\documentclass[12pt, a4paper]{report}
\usepackage[utf8]{inputenc}
\usepackage[acronym,toc,nopostdot,style=super,nonumberlist]{glossaries}
\usepackage[margin=3 cm]{geometry}
\usepackage{hyperref}
\hypersetup{hidelinks,breaklinks,bookmarks=false}
\usepackage{apacite}
\usepackage{graphicx}
\usepackage[titles]{tocloft}
\usepackage[compact]{titlesec}
\usepackage{booktabs}
\usepackage{multirow}
\usepackage{algorithm}
\usepackage{algpseudocode}
\usepackage{amsmath}
\usepackage{xcolor}
\usepackage{caption}
\usepackage{subcaption}
\DeclareUnicodeCharacter{2212}{-}
\titleformat{\chapter}[hang]{\Large\bfseries}{\thechapter.}{1em}{}
\titleformat{\section}{\large\bfseries}{\thetitle}{1em}{}
\titlespacing*{\chapter}{0pt}{-3em}{1.1\parskip}
\titlelabel{\thetitle\quad}

\graphicspath{{Images/}}
\makeglossaries

\begin{document}

\begin{titlepage}
    \begin{center}
    {\textbf{{\large Addressing the issue of stochastic environments and local decision-making in multi-objective reinforcement learning}}}
    
    \vspace{10mm}
    
    \textbf{Kewen Ding}
    
    \vspace{10mm}
    
    \textbf{Supervised by Peter Vamplew and Cameron Foale}
    
    \vspace{30mm}
    
    { In Partial Fulfillment of the Requirements}\\[12pt]
    { For the Degree of}\\[12pt]
    {Bachelor of Science (Honours)} \vfill
    {Institute of Innovation, Science and Sustainability}\\
    {Federation University Australia }\\
    {University Drive, Mount Helen}\\
    {Ballarat Victoria 3353 Australia}
    \vfill
    
    {November 2022}\\
    Kewen Ding              
    \end{center}
\end{titlepage}

\pagenumbering{roman}


\newacronym{SER}{SER}{Scalarised Expected Return}
\newacronym{ESR}{ESR}{Expected Scalarised Return}
\newacronym{RL}{RL}{Reinforcement Learning}
\newacronym{MORL}{MORL}{Multi-objective reinforcement learning}
\newacronym{MDPs}{MDPs}{Markov decision processes}
\newacronym{MOMDP}{MOMDP}{Multi-objective Markov decision process}
\newacronym{HRL}{HRL}{Hierarchical Reinforcement Learning}
\newacronym{IRL}{IRL}{Inverse reinforcement learning}
\newacronym{SMDPs}{SMDPs}{semi-Markov decision processes}
\newacronym{TLO}{TLO}{Thresholded lexicographic ordering}
\newacronym{DRL}{DRL}{Distributional reinforcement learning}

\begin{center}
{{\bf\fontsize{14pt}{14.5pt}\selectfont {Abstract}}}\\\vspace{1cm}
\end{center}
\addcontentsline{toc}{chapter}{Abstract}
\acrfull{MORL} is a relatively new field which builds on conventional \acrfull{RL} to solve multi-objective problems. One of common algorithm is to extend scalar value Q-learning by using vector Q values in combination with a utility function, which captures the user's preference for action selection. This study follows on prior works, and focuses on what factors influence the frequency with which value-based MORL Q-learning algorithms learn the optimal policy for an environment with stochastic state transitions in scenarios where the goal is to maximise the Scalarised Expected Return (SER) - that is, to maximise the average outcome over multiple runs rather than the outcome within each individual episode. The analysis of the interaction between stochastic environment and MORL Q-learning algorithms run on a simple \acrfull{MOMDP} Space Traders problem with different variant versions. The empirical evaluations show that well designed reward signal can improve the performance of the original baseline algorithm, however it is still not enough to address more general environment. A variant of MORL Q-Learning incorporating global statistics is shown to outperform the baseline method in original Space Traders problem, but remains below 100\% effectiveness in finding the find desired SER-optimal policy at the end of training. On the other hand, Option learning is guarantied to converge to desired SER-optimal policy but it is not able to scale up to solve more complex problem in real-life. The main contribution of this thesis is to identify the extent to which the issue of noisy Q-value estimates impacts on the ability to learn optimal policies under the combination of stochastic environments, non-linear utility and a constant learning rate. In conclusion, this study presents several alternative methods that may be more suitable to overcome noisy Q value estimate issue and also find SER optimal policy in MOMDPs with stochastic transitions.  
\begin{center}
{\bf\fontsize{14pt}{14.5pt}\selectfont {Acknowledgements}}\\\vspace{1cm}
\end{center}
\addcontentsline{toc}{chapter}{Acknowledgement}
First and foremost, I would like to thank both of my supervisors, Peter Vamplew and Cameron Foale. They support me as I navigate the difficulties of remote learning, particularly during a difficult period like this pandemic. Also for their insightful comments and participation throughout my entire learning and thesis-writing processes.\newline\newline
Secondly, I would like to thank my Honours course coordinator Rob Bischof. I was able to enrol into the Honours program on time and continue with my study schedule thanks to his assistance, along with my supervisors.\newline\newline
Last but not least, I would want to express my appreciation to my parents. For giving birth to me at the first place and providing for my financial assistance when I was studying abroad.\newline\newline
Thank you all for your support.\newline\newline  
\begin{center}
{\bf\fontsize{14pt}{14.5pt}\selectfont {Statement of authorship}}\\\vspace{1cm}
\end{center}

\addcontentsline{toc}{chapter}{Statement of Authorship} 

Except where explicit reference is made in the text of the thesis, this contains no material published elsewhere or extracted in whole or in part from a thesis by which I have qualified for or been awarded another degree or diploma. No other person's work has been relied upon or used without due acknowledgement in the main text and bibliography of the thesis.

\vfill
\begin{center}
{Signed: \includegraphics[height=1.5\baselineskip]{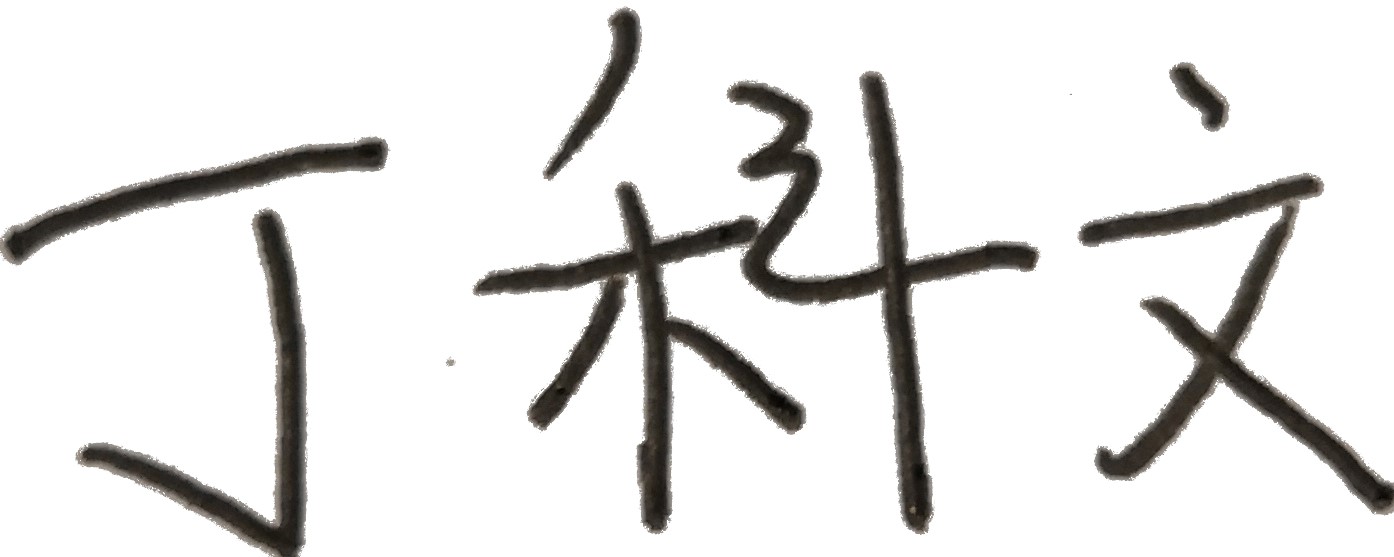} }\\
{Name: Kewen Ding }\\
{Date: November 2022}
\end{center} 

\renewcommand{\contentsname}{\centerline{\textbf{{\large Table of Contents}}}}
\tableofcontents
\newpage

\addcontentsline{toc}{chapter}{\listfigurename} 
\renewcommand*\listfigurename{\centerline{\textbf{{\large List of Figures}}}}
\listoffigures
\newpage

\addcontentsline{toc}{chapter}{\listtablename}
\renewcommand*\listtablename{\centerline{\textbf{{\large List of Tables}}}}
\listoftables
\newpage

\printglossary[type=\acronymtype, toctitle=List of Abbreviations, title=\centerline{\textbf{\large{List of Abbreviations}}}]%
\newpage


\clearpage
\pagenumbering{arabic}

\chapter{Introduction}
\label{Introduction}
\section{Context and Motivation}
\acrfull{RL} \cite{sutton2018reinforcement} is a computational approach which enables an autonomous agent to learn from interactions in a particular environment with a scalar reward signal. The goal of the \acrshort{RL} agent is to learn an optimal policy which determines the best action to take at each time step for maximizing its long-term rewards. It is a major class of machine learning methods to solve sequential decision making problems. \\\\
However, while there are numerous successful applications like AlphaGo \cite{silverMasteringGameGo2016}, the majority of researchers in reinforcement learning only focus on single objective or translate multiple objectives into one scalar value using a simple linear combination function \cite{Hayes2022practicalguide}. This is despite the fact that many real-world problems usually have to trade-offs between multiple or even conflicting objectives. For addressing these type of problems, a new field of study \acrfull{MORL} was introduced. \\\\
Compared with traditional \acrshort{RL}, agent will receive a vector value as reward instead. But this type of vector rewards also create a new challenge in which the best action is less obvious because it requires some methods for sorting those vector rewards. In contrast to \acrshort{RL}, where the best action is the one with the highest expected reward. For this reason, an additional utility function is often used to record user preferences for balancing the trade-offs between the multiple objectives. There are two types of utility function which are linear scalarisation function and non-linear scalarisation function \shortcite{RoijersServey2013}. \\\\
Also there are two types of goal as the result of vector value, \acrfull{ESR} and \acrfull{SER}. The aim of \acrfull{ESR} is to maximise utility outcome within each individual episode. In contrast \acrlong{SER} is for achieving the optimal utility over multiple executions. However, in terms of \acrshort{MORL}, relatively little has been done in comparison to conventional \acrlong{RL} \shortcite{VamplewEmpirical2011}. Additionally, it has just recently been discovered that some of the existing algorithms within the context of non-linear scalarisation function and \acrlong{SER} may not converge to the optimal deterministic policy under stochastic environments \shortcite{VamplewEnvironmental2022}. Taking in account that making decisions in the stochastic environment is the most common thing in the real world. Therefore, the combination of stochastic state transitions and the need for a deterministic policy under \acrshort{MORL} context are likely to arise in a range of applications \shortcite{vamplew2018human}. \\\\
For example, autonomous vehicles should not only reach the desired destination, meet time constraints, provide a comfortable ride, obey road rules, minimise fuel consumption, but also ensure the safety of its occupants, pedestrians and other road users. This makes the problem as multi-objective. Meanwhile autonomous vehicles should also learn to make decision under stochastic environment. In another words, there are always some level of randomness in the driving environment (bad weather, poor road condition and misbehaviour from other road users) which cannot be determined by the autonomous agent. \\\\
As a result, considering the potential range of applications and the gap in the current field of study formed the foundation of the motivation for this research project.
\section{Purpose}
The main purpose of this research is to discover what factors influence the frequency with which value-based \acrshort{MORL} algorithms learn the \acrshort{SER} optimal policy for the environment with stochastic state transition. In order to fulfill this goal, several different methods based on existing algorithms were investigated. The performance of each algorithm was measured on the accuracy of finding the desired optimal policy which will be discussed later in the Methodology chapter.
\section{Scope and aim}
In line with the previous study which identified this problem \shortcite{VamplewEnvironmental2022} this research is scoped to only focus on value-based methods and leave policy-based methods untouched. A more detailed explanation for each terminology will be covered in Literature review chapter. The aim of this project is to answer the following sub research questions.
\begin{itemize}
    \item How different reward signal affects \acrshort{MORL} agent's performance in stochastic environments?
    \item What impact does the augmented state with global statistics have on \acrshort{MORL} agent's ability to learn in stochastic environments?
    \item What effect does the use of the options have on \acrshort{MORL} agent's capacity to learn under stochastic environments?
\end{itemize}
Each of these questions will be addressed in following chapter (\ref{chapter5} - \ref{chapter8}) of the thesis. 
\section{Thesis structure}
The rest of this thesis is organised as follows: Chapter \ref{chapter2} presents comprehensive literature reviews of all related fields (\acrshort{RL}, \acrshort{MORL}). The methodology (chapter \ref{chapter3}) covers all the common aspects of how this research project was conducted, from experiment framework, reward structures, exploration strategy and hyper parameters. This methodology will apply in the following experiments discussed from Chapters \ref{chapter4} to \ref{chapter8}. Chapter \ref{chapter4} first presents the basic multi-objective Q($\lambda$) algorithm and then discusses and reproduces the result of multi-objective Q($\lambda$) using accumulated expected reward from the previous study by \shortciteA{VamplewEnvironmental2022} which will form the baseline for this research. Chapter \ref{chapter5} discusses the reward engineering for original Space Traders test environment and how it affect agent's performance in stochastic environments. Chapter \ref{chapter6} and Chapter \ref{chapter7} propose and evaluate two different setup for collecting and using global statistics in multi-objective Q($\lambda$) algorithm. Chapter \ref{chapter8} provides option learning as an alternative approach. Chapter \ref{chapter9} re-runs all previous method with decay learning rate because of noisy estimates issue which is identified in Option learning. Chapter \ref{chapter10} summarizes the whole thesis as well as highlights the direction for future research.
\chapter{Literature Review}
\label{chapter2}
\section{Introduction}
As mentioned in the scope section from the introduction, the primary focus of this review will be analysis of value-based methods only. But this chapter will still cover reinforcement learning more broadly for better understanding the research problem. For this reason, this literature review will first contain a broad overview of \acrfull{RL} which commences with required background knowledge for this project and formalizing the research problem as \acrfull{MDPs}. Then, in next section, it introduces \acrfull{MORL} and unique features which are different conventional \acrlong{RL}. After that, section 4 and 5 will present the previous research about model-free and model-based MORL methods. Later, section 6 will discuss problem with MORL value-based methods for SER criteria. 
\section{Reinforcement Learning}
\begin{figure}[hbt!]
    \centering
    \includegraphics[width=15cm]{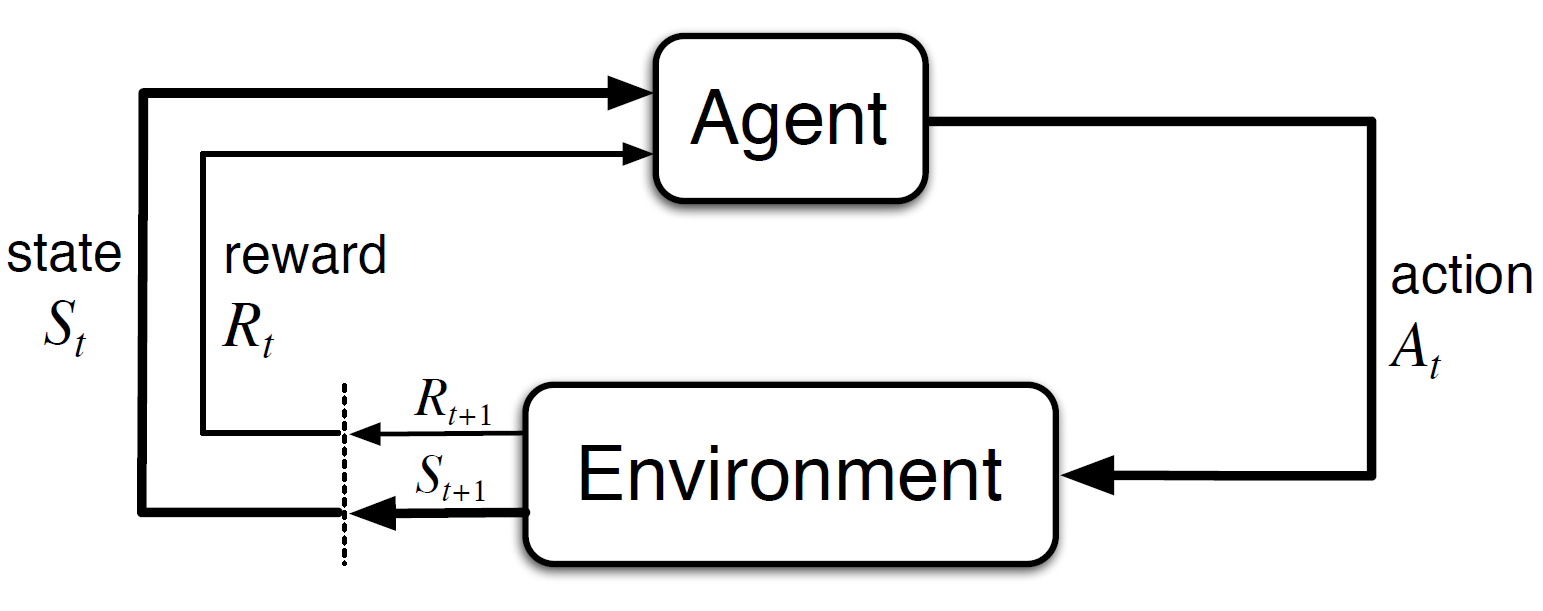}
    \caption[Markov decision process]{Agent-environment interaction in a Markov decision process \protect\cite{sutton2018reinforcement}}
    \label{fig:RL}
\end{figure}

\acrfull{RL} was first inspired by the psychology of animal learning through trial and error \cite{sutton2018reinforcement} and it now has become a major part of machine learning methods.
\acrfull{RL} is the process of learning in an environment, through reward feedback from its own behaviour. It's similar to how children learn to walk without someone showing them how.
The goal of an \acrshort{RL} agent is to learn an optimal policy which determines the best action to take at each step for maximizing its long-term reward. These RL problems are commonly conceptualized as \acrfull{MDPs}. Figure \ref{fig:RL} describes the general \acrshort{RL} framework where an agent repeatedly observes the state from environment, selects and executes an action, and receives a reward which it uses to update its policy. 
A finite Markov decision process is a tuple $\langle S, A, T, R, \mu, \gamma \rangle$ where:
\begin{itemize}
    \item $S$ is a finite set of states ($S^+$ denotes to all state including terminal state)
    \item $A$ is a finite set of actions
    \item $T: S\times A \times S \rightarrow[0,1]$ is a state transition function which define the probability that the agent finds itself in each possible next state after executing action in current state.
    \item $R: S \times A \times S \rightarrow R$ is a reward function which define the expected immediate reward that agent receives in each each possible next state after executing a particular action in current state
    \item $\mu: S \rightarrow[0,1]$ is a probability distribution over initial states
    \item $\gamma\in[0,1)$ is a discount factor specifying the importance of immediate rewards
\end{itemize}
One step of interaction between environment and agent generates a sequence of experience data sample ($s_t,a_t,r_t,s_{t+1}$) which constitutes a trajectory for current episode. The goal of an agent is find a policy $\pi$ mapping states to actions that will maximize the expected sum of discounted rewards defined as
$$G_t=R_{t+1}+\gamma R_{t+2}+{\gamma}^2R_{t+3}+\cdots=\sum_{k=0}^{\infty}{\gamma}^k R_{t+k+1}$$
There are also undiscounted average-reward formulations \shortcite{mahadevan1996average}, but for simplicity we do not consider them in this brief literature review.
\subsection{Deterministic versus Stochastic Environment}
In a deterministic environment, the next state of the environment can always be determined based on the current state and the action chosen by agent. In another word, the state transition function will always be either 0 or 1 for the given states and actions. For example, the board game like Go and Chess can be viewed as deterministic environment. Because at certain board position, a chess player can move knight to F3 and there is only one certain board position.\newline\newline
However, in a stochastic environment, the agent cannot always determine the next state of the environment from the current state by performing a certain action. In another word, there are multiple possible next state for taking certain action and each of them will have a probability which is less than 1. But the sum of those probability will remain to be 1. For example, the card game blackjack is a stochastic environment. Because both player and dealer don't know which card they may get next.
\subsection{Value-based versus Policy-based methods}
Value-based algorithms aim to learn the values of states or actions, such as state-value function or action-value function (i.e. they learn to estimate the future value of $G_t$ starting from each state by following the optimal policy). The policy here is implicit and it can be derived from these value function directly (for the current state, simply pick the action with the best value).\newline\newline
Policy-based algorithms aim to learn the policy directly which can select actions without requiring the value function. A value function may still be used to learn the policy parameter, but is not required for action selection\cite{sutton2018reinforcement}.\newline\newline
This thesis will exclusively discuss value-based methods, ignoring policy-based approaches because value-based methods have been the most commonly used so far in MORL and these are the methods for which the stochastic-environment learning issue has been identified \shortcite{VamplewEnvironmental2022}.
\subsection{Model-Free versus Model-Based method}
One of the most common approaches taken in the RL research are Model-Free methods, so named because they do not required extra knowledge about the dynamic of the environment. In another words, these types of algorithms can learn to find out optimal policy without rewards functions and state transition function. And the most famous model-free method is Q-learning \cite{1989qlearning} which is a value-based reinforcement learning algorithm used to find the optimal action-selection policy through Q function. The optimal Q function $Q^*(s,a)$ can be found using this recursive updating rule which is based on the Bellman equation.
$$Q(s,a)\leftarrow Q(s,a) + \alpha[R(s,a) + \gamma \text{max}Q(s^\prime,a^\prime)-Q(s,a)]$$
Where $\alpha$ is the learning rate and $\text{max}Q(s^\prime,a^\prime)$ refers to the maximum Q value for next state and action\\\\
Another parts of RL researches focus on Model-based Method. Compared with Model-free method, Model-based RL methods need to know the model of the environment (that is, the state transition and reward functions). And it can be divided into two categories
\begin{itemize}
    \item Model-based RL with a known model, where agent plans over a known model, and then learn a optimal policy.
    \item Model-based RL with a learned model, where agent needs to both learn a model and also find out optimal policy. An example is Dyna \cite{1991Dyna}
\end{itemize}
For the Model-based RL with a known model, a good example is AlphaGo \cite{silverMasteringGameGo2016} as the board-game Go is a deterministic environment in which the transition function $T$ is easily defined as it is directly specified by the rules of the game. However RL agent is still required to learn the optimal action in each board position. On the other hand, methods with learned model are useful when the transition function $T$ is not known in advance, which is particularly likely in stochastic environments, or where the agent is interacting with a real world physical environment.\newline\newline
There is another type of method which integrate planning over a learned model. However, this type of approach is not considered model-based RL\shortcite{moerlandModelbasedReinforcementLearning2022}. But it is still worth mentioning here because it could be a potential solution for the research problem.\newline\newline
The main advantage of model-free methods over model-based methods is that there is no need to learn a precise model for the environment. Which means first model-free method can save additional computational cost and also memory for training and storing the model. And second, model-based method can be unstable due to approximation errors in the model \shortcite{moerlandModelbasedReinforcementLearning2022}. But on the other side, model-based method will have better data efficiency because once the model has been learnt there is no need to interact with the environment. So model-based method can reduce the simple cost for example in robotics, the cost for running a series of experiment will be much higher compared with simulation (a model of the environment).
\subsection{Reward Engineering}
Compared with another class of machine learning method - supervised learning, the main advantage of \acrlong{RL} is that does not rely on detailed human instruction. In another word, the rewards received by agent does not depend on knowledge of what correct or wrong actions should be taken. However, any successful reinforcement learning applications strongly depend on how well the reward signal is designed. Here are two aspects need to be considered during the reward designing phase. 
\begin{itemize}
    \item How well the reward signals match with the goal of the application's designer.
    \item How well the reward signals reflect the progress for reaching that goal \shortcite{sutton2018reinforcement}. 
\end{itemize}
Because these reasons, reward design is the most important part of \acrfull{RL}.
Designing the reward here refers to designing the part of an agent’s environment which is responsible for sending reward signal to the agent at each time \textit{t}.\newline\newline
Sometimes, designing a reward function is pretty straightforward when the application's designer has enough knowledge of the problem. For example, considering the game of chess. There are only three possible outcome: win (good), loss (bad), or draw (neutral). So, reward signal for the agent could be $+1$ when it wins the game, $−1$ when it loses, and 0 when it draws or for any other situation. However, this is not always the case. Designing a proper reward function can be a very difficult task because it may have many aspects which need to be considered in order to align the goal of problems with actual desired outcome. For example, consider the driving agent discussed earlier, it is simply not enough to only provide $+1$ for reaching the destination and $-1$ for not arrive at destination. Time constraints, comfortable ride, fuel consumption and most important safety also need to be designed into reward function.\newline\newline
Because of this reason, in practice, designing a reward function is often a trial-and-error and engineering process \shortcite{sutton2018reinforcement}. For example, if the agent fails to learn the desired optimal policy or simply learns too slow. Then the designer need to tweaks the reward signal until agent produces acceptable results.\newline\newline
Of course, this trial-and-error approach is not ideal, and sometimes it can be impractical (for example, increase the training cost for agent) and it could also lead to the undesirable, or even dangerous result. One well known example is CoastRunners \shortcite{Clark_Amodei_2016}. The goal of the game is to finish the boat race quickly and ahead of other players. However, CoastRunners does not gain  reward from the progression of the course, instead it earns higher scores by hitting the targets along the route, and the agent trained on this reward may fail to actually complete the race, instead opting to drive in circles to repeatedly collect targets.\newline\newline
In order to define a better reward function, here are some alternative approaches. The first one is \acrfull{IRL} which learns the reward function via observing an expert demonstrating the task \shortcite{abbeel2004apprenticeship}. However, the first approach is not directly applicable for problem that are difficult for humans to demonstrate. So the second approach is to incorporate human feedback in the RL algorithms and to use this feedback to define the task.\shortcite{christiano2017deep} \shortcite{bignold2021conceptual}. Due to the length constraints of the thesis, other methods would not be covered such as transfer learning for RL.
\subsection{Options Framework}
\begin{figure}[hbt!]
    \centering
    \includegraphics[width=12cm]{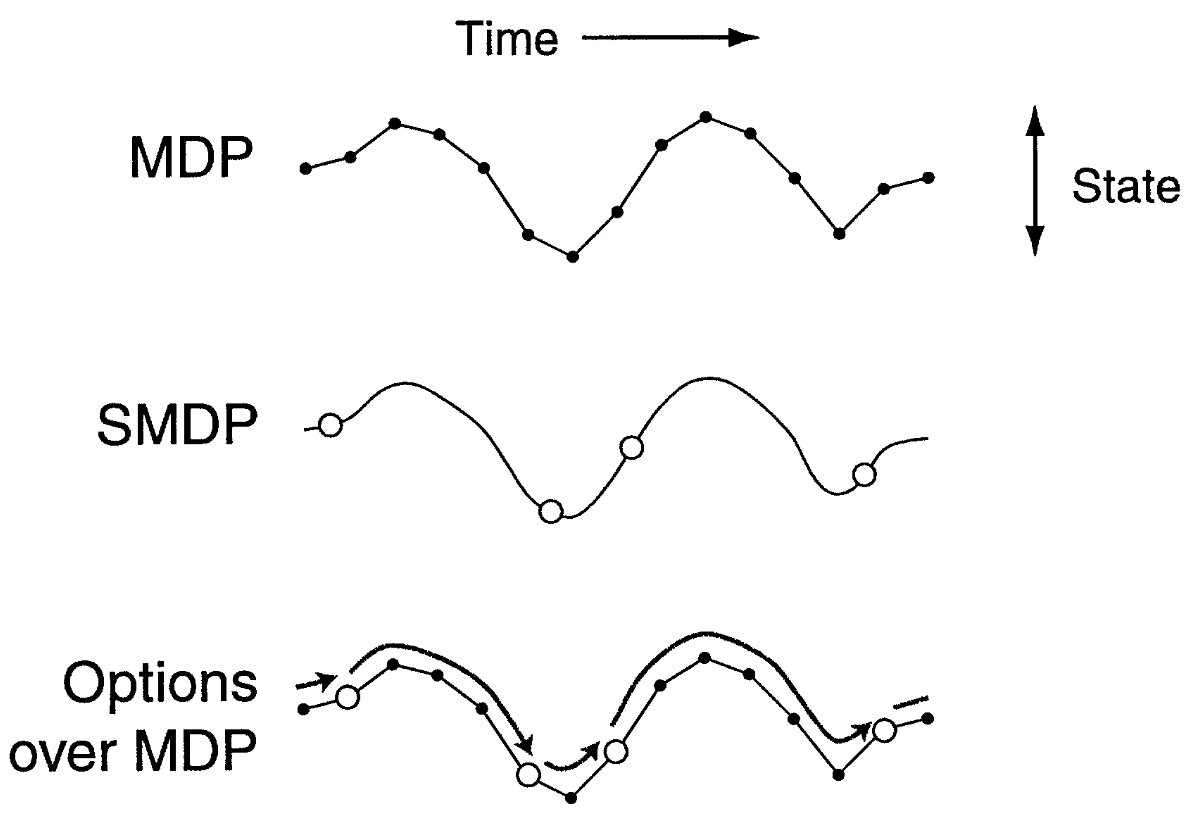}
    \caption[MDP vs SMDP vs Options over MDP]{The different between MDP, SMDP and Options over MDP \protect\cite{sutton1999between}}
    \label{fig:SMDP}
\end{figure}

The capacity to reason at different temporal abstraction levels is one of the fundamental traits of intelligence.
Options, which are temporally expanded courses of action, are frequently used to represent this in reinforcement learning. \shortcite{sutton1999between}. In another word, option is simply following a closed-loop policy by taking actions over a period of time when the termination condition is met. Examples of options include picking up an object, going to lunch, and traveling to another city. Each of these options involves one or more low level actions, such as choosing which muscles to twitch in order to grab the object. Here to formalize the term \textit{Options} $\omega$, it is a 3-tuple $\omega=\langle \mathcal{I}_\omega, \pi_\omega, \beta_\omega \rangle$ where:
\begin{itemize}
    \item $\mathcal{I}_\omega \subseteq \mathcal{S}$ denotes the option's initiation set. (Option is available in state $s_t$ if and only if $s_t\in \mathcal{I}_\omega$)
    \item $\pi_\omega:S\times A \rightarrow [0,1]$ denotes the option's policy where $\sum_a \pi_\omega(\cdot ,a)=1$
    \item $\beta_\omega:S \rightarrow [0,1]$ denotes the option's termination condition.(The probability that option $\omega$ will terminate at a given state)
\end{itemize}
Notice that the actions originally defined in the \acrshort{MDPs} is a special case of options - each action $a$ corresponds to an option whose policy picks the action ($\pi_\omega(s)=a \text{ for all }s\in\mathcal{S}$) and termination function is zero ($\beta_\omega(s)=0 \text{ for all }s\in\mathcal{S}^+$) \shortcite{sutton2018reinforcement}. For this reason, Options are actually interchangeable with low-level actions. For instance, the action-value function $q_\pi$ naturally generalises to an option value function that accepts a state and an option as input and get the expected rewards starting from that state, running that option to termination by executing the policy $\pi$ \shortcite{sutton2018reinforcement}. \newline\newline
The \acrfull{SMDPs} is a special type of decision problem where the options and the actions are closely related. According to \shortciteA{sutton1999between} theorem, "For any MDP, and any set of options defined on that MDP, the decision process that selects only among those options, executing each to termination, is an SMDP". The different between MDP, SMDP and fixed set of options over MDP are suggested in Fig \ref{fig:SMDP}. The top panel shows the state trajectory over discrete time of MDP, the middle layer displays the state changes over continuous random time of SMDP, and the last layer demonstrates how these two levels of analysis can be combined using options. The fundamental basic system in this case is an MDP with regular, one-step transitions, whereas the options define potentially larger transitions, similar to those state in SMDP, that may continue for many discrete steps. This thesis will not go into the details about SMDP.
\section{Multi-objective reinforcement learning}
Now it is time to introduce the basic multi-objective sequential decision problem which can be formalised as a \acrfull{MOMDP}. It is represented by the tuple $\langle S, A, T, \mu, \gamma, \bf{R} \rangle$ where:
\begin{itemize}
    \item $S$ is a finite set of states
    \item $A$ is a finite set of actions
    \item $T: S\times A\times S\rightarrow[0,1]$ is a state transition function
    \item $\mu: S\rightarrow[0,1]$ is a probability distribution over initial states
    \item $\gamma\in[0,1)$ is a discount factor 
    \item $\bf{R}$ $: S\times A\times S\rightarrow R^d$ is a vector-valued reward function which is define the immediate reward for each of the $d\geq2$ objectives.
\end{itemize}
So the main difference between a single-objective MDP and a MOMDP is the vector-valued reward function $\bf{R}$, which specifies a numeric reward for each of the considered objectives. The length of the reward vector is equal to the number of objectives.\newline\newline
Here is a simple example of MORL problem, consider to plan a trip from the current location to a given destination. Selecting on the modes of transportation within a travel plan typically involves a number of objectives, such as minimising the travel time and cost as the same time maximising comfort and reliability. Traveling by car maybe faster and more comfortable compared with subway, but the cost will be more expensive and less reliable due to the possibility of a car accident. When a journey involves many transportation types (e.g. airplane, train, bus or even walking), a policy is required to switch between different travel option due to the delay or malfunction during the journey. In order to solve MORL problem like the previous example, two approaches have been researched so far.
\subsection{Axiomatic versus Utility approach}
\begin{figure}[hbt!]
    \centering
    \includegraphics[width=8cm]{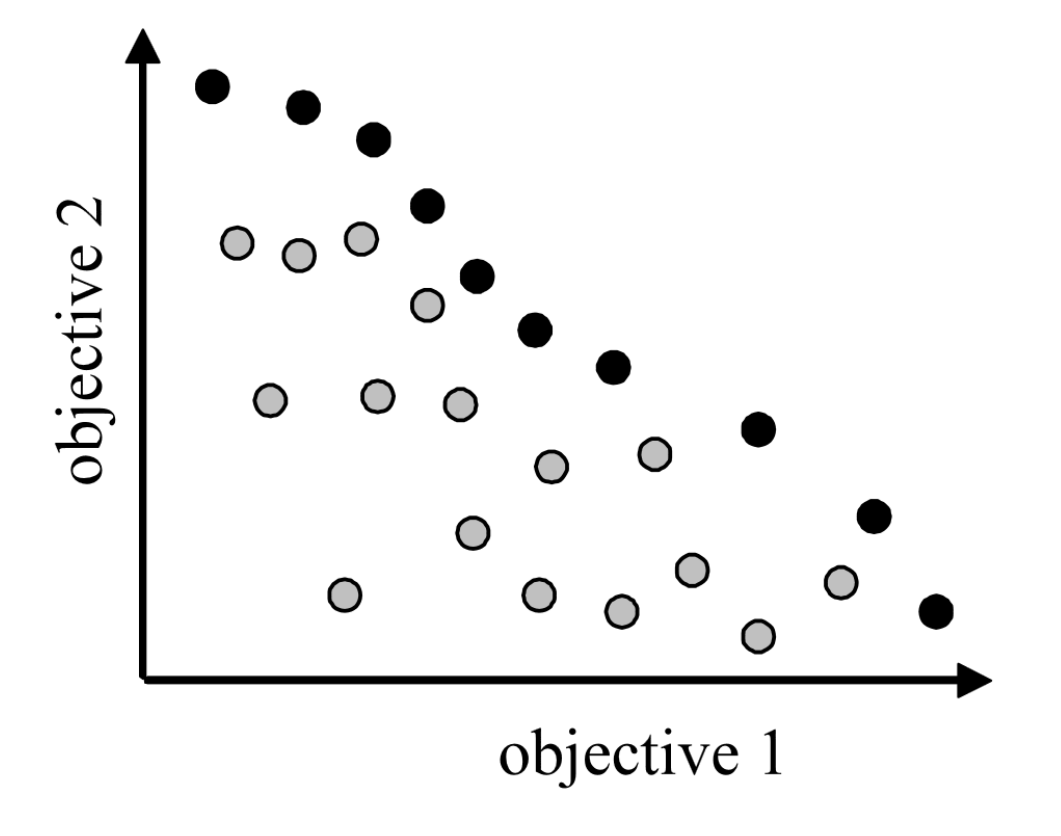}
    \caption[Pareto front]{The black points indicate all non-dominated solutions which are from the Pareto front; The grey points are solutions dominated by at least one solution from Pareto front \protect\shortcite{issabekovEmpiricalComparisonTwo2012}}
    \label{fig:PF}
\end{figure}

The axiomatic approach is to construct the Pareto Front as the optimal solution set which is contained all the non-dominated policy \shortcite{Hayes2022practicalguide}. For each non-dominated policies, there exist no other policy with value that is equal or better in all objectives. In Figure \ref{fig:PF}, each black point indicates a non-dominated policy and each grey point is Pareto dominated by at least one Pareto optimal policy. Non-dominated and Pareto optimal have the same meaning in this thesis. 
However, using Pareto front as the optimal solution set is typically large and it is often computationally expensive to construct. Therefore, there has been a trend in recent literature to adopt a utility-based approach which was proposed by \shortciteA{RoijersServey2013}. Compared with axiomatic approach, the utility approach can utilise domain knowledge for building a utility function which is used for capturing preference of the solution and it can broadly divide into two categories which are linear scalarisation functions and monotonically increasing (nonlinear) scalarisation functions
\subsection{Linear and non-linear scalarisation function}
\label{scalarisation function}
\begin{figure}[hbt!]
    \centering
    \includegraphics[width=8cm]{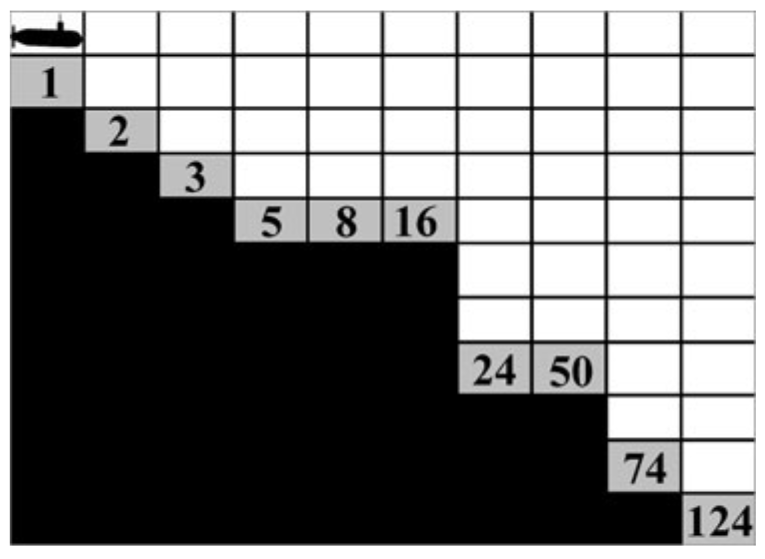}
    \caption[Deep Sea Treasure (DST)]{The Deep Sea Treasure Problem. Black cells indicate the sea-floor; Grey cells indicate the treasure locations. The submarine symbol represent the position in which the agent start at each episode \protect\shortcite{VamplewEmpirical2011}}
    \label{fig:DST}
\end{figure}

Compared with traditional single-objective RL, action selection in MORL agent is not immediately obvious. Because there maybe more than one optimal action to choose from (in term of Pareto optimality). So the most simple and intuitive way to address this issue is to apply a linear scalarisation function. And one of the common approach is to compute the weighted sum of the values for each objective \shortcite{NatarajanDynamic2005}. The advantage of linear approach is to allow the user to have some control over the nature of the solution by placing more or less emphasis on each of the objectives. However, it also suffers from a fundamental disadvantage that any algorithm only combined with linear objective is incapable of finding solutions which lie in the concave region of the Pareto front \shortcite{2008On}. Also in some situation, linear scalarisation function is not enough to handle all types of user preference. For example, human-aligned AI must take into account both its primary goal and its ethical or other constraints in each decision it makes \shortcite{vamplew2018human}. Providing a weight on ethical constraint is simply unacceptable.\newline\newline
Therefore, monotonically increasing (nonlinear) scalarisation functions are introduced. It adheres to the constraint that if a policy increases for one or more of objectives without decreasing any of the objectives, then the scalarized value also increases \shortcite{Hayes2022practicalguide}. One notable example,  \shortciteA{Z1998Multi} first introduced the thresholded lexicographic ordering (TLO) method which allows agent to select actions prioritised in one objective and meet specified thresholds on the remaining objectives.\newline\newline
\shortciteA{RoijersServey2013} previously discussed the issue that under nonlinear function (such as TLO) the rewards are no longer additive which violates the usage of the Bellman equation for value-based method. A good example is Deep Sea Treasure (DST) problem which was originally proposed by \shortciteA{VamplewEmpirical2011} as shown in Fig \ref{fig:DST}. The goal of this agent is to control a submarine for searching deep sea treasure. There are two objectives the first one is to minimise the time taken to reach the treasure and second one is to maximise the value of the treasure. The submarine receives a $-1$ time penalty for each time step it takes to retrieve treasure. When the submarine doesn't find a treasure during that time step, it receives a treasure score of 0, and when it does, it receives a treasure score equal to the value of the treasure. In the previous study \shortcite{VamplewEnvironmental2022}, the threshold for the time objective set to be $-16$, since the TLO agent selects action only according to current state value. As the result, the agent simple ignores the time when deciding whether its future actions will result in exceeding the time threshold.
Therefore, applying scalarisation functions to select action only based on the current state is no enough to generate the result which actually maximise function over the rewards from entire episode. So action selection must be conditioned on current state as well as a summary of the history. \shortciteA{Geibel2008Reinforcement} discussed to use \textit{augmented state} which consisted of both current state value and the accumulating the reward in the current episode for action selection. But this approach only solve non-additive problem in deterministic environment.
\subsection{Single policy versus multi-policies}
Whether an algorithm need to find a single or multiple policies is fully dependent on whether or not user is able to provide the utility function prior to the learning or planning phase. For example, if users already have the ideas about how to balance and trade-off between each objectives. In this case, the utility function is known in advance and fixed, therefore there is need to learn multiple policies as the agent can simply find the optimal policy which maximises that utility. On the other hand, if the futility function can not be designed before the training or the preference could change over time. Then agent has to return a set of all Pareto optimal policies. The use will then select from this set to determine which policy will be used in a particular episode. \newline\newline
Still using previous planning a trip as an example, the traveler  may or may not know exact preferences about getting to the destination in term about when to get there and how much the traveler is willing to spend on this journey. So in this case, the algorithm need to learn all non-dominated policies. However, if the traveler has a preference about how long it need to take in order to arrive the destination or there is a certain budget associated with this trip. Than a single policy will be enough to represent user's preferences.
\subsection{Scalarised expected returns versus Expected scalarised returns}
According to \shortciteA{RoijersServey2013}, there are two distinct optimisation criteria compared with just a single goal in conventional RL\footnote{The conventional single-objective RL does not use a scalarisation function, the ESR and SER criteria are the same in this context.}. The first one is \acrfull{ESR}. In this approach, agent aims to maximise the expected value which is first scalarised by utility function for each state. Equation is shown below (Eq \ref{eq:ESR}) where $w$ is the parameter
vector for utility function $f$, $r_k$ is the vector reward on time-step $k$, and $\lambda$ is the discounting factor
\begin{equation}
\label{eq:ESR}
V^\pi_{\bf w}(s) =  E[f( \sum^\infty_{k=0} \gamma^k \mathbf{r}_k,{\bf w})\ | \ \pi, s_0 = s)
\end{equation}
ESR is the appropriate criteria for problems where the aim is to maximise the expected outcome within each individual episode. A good example is searching a treatment plan for a patient, where there is a trade-off between cure and negative side-effect. Each patient would only care about their own individual outcome instead of total average. The second criteria is \acrfull{SER} which estimate the expected rewards for episode and then maximise the scalarised expected return. Equation is shown below (Eq \ref{eq:SER})
\begin{equation}
\label{eq:SER}
V^\pi_{\bf w}(s) = f( {\bf V}^\pi(s), {\bf w}) =  f(E[\sum^\infty_{k=0} \gamma^k \mathbf{r}_k \ | \ \pi, s_0 = s], {\bf w})
\end{equation}
So \acrshort{SER} formulation is used to achieving the optimal utility over multiple executions. Continuing with the travel example, the employee wants to cut down on the amount of time spent traveling to work each day. Traveling by car would be the good option on average, although there may be rare days on which it is considerably slower due to an accident. 
\section{Model-Free MORL Method}
One of the most common approaches taken in the RL research are Model-Free methods because they do not required extra knowledge about the dynamic of the environment. So the most intuitive and simplest way in MORL literature is to extend those existing RL methods such as Q-learning \cite{1989qlearning} to deal with multiple objectives. There are two changes required in this extended MOQ-learning. First, the agent will store Q-values as vectors instead of scalars. Second, a scalarisation function is designed to select greedy action in each given state. Based on different form of scalarisation function will get new variant of MOQ-learning algorithm. 
\subsection{Weighted sum approach}
The weighted scalarisation approach is well known linear method by assigning different weight for each objective based on their importance. For example, \citeauthor{1995Constrained} \citeyear{1995Constrained} used a weighted sum of the discounted total rewards for multiple reward type. Compared with the fixed weight method, \citeauthor{NatarajanDynamic2005} \citeyear{NatarajanDynamic2005} propose a dynamic preferences method which stores a finite number of policies, choose an appropriate policy for any weight vector and improve upon it. In order to overcome the drawback of the linear scalarisation function mentioned in section \ref{scalarisation function}, \shortciteA{2020Q} proposed a new algorithm called Q-Managed, which is combining the use of linear scalarisation function (no weight for any of the objectives) with $\epsilon$-constraint. This algorithm is also capable to learn viable solutions which is not limited to convex hull. However, they only researched on episodic problem so far in their paper and leave the work for future validation on other class of problems.
\subsection{Ranking approach}
The second approach called ranking or threshold, it aims to provide an ordering or preference among multiple criteria. \citeauthor{Z1998Multi} \citeyear{Z1998Multi} introduced the \acrfull{TLO} for MORL where threshold values were specified for some objectives in order to put the constraints on the objectives and leave last objective unthresholded. \shortciteA{VamplewEmpirical2011} combined TLO with Q-learning (TLQ-learning) in multi-objective reinforcement learning problem. The experiment showed that TLQ-learning converged to a Pareto optimal solution in fewer episodes compared with normal Scalarised Q-learning. However due to the nature of this approach, it requires some knowledge about the problem under consideration. More precisely, first $n-1$ objectives need to have a threshold value which clearly requires some prior expectation about the range of values for each objective. One way to obtain the required knowledge is to use the results produced by the weight sum algorithm \shortcite{issabekovEmpiricalComparisonTwo2012} which is already discussed in the previous section. Another way is to use a dynamic thresholds which proposed by \shortciteA{HayesDynamic2020}. The required knowledge for setting each threshold is extracted from the evaluations of the system's previous performance. 
\subsection{Chebyshev approach}
\shortciteA{2013Scalarized} proposed to use a non-linear scalarisation function, called the weighted Chebyshev scalarisation function. It can not only discover Pareto optimal solutions regardless of the shape of the front but is also not dependent on the actual weights used by comparing with weighted sum approach. Here is one of the application for using Chebyshev scalarisation function in MORL problem. \shortciteA{qin_wang_yi_li_zhai_2021} designed a deadline constrained  scientific workflow  scheduling  algorithm  based  on MORL called DCMORL. It aims to find out  a  solution  set,  which minimizes the cost and energy consumption for cloud computing.
\subsection{Voting approach}
\shortciteA{tozerManyobjectiveStochasticPath2017} proposed a many objective reinforcement learning algorithm called Voting Q-learning (VoQL) which is able to find a set of optimal policies in stochastic environment with several confilicting objectives. This algorithm is based on voting methods from social choice theory. However, they only evaluated VoQL in environment with stochastic rewards but deterministic state transitions. As the result, this algorithm may not work directly in our research problem.
\subsection{Multi-policies approach}
Previous approaches are mainly focus on single policy algorithm, in that situation, user already provides associated utility function. Whether it is a weighted vector or a set of threshold for each objective. However, at the beginning of learning, there may be no domain knowledge for constructing a utility function for action selection. So instead of finding only one optimal policy, providing a set of non-Pareto dominated set will be a good option. \newline\newline
Multi-policy algorithms can be broadly divided into two categories \shortcite{Hayes2022practicalguide}. Outer loop methods operate on a series of single-objective problems to construct an approximate coverage set. So the simplest outer loop methods just iterate through a set of different parameter settings for the utility function and run single-policy MORL method many times for each settings. Compared with outer loop methods, inner loop methods will directly produce multiple policies in a single run and stored in parallel rather than sequentially. \shortciteA{2014Multi} proposed a temporal difference learning (Pareto Q-learning) that learns a set of Pareto dominating polices in a single run. The contribution of this work is to provide a mechanism which learns a separate expected immediate reward vector and the set of expected future discounted reward vectors. But in order to learn the entire Pareto front, each state-action pair is required to be sampled sufficiently. Also they focus on Pareto Q-learning in episodic deterministic environments. When applying Pareto Q-learning in a stochastic environment, the expected immediate and future non-dominated rewards need to be stored in $\langle s, a, s^\prime \rangle$ tuple instead of $\langle s, a \rangle$ pair, also a small model of the transition probabilities need to be learned. In this case, it becomes a model-based method. \shortciteA{RoijersFollowing2021} pointed out that when transition function is stochastic, selecting a value vector and executing the corresponding policy just based knowledge from Pareto Coverage Sets will lead to a following problem in each action selection step.\newline\newline
A similar temporal difference method (MPQ-learning) was described by \shortciteA{Ruiz2017A}. But compared with Pareto Q-learning, this algorithm needs to store some extra information along with the action-vectors inside the sets. So for each vector estimate, it will consist of the current values of the vector estimate and a set of indices. However, high computational cost associated with this algorithm because it aims to learn all deterministic optimal policies paralleled. To improve the efficiency, \shortciteA{mandowPruningDominatedPolicies2018} proposed a modification of MPQ-learning that controls the generation of cycles using suspension rule during the learning.
\section{Model-Based Method}
Another part of RL research focuses on model-based method. Compared with model-free method, model-based RL methods learn to estimate the model from the experiences by interacting with the environment. Some planning algorithm will also be discussed in this section, because only few works have been published in model-based MORL field.\newline\newline
\shortciteA{2007Computing} proposed a multi-objective dynamic programming algorithm called the CON-MODP which can compute all Pareto optimal policies for deterministic multi-objective sequential decision problems. Compared with early work in multi-objective value iteration algorithms \shortcite{WHITE1982639}, this algorithm is much faster to find out all stationary policies by only allows stationary policies to remain in the Pareto optimal set and only single inconsistent state is expanded during policy evaluation step.\newline\newline
Based on CON-MODP algorithm, \shortciteA{wieringModelbasedMultiobjectiveReinforcement2014} proposed a model-based MORL algorithm which combine the model-building methods with a multi-objective dynamic programming method. For more effective building the model, two exploration policies was developed for obtain enough experiences by interacting with the environment. There are least-visited exploration and random-exploration. But still this algorithm is only able to solve deterministic MORL problem.\newline\newline
\shortciteA{barrettLearningAllOptimal2008} described the Convex Hull value Iteration (CHVI) Algorithm which is able to learn optimal policies for all linear weight. But the complexity of this algorithm is exponential when there are more objectives in the problem setting. So this algorithm is restricted to 5 or less objectives problem or applying constraints on the weight for handling many objectives problem.\newline\newline
\shortciteA{yamaguchiModelBasedMultiobjectiveReinforcement2019} proposed an average reward model-based MORL method based on reward occurrence probability (ROP) with unknown weights.The key different between other MORL algorithm is that it learns ROP for each policy instead of normal Q-values. And the Pareto optimal deterministic policies are directly generated from convex hull in the ROP space. So the weight associated with each objective is only calculated once.\newline\newline
\shortciteA{wrayMultiObjectiveMDPsConditional2015} described a model for state-dependent Lexicographic MOMDPs as a subset of MOMDPs and also proposed Lexicographic Value Iteration (LVI) algorithm which combined with slack variables and conditional state-based preferences. By using the concept of slack, it allows some degree of loss in the primary objective in order to obtain gains in the secondary objective. However, there is no threshold for each objective. So this approach is different from TLO method mentioned on previous ranking section.\newline\newline
\shortciteA{bryceProbabilisticPlanningMultiobjective2007} first identified the problem that an agent aiming to maximise the SER cannot rely on local decision-making in \acrshort{MOMDP} with stochastic state transition. According to their research, it is not possible to decide which action is the best only based on current state without considering the actions in other possible state as well. In order to address this problem, they proposed a multi-objective looping AO* (MOLAO*) searching algorithm which is extended the single objective looping AO* (LAO*) algorithm. But in order to apply the MOLAO* algorithm in current research problem, a model of the environment is needed before hand. In addition no implementation of MOLAO* is publically available.
\section{Problem with MORL value-based methods for SER criteria}
After explaining all the required concepts, it is now ready to discuss the problem in stochastic environment. \shortciteA{VamplewEnvironmental2022} reported that using non-linear scalarisation function, current existing value-based model-free MORL methods may fail to find out the \acrshort{SER} optimal policy in environments with stochastic state transitions. Under this type of environment, even following the exactly same policy will receive different rewards each time. Since \acrshort{SER} criteria is used to achieving the optimal utility over multiple execution. The overall policy in order to meet that constraints depends on the probability with which each trajectory is followed as well as the mean outcome of each trajectory. Determining the correct action to select at each possible trajectory requires to consider the actions available at each other trajectory in combination with the probability of that trajectory been followed. In another words, this requirement is fundamentally incompatible with the value-based model-free methods like Q-learning, which is assumed that the best action can be fully determined on the information available to the agent at the current state. The provided information either sum of actual rewards\footnote{The basic MOQ-learning method will be covered in chapter \ref{chapter3}} or expected rewards\footnote{The baseline method will be discussed in chapter \ref{chapter4}} is still insufficient for agent to augment the state. Because both of them still only provide information about the trajectory which has been followed in this episode, rather than all possible trajectory that agent might be able to reach in this same policy.
\section{Summary}
The research discussed in this chapter are significant and have identified most of the crucial areas that relate to the project's aim. The information in this literature review provides a board overview about basic terminology in conventional \acrlong{RL} and existing model-free and model-based methods used in \acrlong{MORL}. Both of them are important to outline the research design and methodology used in this research projected which can be found in the next chapter.
\chapter{Research Design And Methodology}
\label{chapter3}
\section{Introduction}
The research problem discussed in this thesis can be broadly stated as identifying what factors affect action selection in stochastic environment when an agent desires to maximise the \acrfull{SER} under non-linear scalarisation function. To be more specifically, this study focus on using value-based reinforcement learning technique - Q learning combined with different factors and applying them to a simple \acrshort{MOMDP} environment called Space Trader which will be described in the next section. The goal of the agent is to navigate through the environment and learn the optimal policy. This chapter will cover the methodology used in this study, discussing elements common to all the experiments reported in the later chapters, demonstrate the reasons for choosing space trader as testing environment, how each different strategy been implemented and how the results are collected, and lastly how to determine the performance for each algorithm.
\section{Methodology and Environment}
In this research, it uses empirical evaluation as methodology. Following the approach proposed by \shortciteA{VamplewEmpirical2011}, which includes standardized metrics for analysing the quality of the algorithm and benchmark problems with knowledge of actual optimal policies. The reason for using empirical evaluation as methodology is because that is a vital component of machine learning research, particularly in the comparison of algorithms. Unlike other machine learning research, such as Supervised Learning,  which share data set for comparing the accuracy of the algorithm, most of the reinforcement learning algorithm require to publish the implementation of test environments where the agent interacts with and generates data from.\newline\newline
For this study, it uses MORL-Glue \cite{Vamplew2017MORLGlueAB}, a java implementation, as the framework for modelling stochastic environment and comparing each algorithm. Most importantly, the research question was firstly identified by using this framework and also the original testing environment Space Trader was already implemented. So, the result from basic MOQ-learning can be easily reproduced and served as baseline for different algorithms later to compare with.
\section{Experimental Setup}
\subsection{Basic MOQ-Learning}
One of the most common algorithms in the MORL literature is to extend single-objective, model-free value-based Q-learning algorithm. For this paper, it will focus on single-policy form of multi-objective Q-learning and the basic form of algorithm can be found in Algorithm \ref{algo:moql} where utility function $f$ is used to trade-off between multiple Pareto-optimal actions at any state, so that the algorithm will obtain a single policy which is optimal regards to $f$.
\begin{algorithm}
  \caption{A general algorithm for multi-objective Q($\lambda$) which is conditioned on actual reward through out the episode}
  \label{algo:moql}

  \begin{algorithmic}[1]
    \Statex input: learning rate $\alpha$, discounting term $\gamma$, eligibility trace decay term $\lambda$, number of objectives $n$, action-selection function $f$ and any associated parameters
    \For {all states $s$, actions $a$ and objectives $o$}
    	\State initialise $Q_o(s,a)$
    \EndFor
    \For {each episode} 
        \For {all states $s$ and actions $a$}
    		\State $e(s,a)$=0
        \EndFor
        \State sums of prior rewards $P_o$ = 0, for all $o$ in 1..$n$
    	\State observe initial state $s_t$
    	\State $s^A_t$ = $(s_t,P)$ \Comment{create augmented state}
        \State select $a_t$ from an exploratory policy derived using $f(Q(s^A_t))$
        \For {each step of the episode}
 			\State execute $a_t$, observe $s_{t+1}$ and reward $R_t$
 			\State $P = P + R_t$
 			\State $s^A_{t+1}$ = $(s_{t+1},P)$ \Comment{create augmented state}
 			\State $U(s^A_{t+1}) = Q(s^A_{t+1}) + P$ \Comment{create utility value}
 			\State select $a^*$ from a greedy policy derived using $f(U(s^A_{t+1}))$
  			\State select $a^\prime$ from an exploratory policy derived using $f(U(s^A_{t+1}))$
            \State $\delta = R_t + \gamma Q(s^A_{t+1},a^*) - Q(s^A_t,a_t)$
            \State $e(s^A_t,a_t)$ = 1
            \For {each augmented state $s^A$ and action $a$}
            	\State $Q(s^A,a) = Q(s^A,a) + \alpha\delta e(s^A,a)$
                 \If {$a^\prime = a^*$} 
                    \State $e(s^A,a) = \gamma \lambda e(s,a)$
                 \Else
                    \State $e(s^A,a) = 0$                 
                 \EndIf
            \EndFor
            \State $s^A_t = s^A_{t+1}, a_t = a^\prime$
    	\EndFor
    \EndFor
  \end{algorithmic}
\end{algorithm}
\subsection{Space Trader Environment}
\begin{figure}[hbt!]
    \centering
    \includegraphics[width=15cm]{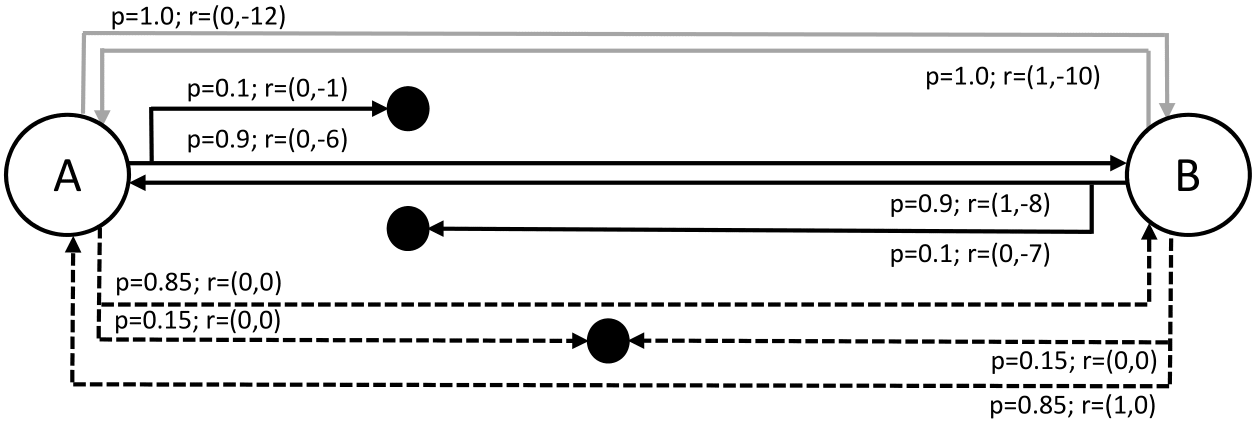}
    \caption[The original Space Traders Environment]
    {The Space Traders MOMDP. Solid black lines show the Direct actions, solid grey line show the Indirect actions, and dashed lines indicate Teleport actions. Sold black circles indicate terminal (failure) state \protect\shortcite{VamplewEnvironmental2022}}
    \label{fig:ST-O}
\end{figure}

Space Traders shown in Fig \ref{fig:ST-O} was first proposed by \shortciteA{VamplewEnvironmental2022}. It is a simple finite-horizon task with only two-steps and it consists of two non-terminal states with three actions (direct, indirect and teleport) available to choose from each state. And agent starts from planet A (State A) and travel to planet B (state B) to deliver shipment and then return back to planet A with the payment. The reward for each action consists of two parts. The first element is whether agent successfully return back to planet A. So agent only receive 1 as reward on last successful action and 0 for all other action including failure. The second element is a negative penalty which indicates how long this action takes to execute. The goal of this agent is to minimise the time taken to complete the travel as well as having at least equal or above 88\% probability of successful completion.
Table \ref{tab:Space-Traders-O1} shows the transition probabilities, immediate reward for each state action pairs and mean rewards as well.
\begin{table}[hbt!]
    \centering
    \resizebox{\columnwidth}{!}{%
    \begin{tabular}{@{}|c|c|c|c|c|c|@{}}
    \toprule
    State              & Action   & P(Success) & Reward on Success & Reward on Failure & Mean Reward \\ \midrule
    \multirow{3}{*}{A} & Indirect & 1.0        & (0, -12)          & N/A               & (0, -12)    \\ \cmidrule(l){2-6} 
                       & Direct   & 0.9        & (0, -6)           & (0, -1)           & (0, -5.5)   \\ \cmidrule(l){2-6} 
                       & Teleport & 0.85       & (0, 0)            & (0, 0)            & (0, 0)      \\ \midrule
    \multirow{3}{*}{B} & Indirect & 1.0        & (1, -10)          & N/A               & (1, -10)    \\ \cmidrule(l){2-6} 
                       & Direct   & 0.9        & (1, -8)           & (0, -7)           & (0.9, -7.9) \\ \cmidrule(l){2-6} 
                       & Teleport & 0.85       & (1, 0)            & (0, 0)            & (0.85, 0)   \\ \bottomrule
    \end{tabular}%
    }
    \caption[The State Action pair in original Space Traders Environment]
    {The probability of success and reward values for each state-action pair in the Space Traders MOMDP \protect\shortcite{VamplewEnvironmental2022}}
    \label{tab:Space-Traders-O1}
\end{table}
The reason for selecting Space Traders as testing environment is because it is relatively small test environment. So, it is easy to list all of the nine possible deterministic policies which are shown in Table \ref{tab:Space-Traders-O2}
\begin{table}[hbt!]
    \centering
    \begin{tabular}{@{}|c|c|c|c|@{}}
    \toprule
    Policy identifier & Action in state A & Action in state B & Mean Reward     \\ \midrule
    II                & Indirect          & Indirect          & (1, -22)        \\ \midrule
    ID                & Indirect          & Direct            & (0.9, -19.9)    \\ \midrule
    IT                & Indirect          & Teleport          & (0.85, -12)     \\ \midrule
    DI                & Direct            & Indirect          & (0.9, -14.5)    \\ \midrule
    DD                & Direct            & Direct            & (0.81, -12.61)  \\ \midrule
    DT                & Direct            & Teleport          & (0.765, -5.5)   \\ \midrule
    TI                & Teleport          & Indirect          & (0.85, -8.5)    \\ \midrule
    TD                & Teleport          & Direct            & (0.765, -6.715) \\ \midrule
    TT                & Teleport          & Teleport          & (0.7225, 0)     \\ \bottomrule
    \end{tabular}
    \caption[Mean return for nine deterministic policies in original Space Traders Environment]{Nine available deterministic policies mean return for Space Traders Environment\protect\shortcite{VamplewEnvironmental2022}}
    \label{tab:Space-Traders-O2}
\end{table}
The parameters used for Space Traders environment can be found in Table \ref{tab:Space-Traders-O3}.
\begin{table}[hbt!]
    \centering
    \resizebox{\columnwidth}{!}{%
    \begin{tabular}{@{}|c|c|c|c|c|c|c|@{}}
    \toprule
    Parameter & $\alpha$ & $\lambda$ & $\gamma$ & softmax-t temperature initial & softmax-t temperature final & Number of episodes per training \\ \midrule
    Value     & 0.01     & 0.95      & 1        & 10                            & 2                           & 20,000                          \\ \bottomrule
    \end{tabular}%
    }
    \caption[Parameter table for Space Traders]{Hyper Parameter used in the Space Traders environment}
    \label{tab:Space-Traders-O3}
\end{table}
\section{Strategy}
This section will discuss four \acrshort{RL} methods including baseline method which have been explored in this research and how each one was implemented. 
\begin{itemize}
\item Baseline approach - It replaces actual accumulated reward in basic MOQ-learning with expected accumulated reward. This is discussed in Chapter 4.
\item Reward engineering approach - It modifies original Space Trader rewards signal and keep using baseline algorithm. This is discussed in Chapter 5.
\item Global statistic approach - It introduces two novel algorithms which include global statistic information for action selection. This is discussed in Chapters 6 and 7.
\item Option approach - It introduces Option as a 'meta-action' which determines the action selection over multiple time-steps compared with single time-step in Baseline method. This is discussed in Chapter 8.
\end{itemize}
\section{Performance Measure}
When using \acrlong{RL} in real-world applications, it is preferable to have algorithms that consistently achieve high levels of performance without the need for extensive human interaction in term of parameter tuning and less computation time\shortcite{jordan2020evaluating}. Therefore empirical RL research often focuses on how quickly an algorithm learns. However learning speed is of secondary importance to fining the optimal policy. So for this thesis, it focuses on which approach can converge to desired optimal policy.
\subsection{Metrics}
In term of metrics, each approach will run through 20 trials of experiments. How many trials out of 20 can agent find out desired \acrshort{SER} optimal policy will be the measurement to decide which approach is the best one. In other words, the more times the algorithm can find out the optimal policy is the better one. 
\subsection{Data Collection}
For each of the four approaches implemented in this thesis, the following data are collected for each of the 20 trials in each experiment.
\begin{itemize}
\item The reward which is collected by agent during 20,000 episodes of training
\item After training, the final policy which has been learned by the agent.
\end{itemize}
\subsection{Reproducible}
To make sure a test is reproducible, the seed for random number generator for each approach is kept. So to reproduce a test result is the matter of reusing the same random number seed from previous test
\section{Conclusion}
This chapter describes the methodology for this thesis, including the common elements which shared by all the experiments, the Space Trader testing environment, the hyper parameters been used, how the results are collected and lastly how to measure the performance for each approach. The next chapter, Chapter 4, discusses the baseline approach which replicates the results reported in \shortciteA{VamplewEnvironmental2022}.
\chapter{Baseline method}
\label{chapter4}
For a fully deterministic test environment, like Deep Sea Treasure (DST) \shortcite{VamplewEmpirical2011} as mentioned early in chapter 2, the SER deterministic optimal policy is the same as ESR deterministic optimal policy. The basic MOQ-learning algorithm \ref{algo:moql} is enough to solve this question using an augmented state which conditioned on current episode. However, in stochastic environment, the deterministic SER optimal policy may be totally different from ESR optimal policy. As SER optimisation cares about the mean result over all episodes, which makes basic MOQ-learning algorithm \ref{algo:moql} not appropriate any more. Even an agent conditioned on the accumulated expected immediate reward when selecting the action is still not enough to address this issue as it is mentioned early in chapter 2. This is the modified version of the basic MOQ-learning algorithm, proposed by \shortciteA{VamplewEnvironmental2022}, who demonstrated that it could reliably find SER-optimal policies for environments with stochastic rewards and deterministic state transitions, but not for environments like SpaceTraders which have stochastic state transitions. The purpose of this chapter is to replicate the results of \shortciteA{VamplewEnvironmental2022}, to serve as a baseline method for comparison with the approaches considered in the later chapters. 
\section{Algorithm}
Here is the accumulated expected reward version of the basic MOQ-learning algorithm (Algorithm \ref{algo:moql-expected}). In order to learn the expected immediate reward, the agent needs to maintain an estimate of these expected immediate rewards for each state-action pair (Line 3 and Line 15 in Algorithm \ref{algo:moql-expected}).
\begin{algorithm}[hbt!]
  \caption{Multi-objective Q($\lambda$) using accumulated expected reward as an approach to finding deterministic policies for the SER context. The differences from Algorithm \ref{algo:moql} have been highlighted in red text}
  \label{algo:moql-expected}

  \begin{algorithmic}[1]
    \Statex input: learning rate $\alpha$, discounting term $\gamma$, eligibility trace decay term $\lambda$, number of objectives $n$, action-selection utility function $f$ and any associated parameters
    \For {all states $s$, actions $a$ and objectives $o$}
    	\State initialise $Q_o(s,a)$
    	\State \textcolor{red}{initialise $I_o(s,a)$ \Comment{estimated immediate (single-step) reward}}
    \EndFor
    \For {each episode} 
        \For {all states $s$ and actions $a$}
    		\State $e(s,a)$=0
        \EndFor
        \State sums of prior expected rewards $P_o$ = 0, for all $o$ in 1..$n$
    	\State observe initial state $s_t$
    	\State $s_t$ = $(s_t,P)$ \Comment{create augmented state}
        \State \parbox[t]{\dimexpr\linewidth-\algorithmicindent\relax}{%
            \setlength{\hangindent}{\algorithmicindent}%
            select $a_t$ from an exploratory policy derived using $f(Q(s))$
        }\strut
        \For {each step of the episode}
 			\State execute $a_t$, observe $s_{t+1}$ and reward $R_t$
 			\State \textcolor{red}{update $I(s_t,a_t)$ based on $R_t$}
 			\State \textcolor{red}{$P = P + I(s_t,a_t)$}
 			\State $s_{t+1}$ = $(s_{t+1},P)$ \Comment{create augmented state}
 			\State $U(s_{t+1}) = Q(s_{t+1}) + P$ \Comment{create value vector}
 			\State select $a^*$ from a greedy policy derived using $f(U(s_{t+1}))$
  			\State select $a^\prime$ from an exploratory policy derived using $f(U(s_{t+1}))$
            \State $\delta = R_t + \gamma Q(s_{t+1},a^*) - Q(s_t,a_t)$
            \State $e(s_t,a_t)$ = 1
            \For {each state $s$ and action $a$}
            	\State $Q(s,a) = Q(s,a) + \alpha\delta e(s,a)$
                 \If {$a^\prime = a^*$}
                 	\State $e(s,a) = \gamma \lambda e(s,a)$
                 \EndIf
            \EndFor
            \State $s_t = s_{t+1}, a_t = a^\prime$
    	\EndFor
    \EndFor
  \end{algorithmic}
\end{algorithm}
\section{Result}
\begin{figure}[hbt!]
     \centering
     \begin{subfigure}[h]{0.45\textwidth}
         \centering
         \includegraphics[width=\textwidth]{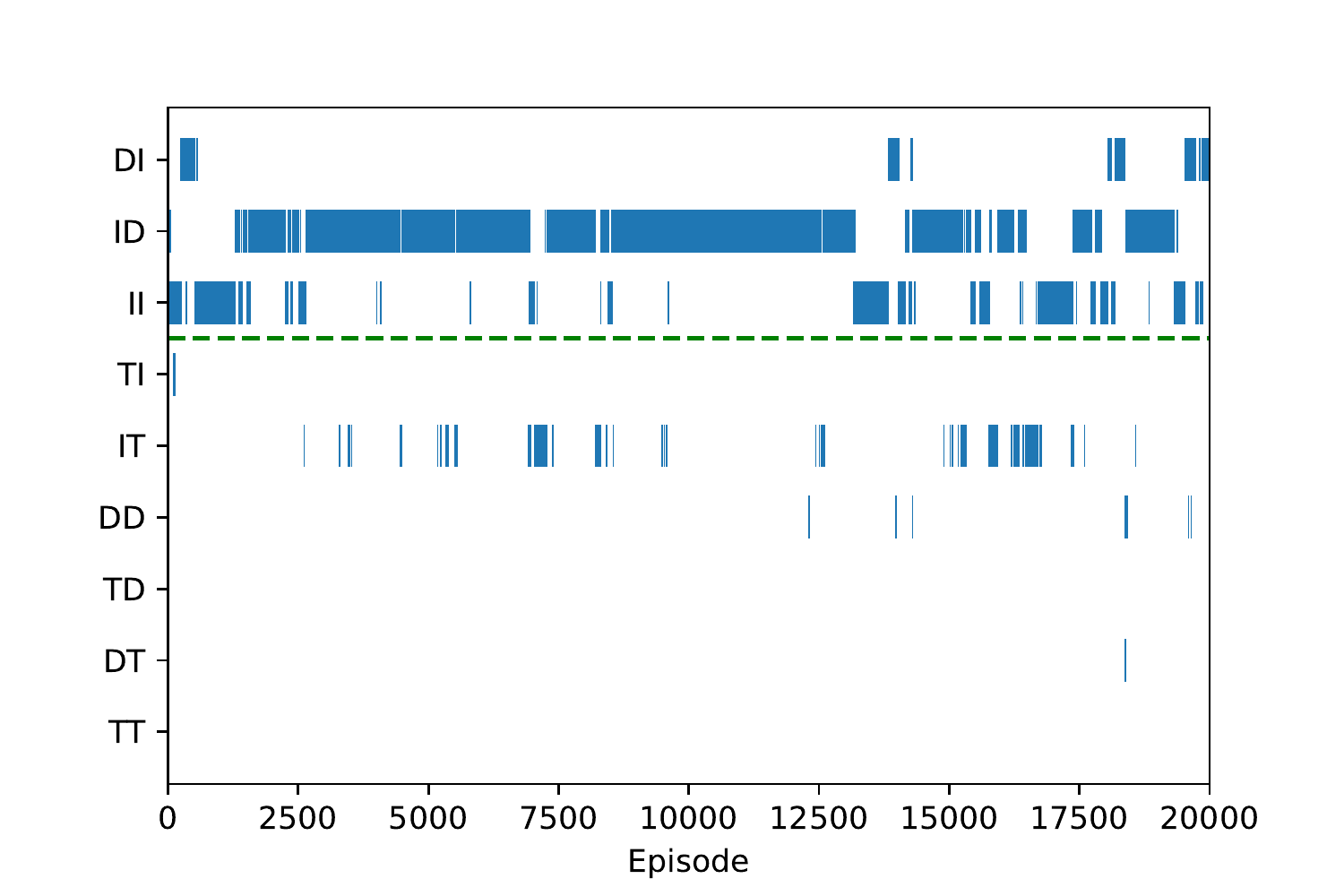}
         \caption{Policy DI}
         \label{fig:Baseline DI}
     \end{subfigure}
     \begin{subfigure}[h]{0.45\textwidth}
         \centering
         \includegraphics[width=\textwidth]{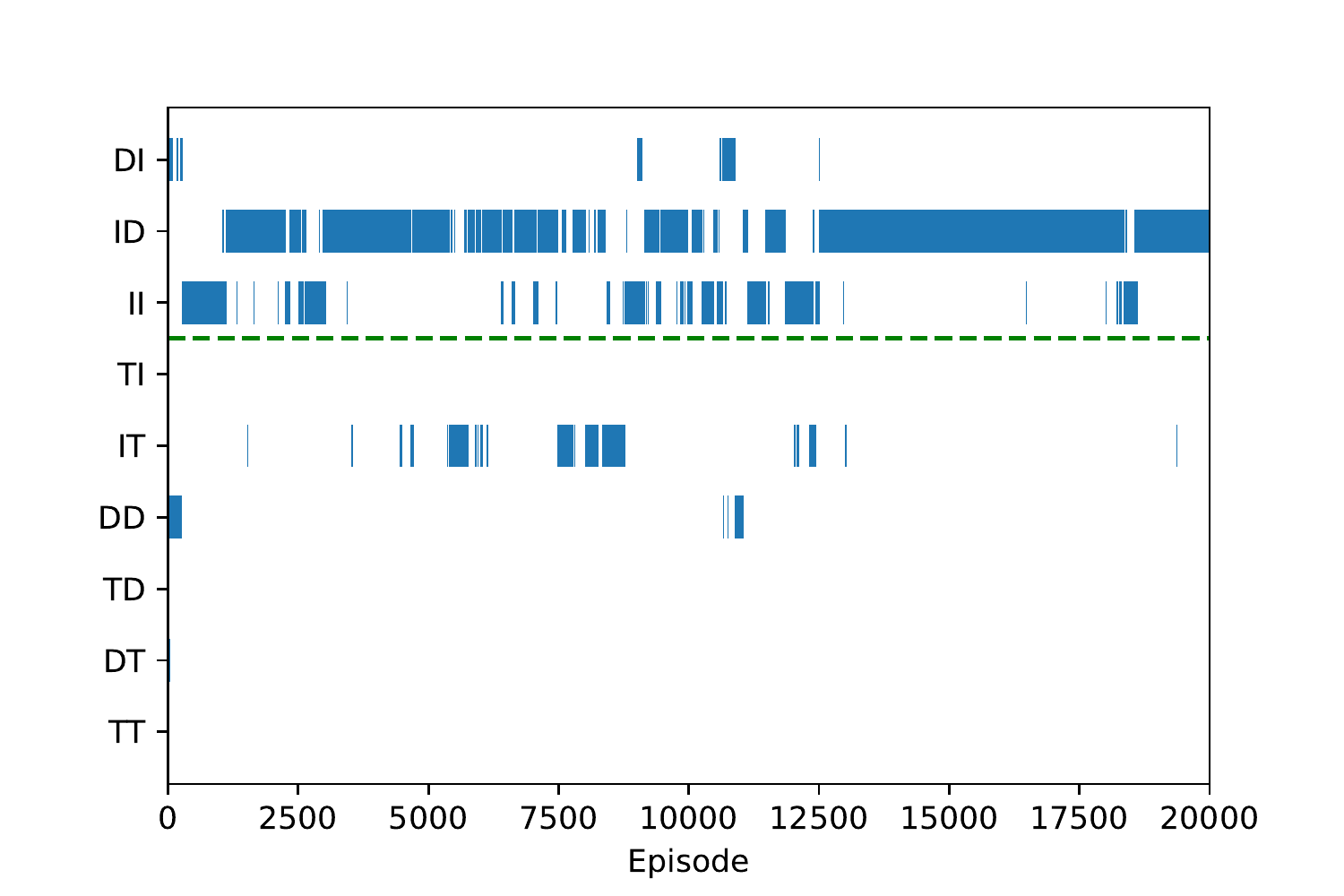}
         \caption{Policy ID}
         \label{fig:Baseline ID}
     \end{subfigure}
     \begin{subfigure}[h]{0.45\textwidth}
         \centering
         \includegraphics[width=\textwidth]{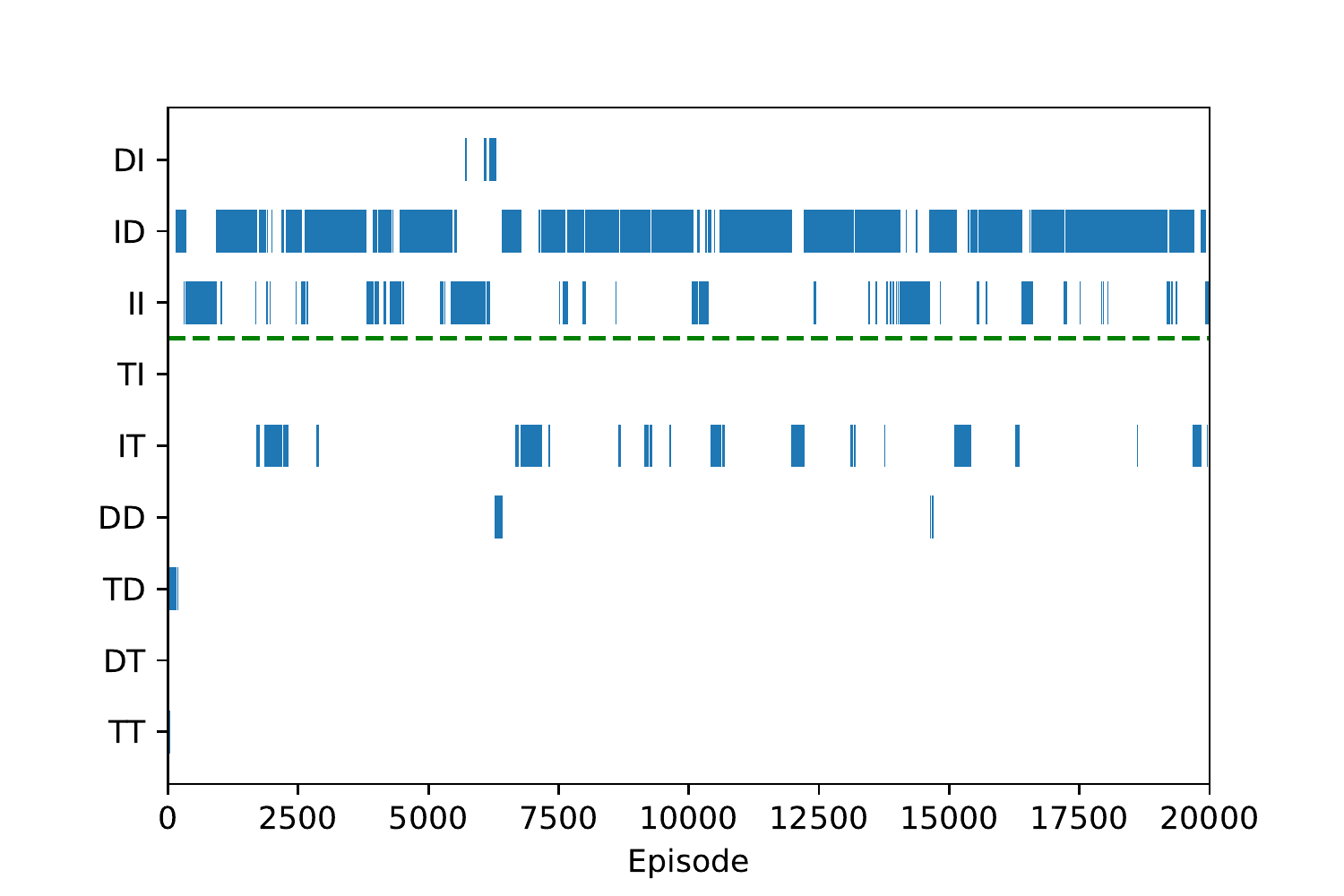}
         \caption{Policy II}
         \label{fig:Baseline II}
     \end{subfigure}
     \begin{subfigure}[h]{0.45\textwidth}
         \centering
         \includegraphics[width=\textwidth]{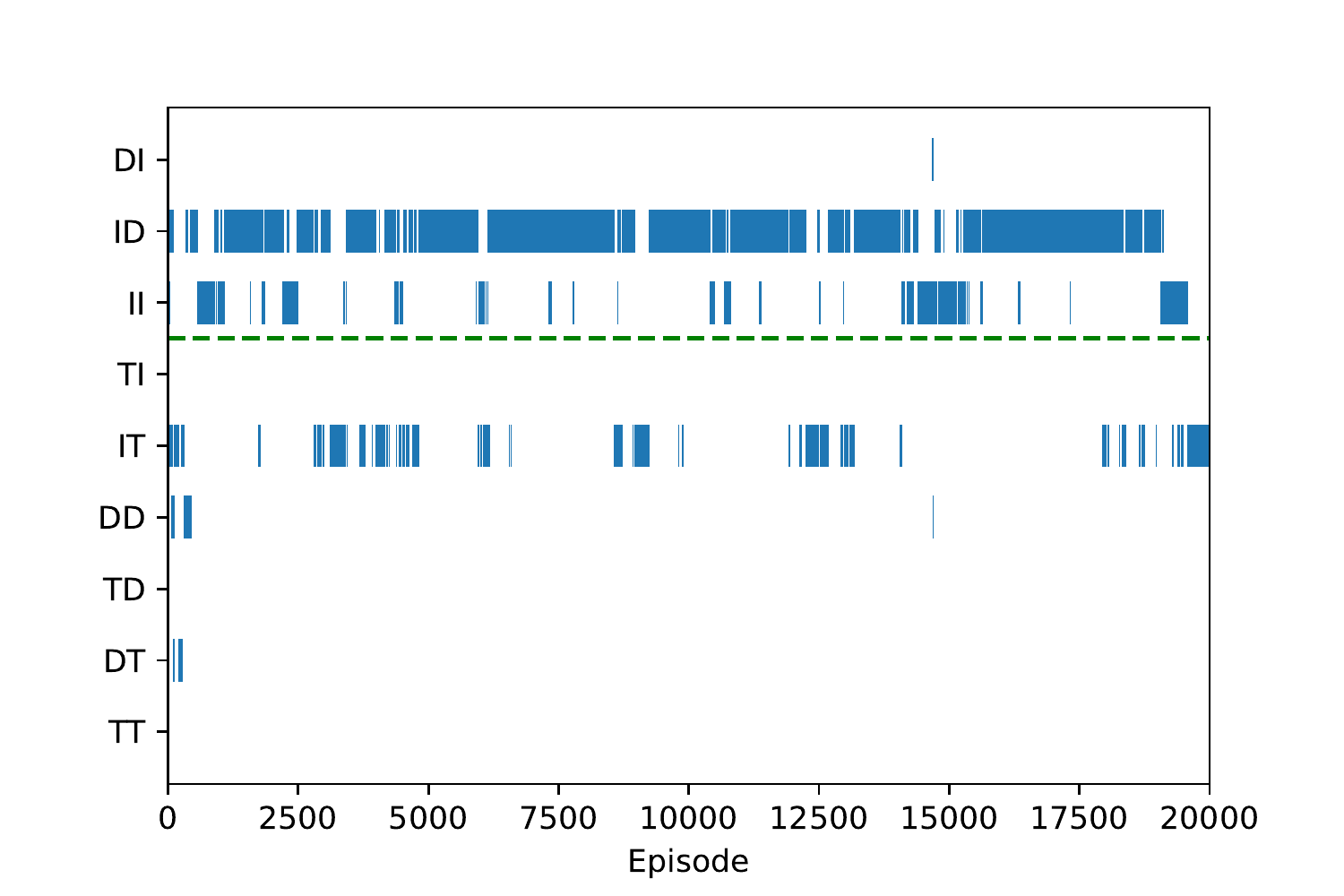}
         \caption{Policy IT}
         \label{fig:Baseline IT}
     \end{subfigure}
     \caption[Policy charts for Baseline method]{4 possible greedy policies produced by Baseline method}
     \label{fig:Baseline Policy Charts}
\end{figure}

Table \ref{tab:Space-Traders-Baseline} lists all 20 independent training results which are reproduced based on \shortciteA{VamplewEnvironmental2022} paper. This is used as benchmark to cross compare between each method later in this thesis.\newline\newline
\begin{table}[hbt!]
    \centering
    \begin{tabular}{@{}|c|c|c|c|c|@{}}
    \toprule
    Policy & DI & ID & II & IT  \\ \midrule
    Baseline & 1 & 13 & 4  & 2    \\ \bottomrule
    \end{tabular}%
    \caption[20 independent runs of the Algorithm \ref{algo:moql-expected}]{The final greedy policies learned in 20 independent runs of the Algorithm \ref{algo:moql-expected} for Space Traders environment}
    \label{tab:Space-Traders-Baseline}
\end{table}
Figure \ref{fig:Baseline Policy Charts} visualises the learning behaviour of the Baseline method (Algorithm \ref{algo:moql-expected}) in 20 independent trials. Each sub-part of the figure illustrates a single run of the baseline algorithm - an example was randomly chosen for each of the four different final policies listed in Table \ref{tab:Space-Traders-Baseline}. The graph shows for each episode, the policy which the agent believed to be optimal at that stage of its learning. The green dash line indicate the threshold for first objective. Above the threshold line, only DI, ID and II policy meet the constraint and the blue bar denotes for each episode which policy the agent believed to be optimal. As we can see from all of these policy charts, the agent's behaviour is unstable with frequent changes in its choice of optimal policy. Policy ID is the most frequently selected across 20,000 episodes, which reflects why it is the most frequent final outcome, but in many runs the agent winds up with a different final policy. In particular it can be seen that policies beneath the threshold are regarded as optimal on an intermittent basis, which indicates that the agent's estimate of the value of these policies must be inaccurate.
\section{Discussion}
The empirical results from 20 trials show that the desired optimal policy (DI) was not converged to in practice, with it being identified as the best policy in only 1 of 20 runs. A closer examination of agent's behaviours reveals that regardless of which action agent selected at state A, if state B is successfully reached, then a zero reward will have been received by the agent for the first objective. In another words, the accumulated expected reward for first objective at state B is zero. Therefore, the choice of action at state B is
purely based on state action value. Now looking at the mean action values for state B which is reported in Table \ref{tab:Space-Traders-O1} from previous chapter. It can be seen that teleport action will be eliminated because it fails to meet the threshold for the first objective, and the direct action will be preferred over indirect action as both meet the threshold, and Direct action takes less time penalty in second objective. Therefore, agent will choose direct action at state B regardless of which action agent selected at state A. As the result, this agent at state A will only consider Policy ID, DD and TD and since only policy ID is above the threshold for first objective if we look back the mean reward in table \ref{tab:Space-Traders-O2}. Therefore agent converges to sub-optimal policy ID in practice. In addition the issue of noisy estimates means that the agent will sometimes settle on another policy, including a policy which does not even meet the success threshold.
\chapter{Designing Reward Signals}
\label{chapter5}
The first approach we consider is to modify the reward structure of original Space Traders. As we discussed early in chapter 2, in practice, designing a reward function is often a trial-and-error process. Sometimes with a smart reward design, agent could perform much better compared with a bad reward signal. Therefore, the most simple and natural approach is to modify the reward structure first without actually changing the original MOQ-learning algorithm. The new reward design will be discussed in next section.
\section{New reward design}
\begin{figure}[hbt!]
    \centering
    \includegraphics[width=15cm]{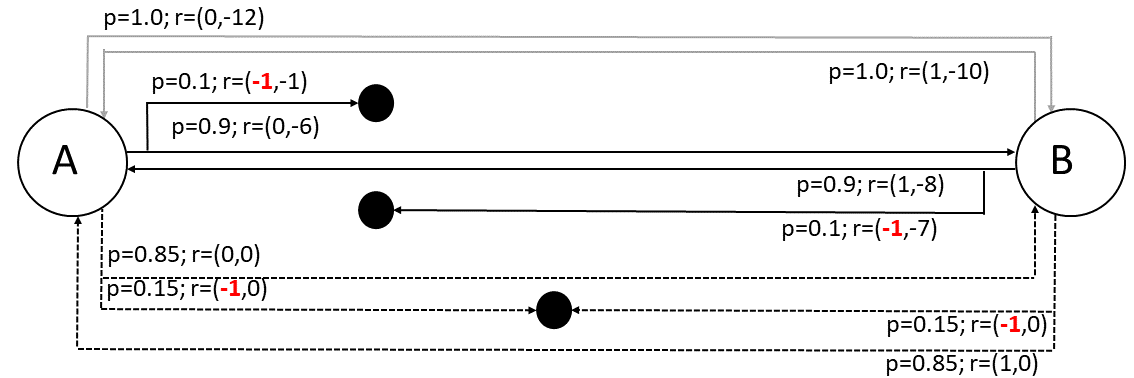}
    \caption[The reward design version of Space Traders Environment]
    {The reward design version of Space Traders MOMDP. All the changes have been highlighted in red color}
    \label{fig:ST-V2}
\end{figure}

New reward design version of Space Traders is shown in Figure \ref{fig:ST-V2}. Every time, agent will receive a -1 reward for the first objective when visiting one of the terminal state, receives +1 when reaching the goal state, and 0 for other intermediate transitions. The motivation here is to avoid the situation arising from the original reward structure as discussed in the previous chapter, where the accumulated expected reward for the first objective when reaching state B is always zero, regardless of the action selected in state A. As can be seen from the top-half of Table 5.1, under our new reward design, the 3 actions from state A have differing expected values of 0, -0.1, and -0.15. \newline\newline
As a consequence, the threshold value for first element also need to be updated. Because the total rewards for first element are now ranging for -1 to 1 instead of 0 to 1. So the equivalent threshold value 0.88 will become to $0.88*1+0.12*(-1)=0.76$.
\begin{table}[hbt!]
    \centering
    \resizebox{\columnwidth}{!}{%
    \begin{tabular}{@{}|c|c|c|c|c|c|@{}}
    \toprule
    State              & Action   & P(Success) & Reward on Success & Reward on Failure & Mean Reward \\ \midrule
    \multirow{3}{*}{A} & Indirect & 1.0        & (0, -12)          & N/A               & (0, -12)    \\ \cmidrule(l){2-6} 
                       & Direct   & 0.9        & (0, -6)           & (-1, -1)           & (-0.1, -5.5)   \\ \cmidrule(l){2-6} 
                       & Teleport & 0.85       & (0, 0)            & (-1, 0)            & (-0.15, 0)      \\ \midrule
    \multirow{3}{*}{B} & Indirect & 1.0        & (1, -10)          & N/A               & (1, -10)    \\ \cmidrule(l){2-6} 
                       & Direct   & 0.9        & (1, -8)           & (-1, -7)           & (0.8, -7.9) \\ \cmidrule(l){2-6} 
                       & Teleport & 0.85       & (1, 0)            & (-1, 0)            & (0.7, 0)   \\ \bottomrule
    \end{tabular}%
    }
    \caption[The State Action pair in Reward Design Space Traders Environment]
    {The probability of success and reward values for each state-action pair in Reward Design Space Traders MOMDP}
    \label{tab:RewardDesign-Space-Traders-O1}
\end{table}
\section{Result}
\begin{figure}[hbt!]
     \centering
     \begin{subfigure}[h]{0.45\textwidth}
         \centering
         \includegraphics[width=\textwidth]{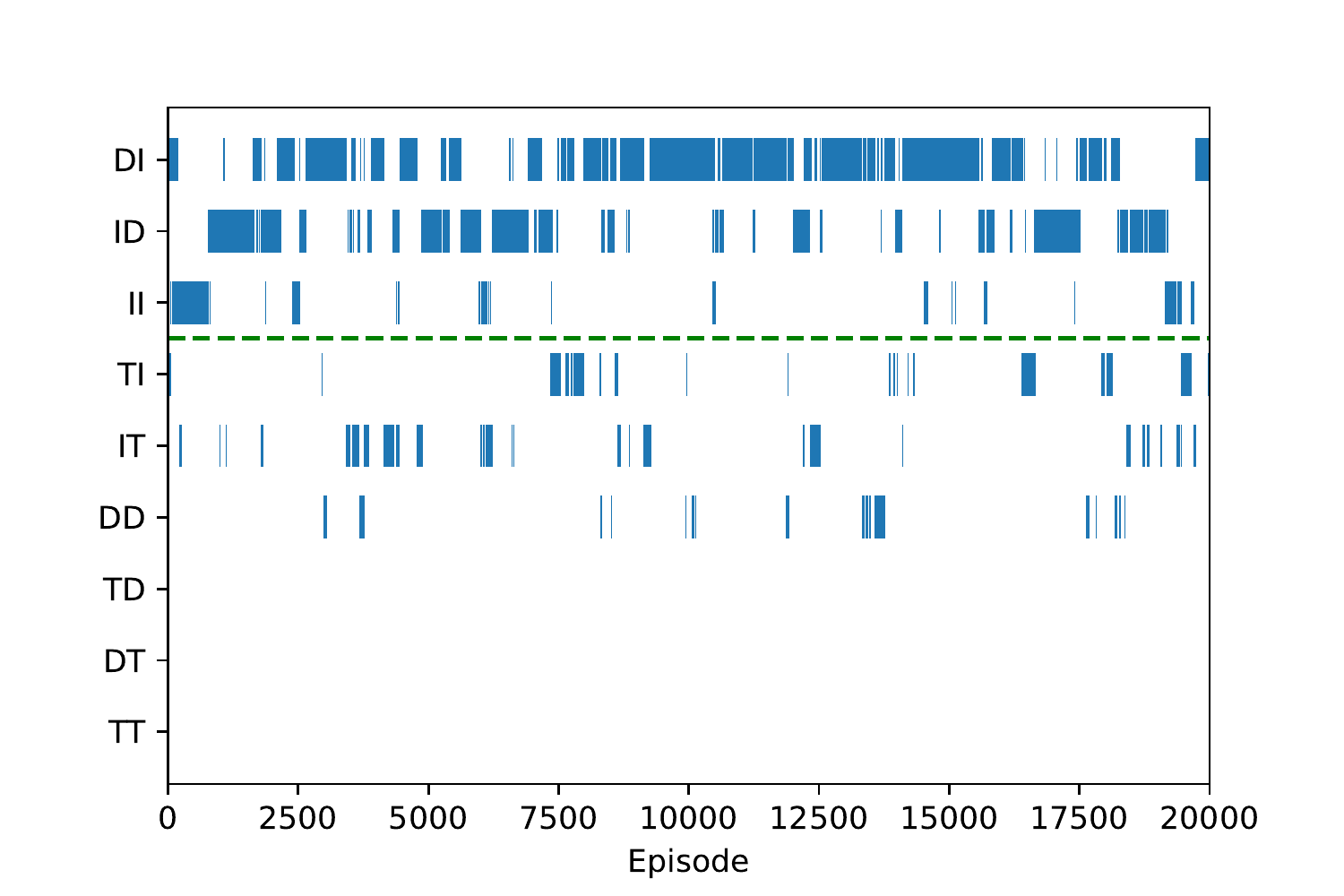}
         \caption{Policy DI}
         \label{fig:RewardDesign DI}
     \end{subfigure}
     \begin{subfigure}[h]{0.45\textwidth}
         \centering
         \includegraphics[width=\textwidth]{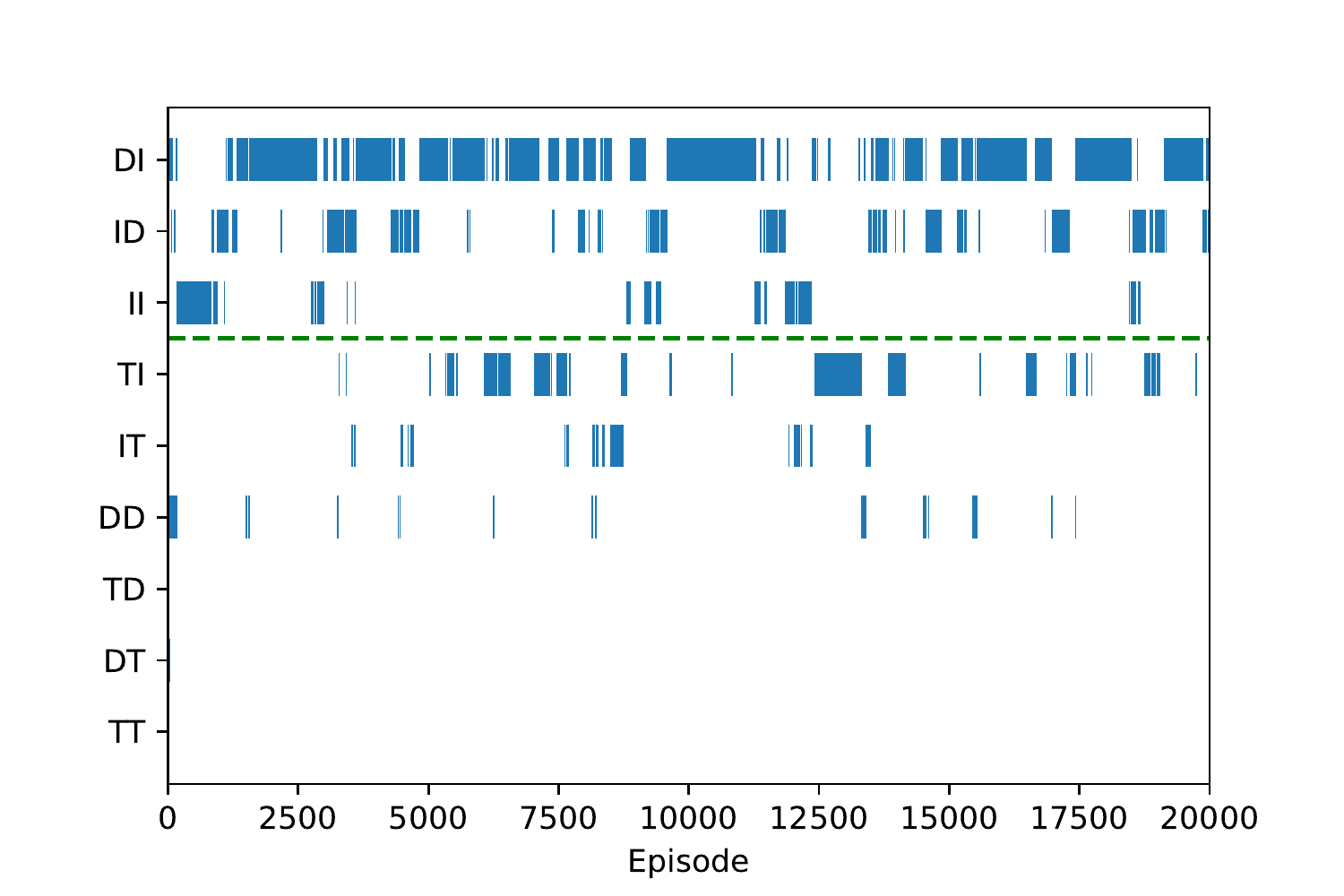}
         \caption{Policy ID}
         \label{fig:RewardDesign ID}
     \end{subfigure}
     \begin{subfigure}[h]{0.45\textwidth}
         \centering
         \includegraphics[width=\textwidth]{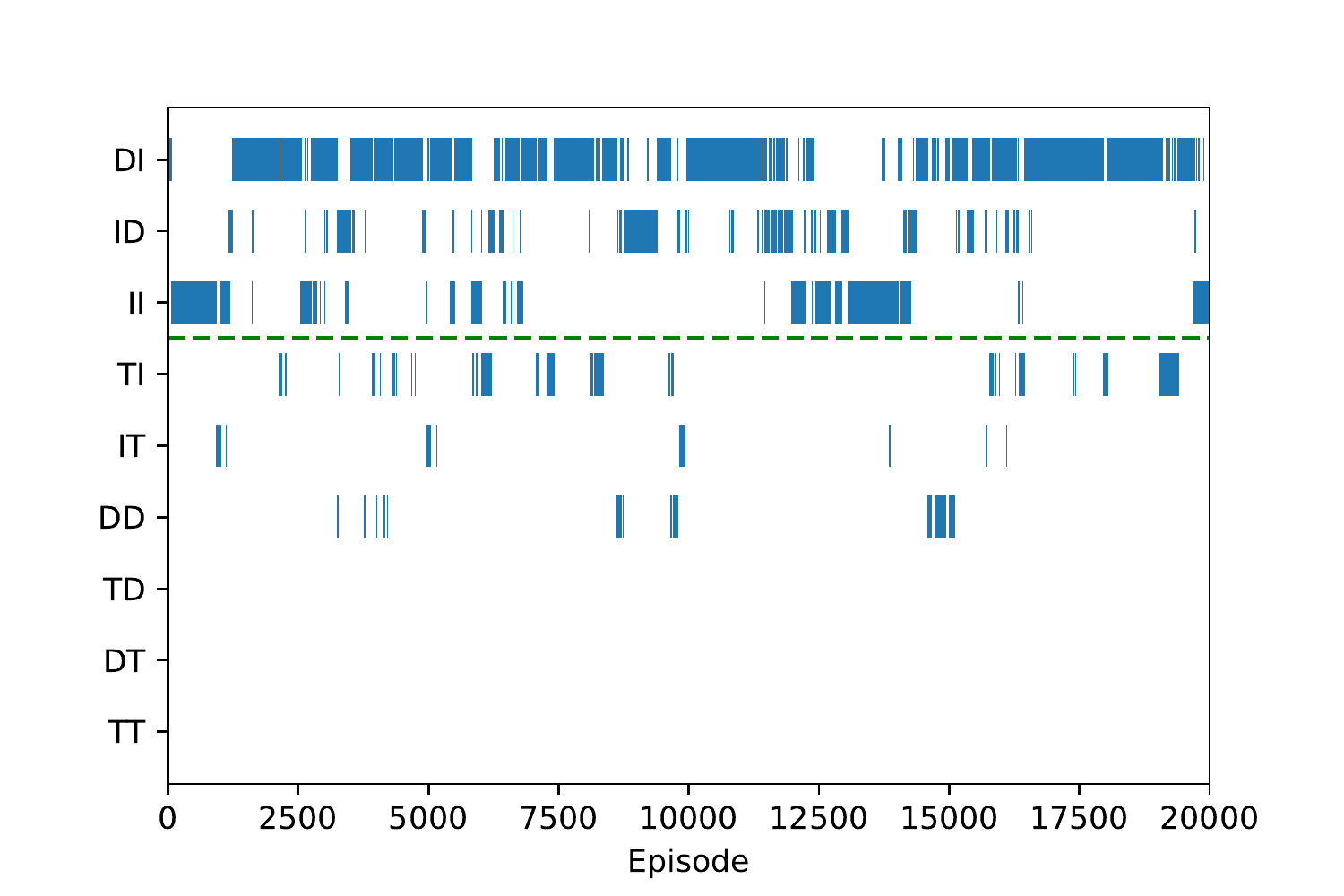}
         \caption{Policy II}
         \label{fig:RewardDesign II}
     \end{subfigure}
     \begin{subfigure}[h]{0.45\textwidth}
         \centering
         \includegraphics[width=\textwidth]{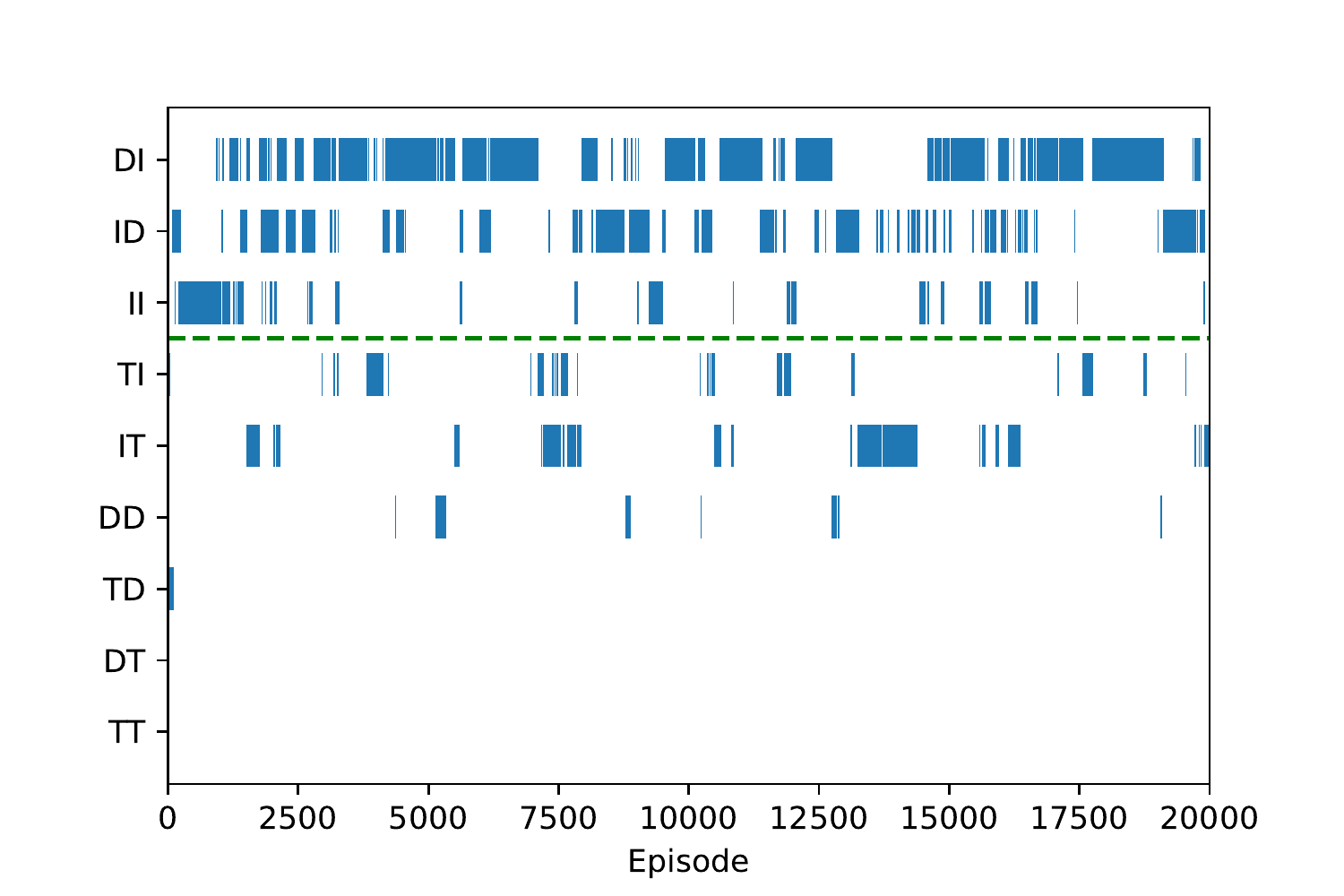}
         \caption{Policy IT}
         \label{fig:RewardDesign IT}
     \end{subfigure}
          \begin{subfigure}[h]{0.45\textwidth}
         \centering
         \includegraphics[width=\textwidth]{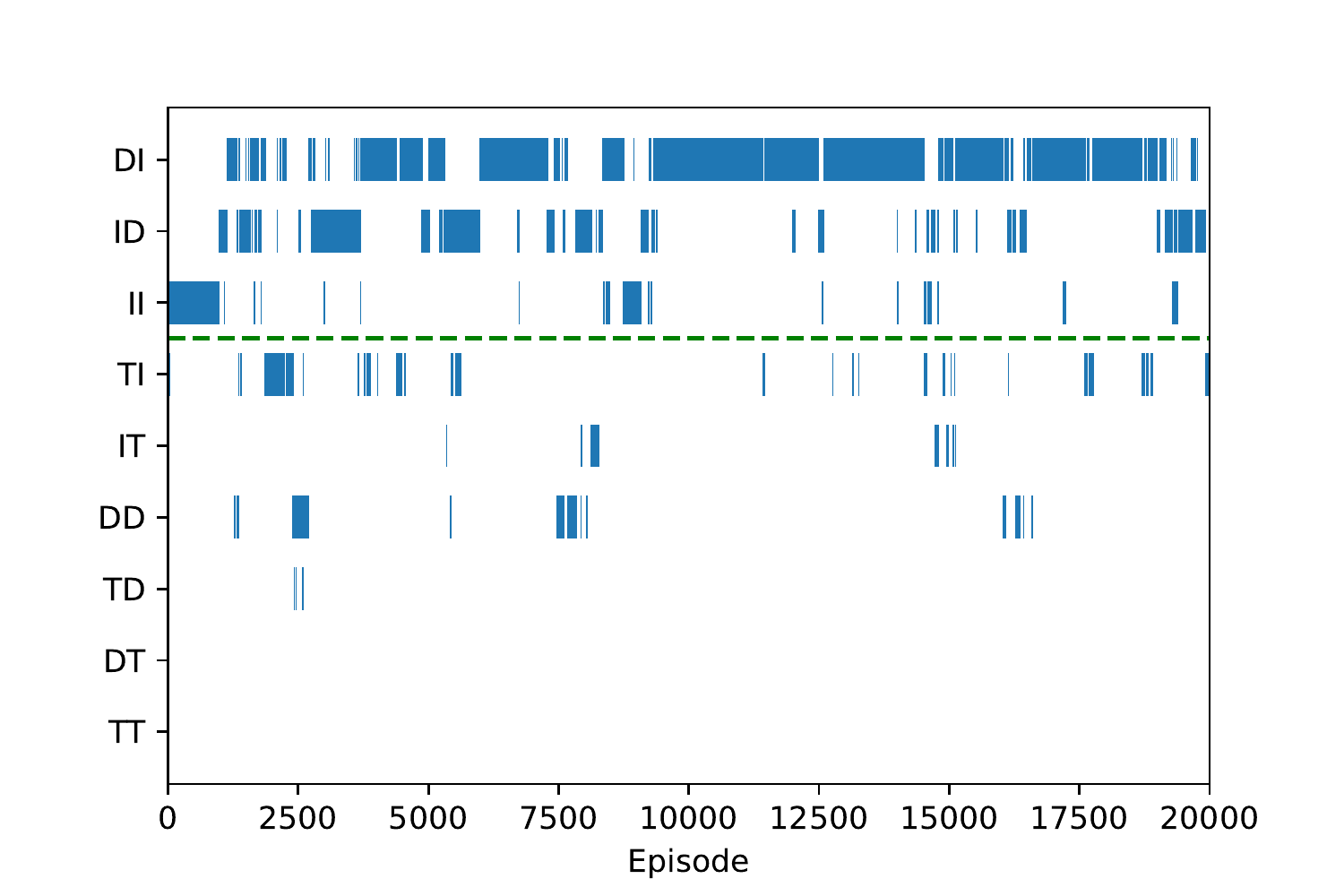}
         \caption{Policy TI}
         \label{fig:RewardDesign TI}
     \end{subfigure}
         \begin{subfigure}[h]{0.45\textwidth}
         \centering
         \includegraphics[width=\textwidth]{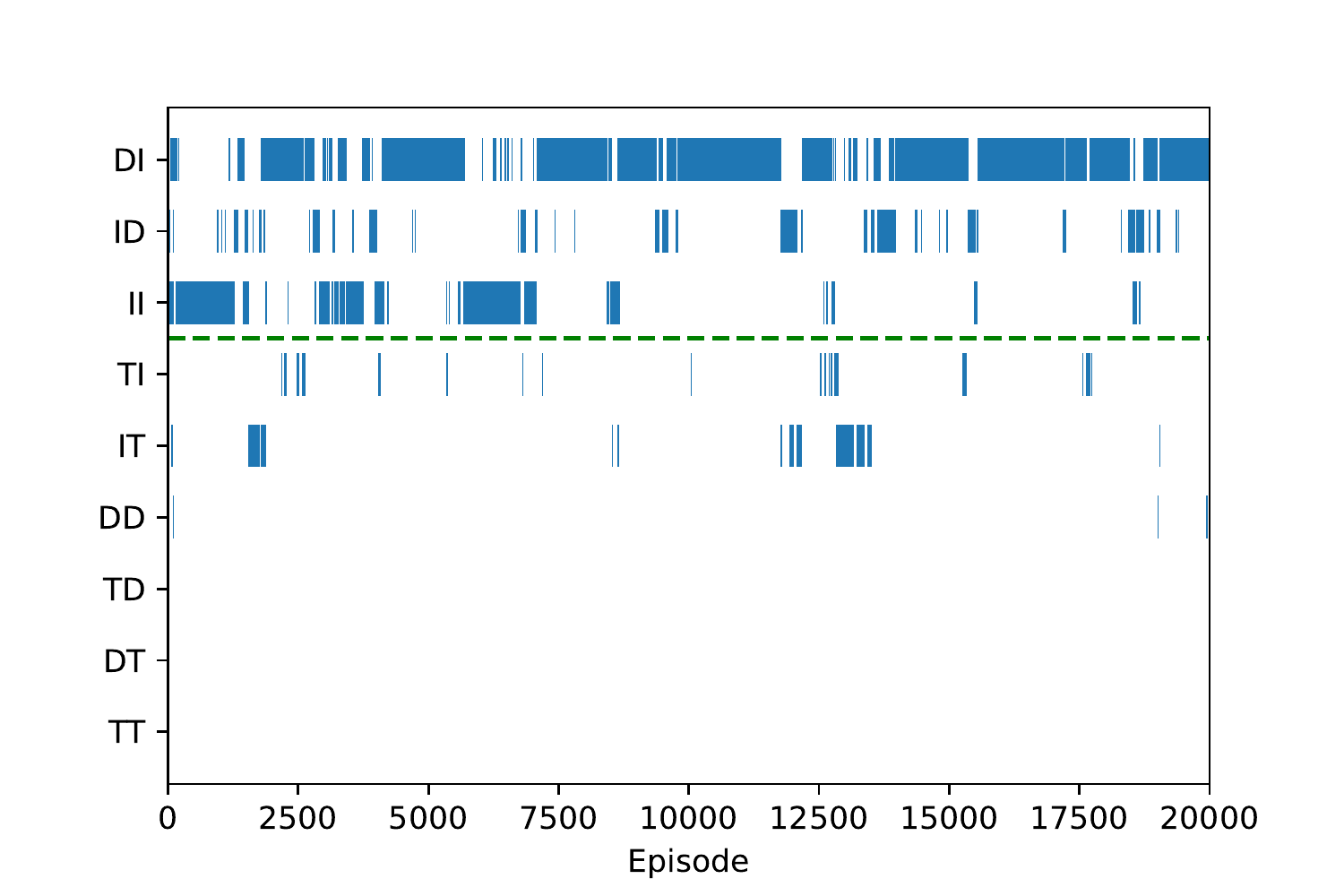}
         \caption{Policy DD}
         \label{fig:RewardDesign DD}
     \end{subfigure}
     \caption[Policy charts for Reward Design method]{6 policy charts}
     \label{fig:Reward Design Policy Charts}
\end{figure}
\begin{table}[hbt!]
    \centering
    \begin{tabular}{@{}|c|c|c|c|c|c|c|@{}}
    \toprule
    Policy & DI & ID & II & IT & TI & DD\\ \midrule
    Baseline & 1 & 13 & 4  & 2  & 0 & 0\\ \midrule
    Reward Design & 10 & 5 & 1 & 1 & 2 & 1\\ \bottomrule
    \end{tabular}%
    \caption[20 independent runs of the Algorithm \ref{algo:moql-expected}]{The final greedy policies learned in 20 independent runs of the Algorithm \ref{algo:moql-expected} for Space Traders environment}
    \label{tab:Space-Traders-RewardDesign}
\end{table}
As we can see from table \ref{tab:Space-Traders-RewardDesign}, the most common outcome (10/20 runs) is the desired DI policy. This is a substantial improvement over the single occurrence of this policy under the original reward design. But on the another hand, ID policy (5 repetitions) is the second common outcome, TI policy (2 repetitions) and II policy (1) also occur in some runs. From figure \ref{fig:Reward Design Policy Charts}, most of time across 20,000 episodes, agent stays on policy DI which is our desired optimal policy. However the intermittent identification of the other policies as optimal means that overall this approach still only yields the correct policy 50\% of the time.
\section{Discussion}
\begin{figure}[hbt!]
    \centering
    \includegraphics[width=15cm]{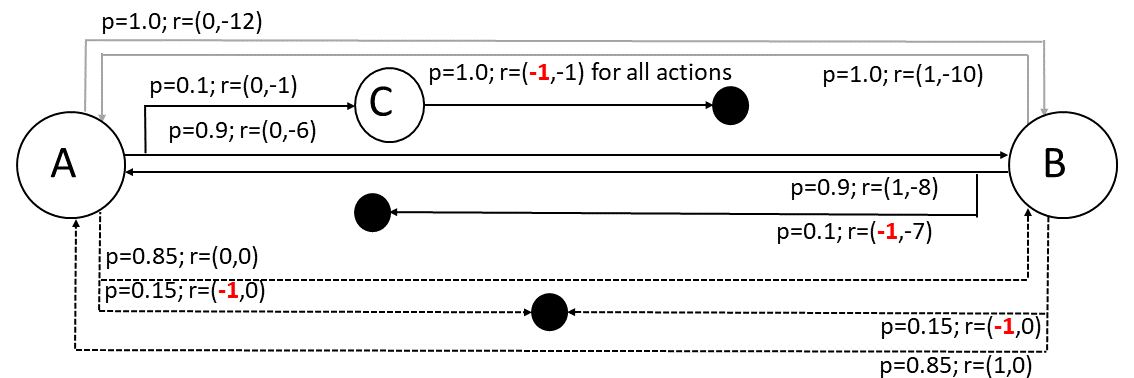}
    \caption[The new variant of reward design for Space Traders Environment]
    {The new variant of reward design for Space Traders MOMDP with a more complex state structure. All the changes have been highlight in red color}
    \label{fig:ST-V3}
\end{figure}
\begin{table}[hbt!]
    \centering
    \begin{tabular}{@{}|c|c|c|c|c|c|c|c|@{}}
    \toprule
    Policy & DI & ID & II & IT & TI & DD & TD\\ \midrule
    Reward Design & 10 & 5 & 1 & 1 & 2 & 1 & 0\\ \midrule
    Extra State & 0 & 14 & 2 & 1 & 2 & 0 & 1\\ \bottomrule
    \end{tabular}%
    \caption[20 independent runs of the Algorithm \ref{algo:moql-expected}]{The final greedy policies learned in 20 independent runs of the Algorithm \ref{algo:moql-expected} for new variant of reward design Space Traders environment}
    \label{tab:Space-Traders-RewardDesign-v}
\end{table}
In order to test whether this reward design strategy is going to work for general problem as well or it will still suffer the same problem as with the original Space Traders. The new Space Traders environment has been introduced as shown in Figure \ref{fig:ST-V3}. It includes a new state C when agent select direct action at state A. The empirical results from 20 trials show that the desired optimal policy (DI) was not converged to in practice for this new space traders environment. A closer examination of the behaviour of the agent in first reward design shows when agent selects different action in state B will have different accumulated expected reward for first objective. For example, if agent selects direct action and successfully reaches to state B. Then the ideal accumulated expected reward for first objective will be $0.9*0+0.1*(-1)=-0.1$ when the action values are learned with sufficient accuracy. This time agent will select indirect action in state B as combining with accumulated expected reward $-0.1$ both direct action and teleport action's utility value will below the threshold 0.76 in first objective which can be checked in table \ref{tab:RewardDesign-Space-Traders-O1}. But in the extra state Space Traders environment, the accumulated expected reward goes back to zero again when agent reach state by taking direct action. As the result, agent converges to sub-optimal policy ID in practice again.\newline\newline
What this example illustrates is that while it maybe possible in some cases to encourage SER-optimal behaviour via a careful designing of rewards, in other cases the structure of the environment may make it difficult or impossible to identify a suitable reward design.
\section{Conclusion}
The new reward design does improve the baseline method in original Space Traders problem. But for this particular environment structure it is possible to design the reward signal which essentially captures the required information such as the transition probabilities within the accumulated expected reward for the first objective. However more generally this may not be easy or even possible to achieve. Therefore, simply changing the reward signal is insufficient to address issues in stochastic environments under SER criteria. In addition, even with a suitable reward design, the baseline MOQ-learning algorithm may still fail to find out the SER-optimal policy occasionally. One potential solution is to add extra global information for building the augmented state and this will be discuss in next chapter.

\chapter{Single-Phase MOSS}
\label{chapter6}
As identified in previous study \shortcite{VamplewEnvironmental2022}, the main issue for applying MOQ-learning algorithm to stochastic environment is that the action selection at given state is purely based on local information (the Q value for current state) and current episode information (accumulated expected reward). This is the same issue previously identified for multi-objective planning algorithms by \shortciteA{bryceProbabilisticPlanningMultiobjective2007}. However, in order to maximise the expected utility over multiple episodes (SER criteria) the agent must also consider expected return on other episodes where current state is not reached as well. In another words, the agent must also have some level of knowledge about global statistics in order to maximise \acrfull{SER}. Therefore the second approach is to include extra global information for current MOQ-learning algorithm. 
\section{Algorithm}
To support this idea, Multi-objective Stochastic State Q-learning (MOSS)(Algorithm \ref{algo:mossql}) is introduced. Here are the changes compared with previous MOQ-learning (Algorithm \ref{algo:moql-expected})
\begin{itemize}
  \item The agent maintains two pieces of global information: the total number of episodes experienced ($v_\pi$), and an estimate of the average per-episode return ($E_\pi$).
  \item For every state, the agent maintains a counter of episodes in which this state was visited at least once ($v(s)$), and the estimated average return in those episodes ($E(s)$).
  \item When selecting an action, the agent uses those values to estimate the average return in episodes where the current state is not visited. This value is then combined with estimated accumulated rewards $P(s)$ and Q value $Q(s)$ to estimate the return for each action, which taking in to account all episodes (both the episodes in which this state is visited, and those in which it is not visited). Action selection is then based on this holistic measure of the value for each action, which should make the action selection more compatible with the goal of finding the SER-optimal policy.
\end{itemize}
\begin{algorithm}[hbt!]
  \caption{The multi-objective stochastic state Q($\lambda$) algorithm (MOSSQ-learning). Highlighted text identifies the changes and extensions introduced relative to multi-objective Q($\lambda$) as previously described in Algorithm \ref{algo:moql-expected}}
  \label{algo:mossql}
  \begin{algorithmic}[1]
    \Statex input: learning rate $\alpha$, discounting term $\gamma$, eligibility trace decay term $\lambda$, number of objectives $n$, action-selection function $f$ and any associated parameters
    \For {all states $s$, actions $a$ and objectives $o$}
    	\State initialise $Q_o(s,a)$
    	\textcolor{red}{
    	    \State initialise $P_o(s)$ \Comment{expected cumulative reward when $s$ is reached}
    	    \State initialise $v(s)=0$ \Comment{count of visits to s}
    	}
    \EndFor
    \State \textcolor{red}{initialise $E_\pi$ \Comment{estimated return over all episodes}
    }
    \State \textcolor{red}{initialise $v_\pi=0$ \Comment{count of all episodes}}
   \For {each episode} 
            \State \textcolor{red}{$v_\pi = v_\pi + 1$
            \Comment{increment episode counter}}
        \For {all states $s$ and actions $a$}
    		\State $e(s,a)$=0; 
    		\textcolor{red}{ $b(s)=0$ \Comment{binary flag - was $s$ visited in this episode?}
        }
        \EndFor
        \State sums of prior rewards $P_o$ = 0, for all $o$ in 1..$n$
    	\State observe initial state $s_t$
    	\textcolor{red}{
    	    \\ \Comment{call helper algorithm to update stats and create augmented state and utility vector}
    	    \State $s^A_t$, $U(s^A_t)$ = update-statistics($s_t$,$P$)
            \State select $a_t$ from an exploratory policy derived using $f(U(s^A_t))$
        }\strut
        \For {each step of the episode}
 			\State execute $a_t$, observe $s_{t+1}$ and reward $R_t$
 			\State $P = P + R_t$
    	    \textcolor{red}{
    	        \State $s^A_{t+1}$, $U(s^A_{t+1})$ = update-statistics($s_{t+1}$,$P$)
            } 			
 			\State select $a^*$ from a greedy policy derived using  $f(U(s^A_{t+1}))$
  			\State select $a^\prime$ from an exploratory policy derived using $f(U(s^A_{t+1}))$
            \State $\delta = R_t + \gamma Q(s^A_{t+1},a^*) - Q(s^A_t,a_t)$
            \State $e(s^A_t,a_t)$ = 1
            \For {each augmented state $s^A$ and action $a$}
            	\State $Q(s^A,a) = Q(s^A,a) + \alpha\delta e(s^A,a)$
                 \If {$a^\prime = a^*$} 
                    \State $e(s^A,a) = \gamma \lambda e(s,a)$
                 \Else
                    \State $e(s^A,a) = 0$                 
                 \EndIf
            \EndFor
            \State $s^A_t = s^A_{t+1}, a_t = a^\prime$
    	\EndFor
    	\textcolor{red}{
    	    \State $E_\pi = E_\pi + \alpha(P-E_\pi)$ \Comment update estimates of per-episode return
            \For {all states with $b(s)\neq0$}
              \State $E(s) = E(s) + \alpha(P-E(s))$
            \EndFor
        }    	
    \EndFor
  \end{algorithmic}
\end{algorithm}
And the update-statistics helper algorithm can be found in Algorithm \ref{algo:update-stats1} which calculates the augmented state $s^A$ and the utility vector value $U$ for action selection in MOSSQ-learning (Algorithm \ref{algo:mossql}).\newline
\begin{algorithm}[hbt!]
  \caption{The update-statistics helper algorithm for MOSSQ-learning (Algorithm \ref{algo:mossql}). Given a particular state $s$ it updates the global variables which store statistics related to $s$. It will then return an augmented state formed from the concatenation of $s$ with the estimated mean accumulated reward when $s$ is reached, and a utility vector $U$ which estimates the mean vector return over all episodes for each action available in $s$}
  \label{algo:update-stats1}
  \begin{algorithmic}[1]
    \Statex input: state $s$, accumulated rewards in the current episode $P$
    \If {$b(s)=0$} \Comment{first visit to $s$ in this episode}
        \State $v(s) = v(s) + 1$ \Comment{increment count of visits to $s$}
        \State $b(s)=1$ \Comment{set flag so duplicate visits within an episode are not counted}
    \EndIf
    \State $P(s) = P(s) + \alpha (P-P(s))$
    \State $s^A$ = $(s,P(s))$ \Comment{augmented state}
    \State $p(s) = v(s)/v_\pi$    \Comment{estimated probability of visiting $s$ in any episode}
    \If {p(s)=1} \Comment{treat states which are always visited as a special case}
        \For {each action $a$}
            \State $U(a) = P(s)+Q(s^A,a))$ 
        \EndFor
    \Else
        \State $E_{\not{s}}=(E_\pi - p(s)E_s)/(1-p(s)$ \Comment estimated return in episodes where $s$ is not visited
        \State \Comment{calculate estimated value over all episodes, assuming a is executed in $s^A$}
        \For {each action $a$}
            \State $U(a) = p(s)(P(s)+Q(s^A,a)) + (1-p(s))E_{\not{s}}$ 
        \EndFor
    \EndIf
    \State return $s^A$, $U$
  \end{algorithmic}
\end{algorithm}
\section{Result}
\begin{figure}[hbt!]
     \centering
     \begin{subfigure}[h]{0.45\textwidth}
         \centering
         \includegraphics[width=\textwidth]{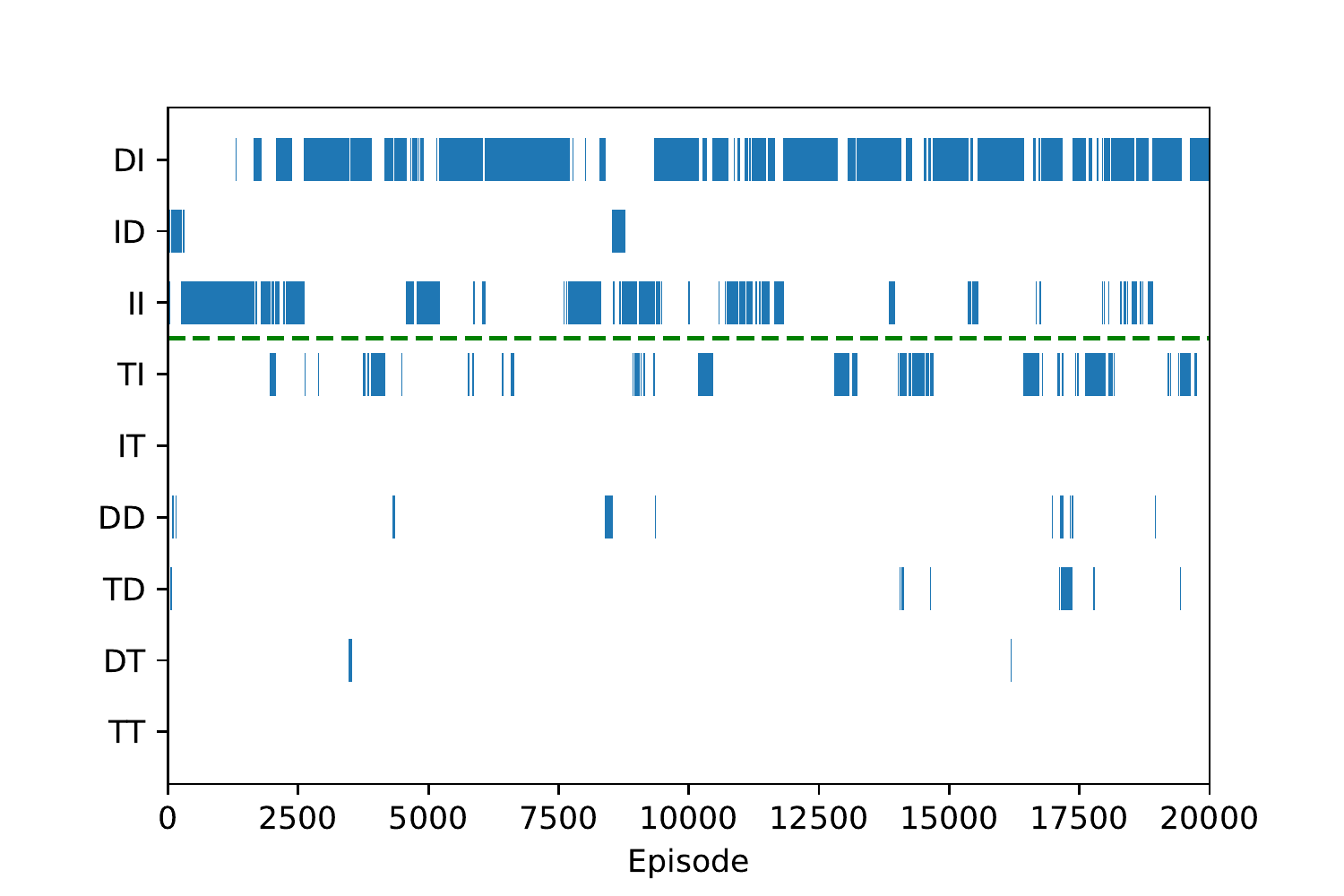}
         \caption{Policy DI}
         \label{fig:MOSS DI}
     \end{subfigure}
     \begin{subfigure}[h]{0.45\textwidth}
         \centering
         \includegraphics[width=\textwidth]{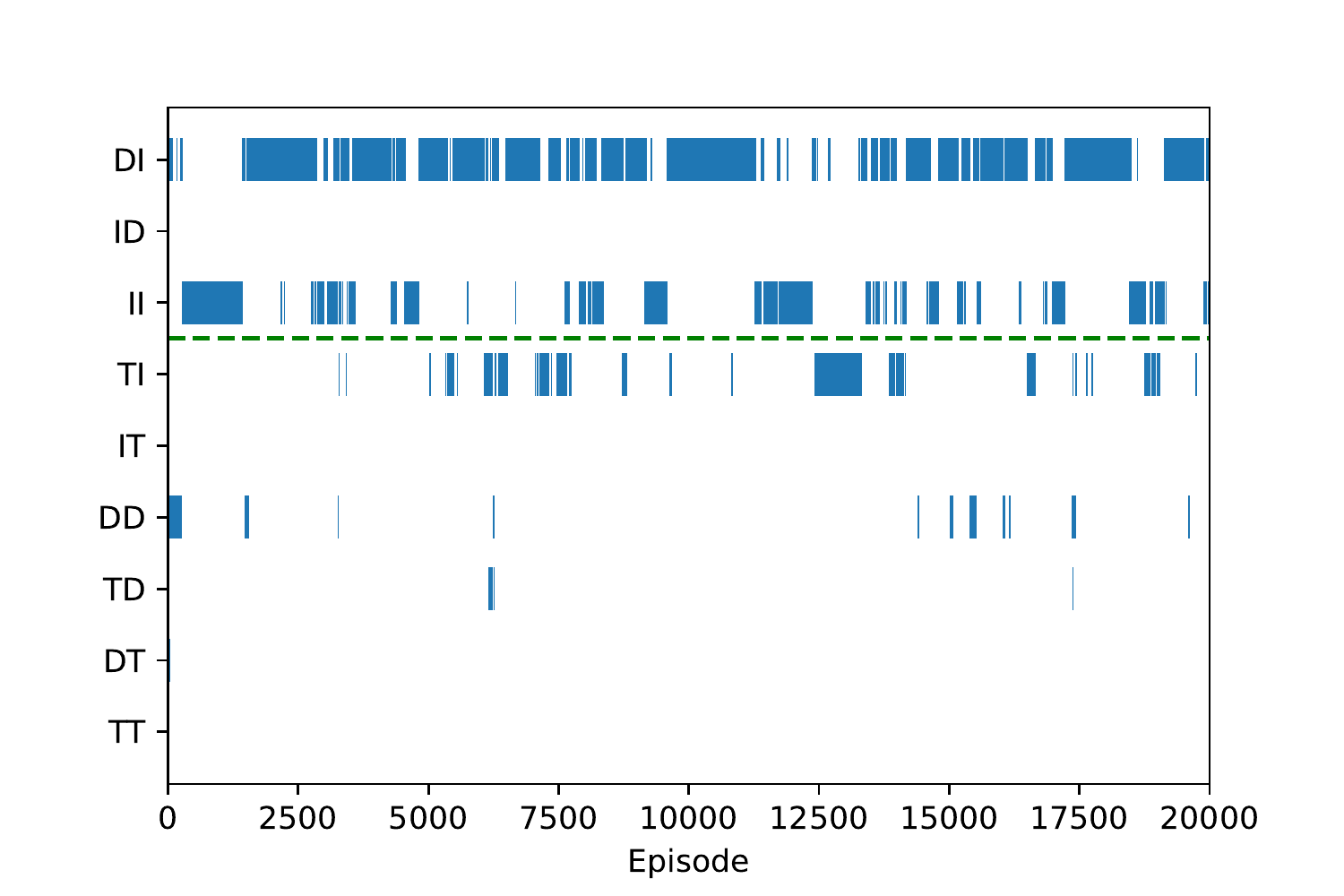}
         \caption{Policy II}
         \label{fig:MOSS II}
     \end{subfigure}
     \begin{subfigure}[h]{0.45\textwidth}
         \centering
         \includegraphics[width=\textwidth]{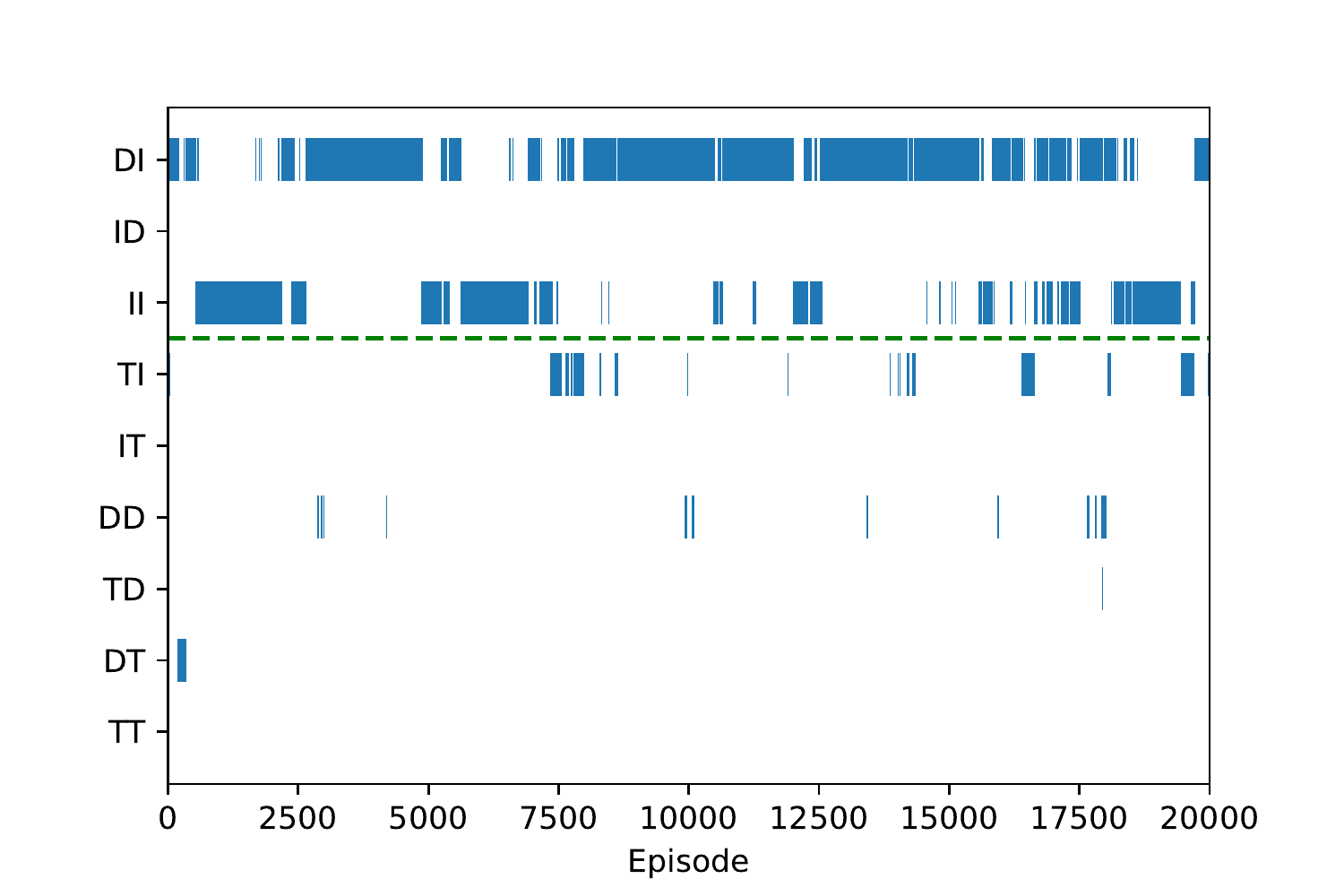}
         \caption{Policy TI}
         \label{fig:MOSS TI}
     \end{subfigure}
     \caption[Policy charts for Single-Phase MOSS]{3 policy charts for Single-Phase MOSS Algorithm}
     \label{fig:MOSS Policy Charts}
\end{figure}
\begin{table}[hbt!]
    \centering
    \begin{tabular}{@{}|c|c|c|c|c|c|c|@{}}
    \toprule
    Policy & DI & ID & II & IT & TI & DD\\ \midrule
    Baseline & 1 & 13 & 4  & 2  & 0 & 0\\ \midrule
    Reward Design & 10 & 5 & 1 & 1 & 2 & 1\\ \midrule
    MOSS & 15 & 0 & 0 & 3 & 2 & 0\\ \bottomrule
    \end{tabular}%
    \caption[20 independent runs of the Algorithm \ref{algo:mossql}]{The final greedy policies learned in 20 independent runs of the Single-Phase MOSS algorithm for Space Traders environment}
    \label{tab:Space-Traders-MOSS}
\end{table}
As we can see from table \ref{tab:Space-Traders-MOSS}, the most common result (15/20 runs)
is the DI policy, which is the desired optimal policy, but the IT policy (3 repetitions) and TI policy (2) also occur in some trials. Figure \ref{fig:MOSS Policy Charts} also indicates that even the final policy converge to Policy II and Policy TI at the end of training. Most of time, agent believes policy DI is the desired optimal policy. But surprisingly, ID policy becomes less popular compared with policy ID and II.
\section{Discussion}
\begin{figure}[hbt!]
    \centering
    \includegraphics[width=15cm]{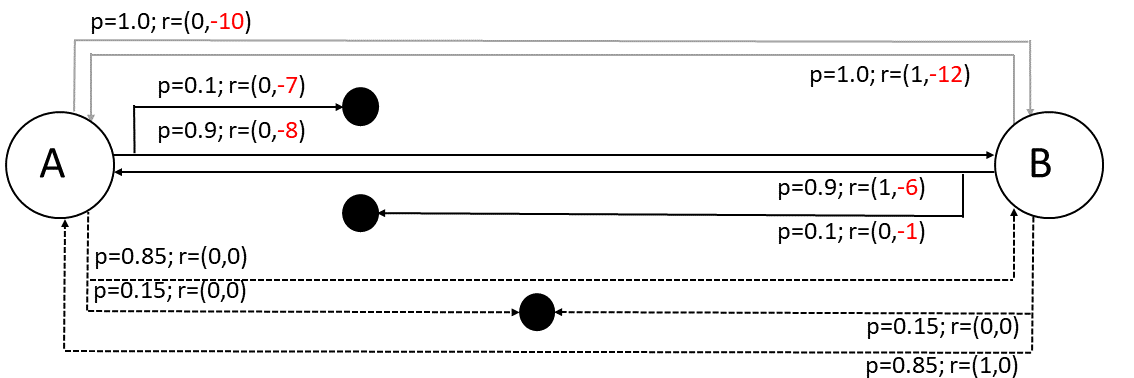}
    \caption[New variant of Space Traders Environment]
    {New variant of Space Traders MOMDP for testing MOSS Algorithm \ref{algo:mossql}. All the changes compared with original have been highlight in red color}
    \label{fig:ST-V1}
\end{figure}

Compared with benchmark result in table \ref{tab:Space-Traders-MOSS}, Single Phase MOSS algorithm clearly outperform the baseline method in original Space Traders problem. But in order to test whether this MOSS algorithm is going to work for more general problem as well. The new variant of Space Traders Problem has been introduced as shown in Figure \ref{fig:ST-V1}.
\subsection{New variant of Space Traders}
\begin{table}[hbt!]
    \centering
    \resizebox{\columnwidth}{!}{%
    \begin{tabular}{@{}|c|c|c|c|c|c|@{}}
    \toprule
    State              & Action   & P(Success) & Reward on Success & Reward on Failure & Mean Reward \\ \midrule
    \multirow{3}{*}{A} & Indirect & 1.0        & (0, -10)          & N/A               & (0, -10)    \\ \cmidrule(l){2-6} 
                       & Direct   & 0.9        & (0, -8)           & (0, -7)           & (0, -7.9)   \\ \cmidrule(l){2-6} 
                       & Teleport & 0.85       & (0, 0)            & (0, 0)            & (0, 0)      \\ \midrule
    \multirow{3}{*}{B} & Indirect & 1.0        & (1, -12)          & N/A               & (1, -12)    \\ \cmidrule(l){2-6} 
                       & Direct   & 0.9        & (1, -6)           & (0, -1)           & (0.9, -5.5) \\ \cmidrule(l){2-6} 
                       & Teleport & 0.85       & (1, 0)            & (0, 0)            & (0.85, 0)   \\ \bottomrule
    \end{tabular}%
    }
    \caption[The State Action pair in the new variant Space Traders Environment]
    {The probability of success and reward values for each state-action pair in the new variant Space Traders MOMDP}
    \label{tab:Space-Traders-MOSS-O1}
\end{table}
\begin{table}[hbt!]
    \centering
    \begin{tabular}{@{}|c|c|c|c|@{}}
    \toprule
    Policy identifier & Action in state A & Action in state B & Mean Reward     \\ \midrule
    II                & Indirect          & Indirect          & (1, -22)        \\ \midrule
    ID                & Indirect          & Direct            & (0.9, -15.5)    \\ \midrule
    IT                & Indirect          & Teleport          & (0.85, -10)     \\ \midrule
    DI                & Direct            & Indirect          & (0.9, -18.7)    \\ \midrule
    DD                & Direct            & Direct            & (0.81, -12.85)  \\ \midrule
    DT                & Direct            & Teleport          & (0.765, -7.9)   \\ \midrule
    TI                & Teleport          & Indirect          & (0.85, -10.2)    \\ \midrule
    TD                & Teleport          & Direct            & (0.765, -4.675) \\ \midrule
    TT                & Teleport          & Teleport          & (0.7225, 0)     \\ \bottomrule
    \end{tabular}
    \caption[Mean return for nine deterministic policies in new variant Space Traders Environment]{Nine available deterministic policies mean return for the new variant Space Traders Environment}
    \label{tab:Space-Traders-MOSS-O2}
\end{table}
\begin{figure}[hbt!]
     \centering
     \begin{subfigure}[h]{0.45\textwidth}
         \centering
         \includegraphics[width=\textwidth]{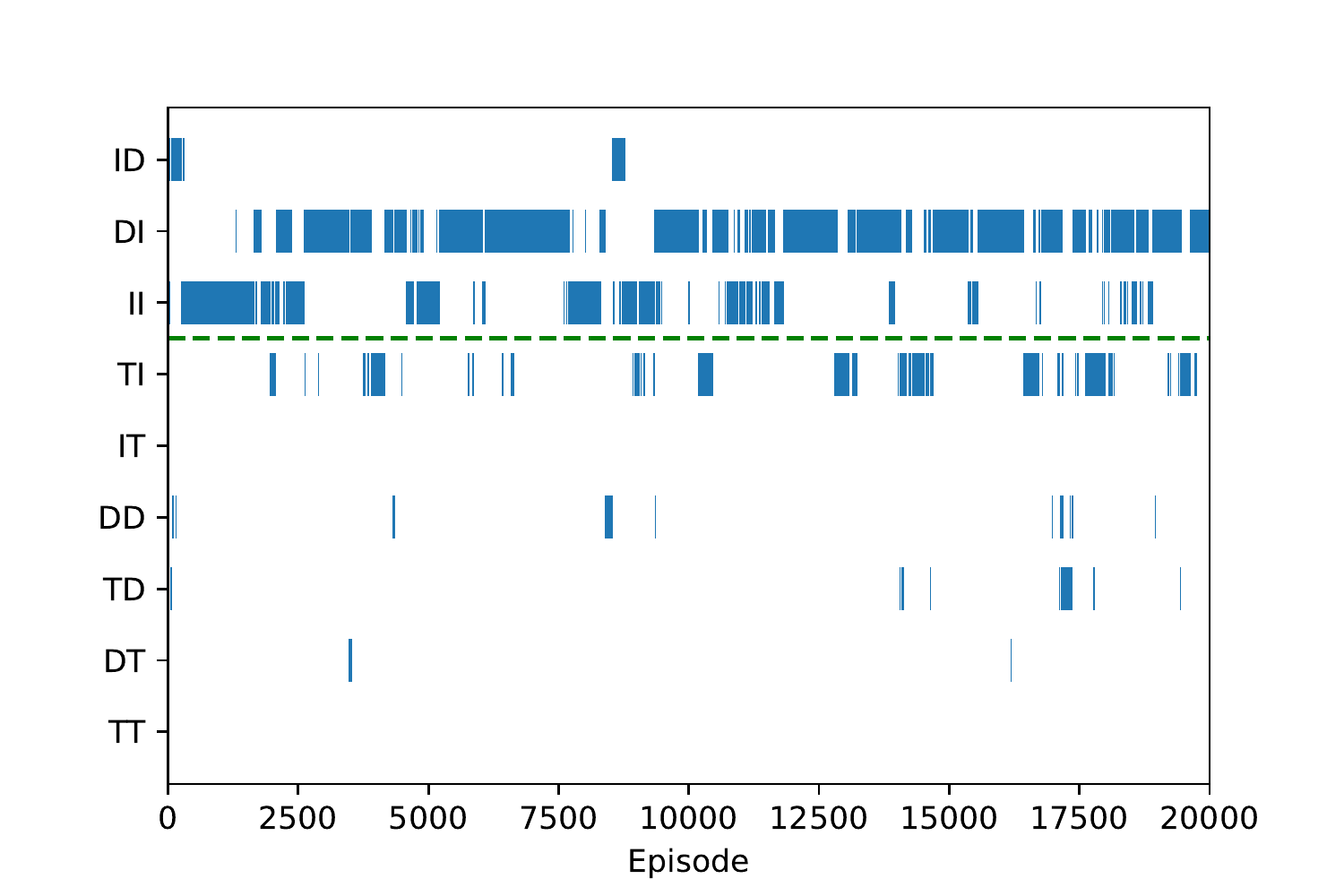}
         \caption{Policy DI}
         \label{fig:MOSS-v DI}
     \end{subfigure}
     \begin{subfigure}[h]{0.45\textwidth}
         \centering
         \includegraphics[width=\textwidth]{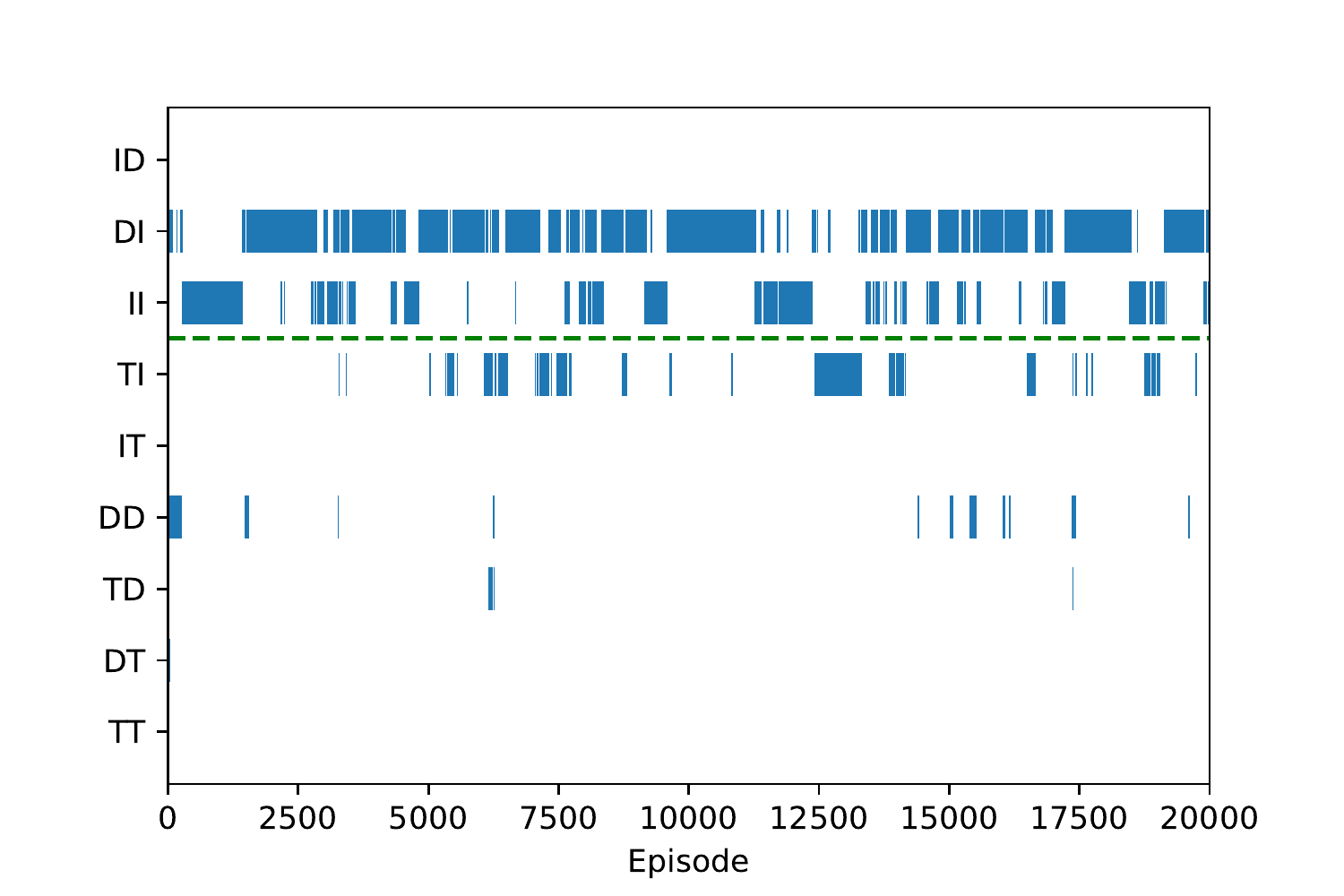}
         \caption{Policy II}
         \label{fig:MOSS-v II}
     \end{subfigure}
     \begin{subfigure}[h]{0.45\textwidth}
         \centering
         \includegraphics[width=\textwidth]{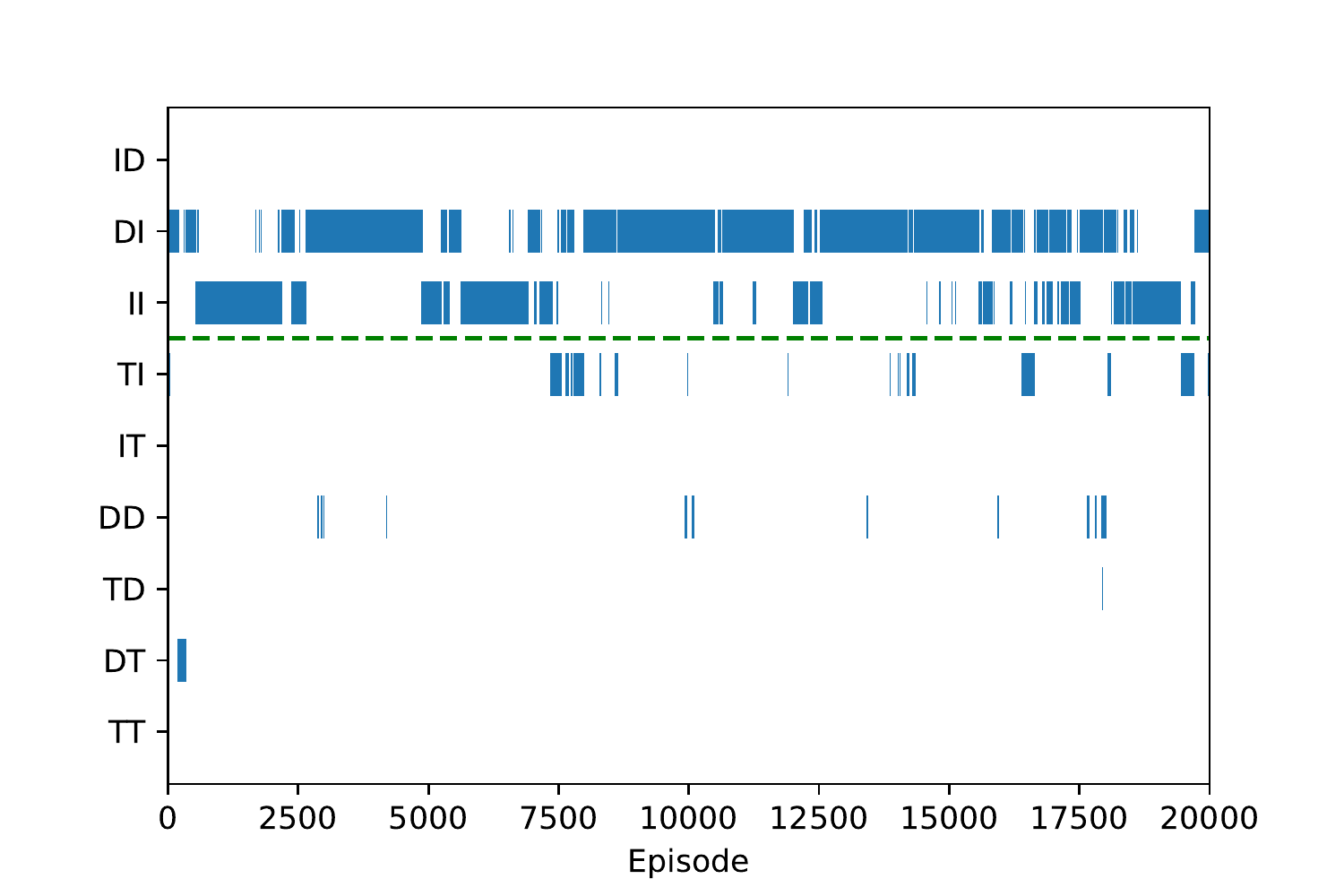}
         \caption{Policy TI}
         \label{fig:MOSS-v TI}
     \end{subfigure}
     \caption[Policy charts for Single-Phase MOSS in new Space Traders MOMDPs]{3 policy charts for Single-Phase MOSS in new variant of Space Traders MOMDPs}
     \label{fig:MOSS-v Policy Charts}
\end{figure}
\begin{table}[hbt!]
    \centering
    \begin{tabular}{@{}|c|c|c|c|c|c|c|@{}}
    \toprule
    Policy & DI & ID & II & IT & TI & DD\\ \midrule
    Original & \textcolor{red}{15} & 0 & 0 & 3 & 2 & 0\\ \midrule
    New variant & 15 & \textcolor{red}{0} & 0 & 3 & 2 & 0\\ \bottomrule
    \end{tabular}%
    \caption[20 independent runs of the Algorithm \ref{algo:mossql}]{The final greedy policies learned in 20 independent runs of the Algorithm \ref{algo:mossql} for new Space Traders environment. The red color indicate the number of desired optimal policy which agent converges to}
    \label{tab:Space-Traders-MOSS-v}
\end{table}
All the changes compared with original one have been highlight in red color. The main different is that the time penalty for each action has been swapped from state A to state B. The new probability of success and reward values for each state-action pair in the new variant Space Traders has showed in table \ref{tab:Space-Traders-MOSS-O1}. Since the only different between policy DI and ID in original Space Traders Problem is the second objective - time penalty. Therefore in this new variant of Space Traders problem, policy ID has become the desired optimal policy as we can see from table \ref{tab:Space-Traders-MOSS-O2}.\newline
After 20 trials of empirical results show that the desired optimal policy (ID) was not converged to in practice. The most common result (15/20 runs) is still policy (DI). As the result show in policy chart (Figure \ref{fig:MOSS-v Policy Charts}),
agent stays on policy DI for most of the time across 20,000 episodes which has show the same pattern in original Space Trader problem. A closer examination of MOSS algorithm \ref{algo:mossql} reveals that the estimated values on which that decision is based $s_t$, $P(s_t)$, $p(s_t)$ and $E_{\not{s_{t+1}}}$ must be based only on the trajectories produced during execution of the greedy policy, whereas in the current algorithm \ref{algo:mossql} they are derived from all trajectories. As the result, the value of $p(s_t)$ \footnote{The estimated probability of visiting state s in any episode} is below 1 because of exploratory actions. The $U(a)$ value at state B for direct and teleport action are below threshold for first objective. So it can already be seen that this agent will not converge to the desired policy ID.
\section{Conclusion}
MOSS algorithm is clearly on the right track to address the stochastic SER issue as it outperforms the baseline method in original Space Traders problem. One potential solution for solving the problem in the new variant of Space Traders is to separate MOSS algorithm into two phases - data-gathering phase and learning phase and this will be discussed in next chapter.
\chapter{Two-Phase MOSS}
\label{chapter7}
As we discussed early in chapter 6, a potential problem could occur in previous introduced MOSSQ-learning algorithm(Algorithm \ref{algo:mossql}). In order to select greedy action in Line 22, the global information must be based only on the trajectories produced during execution of the greedy policy. One potential approach to address this problem is to separate algorithm into two phases: the global statistic data-gathering phase and normal learning phase for Q values. During the data-gather phase, agent only execute actions based on current best knowledge of greedy policy and meanwhile associated global statistic data would be gathered. And during learning phase, agent could select exploratory actions and meanwhile the Q values are updated. Ultimately those estimated global statistic values should converge to those values which are associated with the greedy policy, but it is still possible that some errors occurred in these estimates may prevent agent to find out the SER-optimal policy.
\section{Algorithm}
Therefore, to support this idea, algorithm \ref{algo:mossql-phased} introduces a variant of the previous MOSSQ-learning (Algorithm \ref{algo:mossql}) in which the global statistics are derived directly from episodes where greedy policy is strictly followed.\newline
As a consequence of this implementation, a potential issue could arise during the second phase. Either due to following a exploratory policy or the stochastic environment, agent may end up into a state which was not visited during the previous data-gathering phase. Because of this, the estimated probability of occurrence $p(s)$ will be 0. In this case, all actions would have the same utility value for this state. In another words, all actions including those which are Pareto-dominated action would have equal likelihood to be selected by agent. So, to avoid this, a small change to the update-statistics algorithm has  been introduced, highlighted in red in Algorithm \ref{algo:update-stats2}
\begin{algorithm}[hbt!]
  \caption{The update-statistics helper algorithm for the two-phased form of MOSSQ-learning (Algorithm \ref{algo:mossql-phased}). Changes from Algorithm \ref{algo:update-stats1} are highlighted in red.}
  \label{algo:update-stats2}

  \begin{algorithmic}[1]
    \Statex input: state $s$, accumulated rewards in the current episode $P$, \textcolor{red}{lower bound on probability weighting $\epsilon$}
    \If {$b(s)=0$} \Comment{first visit to $s$ in this episode}
        \State $v(s) = v(s) + 1$ \Comment{increment count of visits to $s$}
        \State $b(s)=1$ \Comment{set flag so duplicate visits within an episode are not counted}
    \EndIf
    \State $P(s) = P(s) + \alpha (P-P(s))$
    \State $s^A$ = $(s,P(s))$ \Comment{augmented state}
    \State $p(s) = v(s)/v_\pi$    \Comment{estimated probability of visiting $s$ in any episode}
    \If {p(s)=1} \Comment{treat states which are always visited as a special case}
        \For {each action $a$}
            \State $U(a) = P(s)+Q(s^A,a))$ 
        \EndFor
    \Else
        \State $E_{\not{s}}=(E_\pi - p(s)E_s)/(1-p(s))$ \Comment estimated return in episodes where $s$ is not visited
        \State \Comment{calculate estimated value over all episodes, assuming a is executed in $s^A$}
        \textcolor{red}{\State$p^\epsilon=max(p(s),\epsilon)$}
        \For {each action $a$}
            \State $U(a) =  \textcolor{red}{p^\epsilon}(P(s)+Q(s^A,a)) + (1-\textcolor{red}{p^\epsilon})E_{\not{s}}$ 
        \EndFor
    \EndIf
    \State return $s^A$, $U$
  \end{algorithmic}
\end{algorithm}

\begin{algorithm}[tbp]
  \footnotesize
  \caption{A two-phased variant of multi-objective stochastic state Q($\lambda$) algorithm (MOSSQ-learning).}
  \label{algo:mossql-phased}

  \begin{algorithmic}[1]
    \Statex input: learning rate $\alpha$, discounting term $\gamma$, eligibility trace decay term $\lambda$, number of objectives $n$, action-selection function $f$ and any associated parameters, duration of data-gathering and learning phases in episodes $D_D$ and $D_L$
    \For {all states $s$, actions $a$ and objectives $o$}
    	\State initialise $Q_o(s,a)$
    \EndFor
    \While {not finished}
	    \State initialise $P_o(s)$, $v(s)=0$, $E_\pi$, $v_\pi=0$
        \For {each episode $1..D_D$} \Comment {Data-gathering phase}
            \State $v_\pi = v_\pi + 1$ \Comment{increment episode counter}
            \For {all states $s$ and actions $a$}
        		\State $b(s)=0$ \Comment{binary flag - was $s$ visited in this episode?
            }
            \EndFor
            \State sums of prior rewards $P_o$ = 0, for all $o$ in 1..$n$
        	\State observe initial state $s_t$
        	\\ \Comment{call helper algorithm to update stats, create augmented state \& utility vector}
        	\State $s^A_t$, $U(s^A_t)$ = update-statistics($s_t$,$P$)
            \State select $a^*$ from a greedy policy derived using $f(U(s^A_t))$
            \For {each step of the episode}
 			    \State execute $a^*_t$, observe $s_{t+1}$ and reward $R_t$
 			    \State $P = P + R_t$
    	        \State $s^A_{t+1}$, $U(s^A_{t+1})$ = update-statistics($s_{t+1}$,$P$)
 			\State select $a^*$ from a greedy policy derived using  $f(U(s^A_{t+1}))$
    	\EndFor
    	\State $E_\pi = E_\pi + \alpha(P-E_\pi)$ \Comment update estimates of per-episode return
        \For {all states $s$ with $b(s)\neq0$}
              \State $E(s) = E(s) + \alpha(P-E(s))$
        \EndFor
    \EndFor
    \For {each episode $1..D_L$} \Comment {Learning phase}
            \For {all states $s$ and actions $a$}
        		\State $e(s,a)$=0; $b(s)=0$
            \EndFor
        	\State observe initial state $s_t$
        	\State $s^A_t$ = $(s_t,P(s_t))$ \Comment{create augmented state \& utility vector}
            \For {each action $a$}
                \State $U(s^A,a) = p(s^A_t)(P(s^A_t)+Q(s^A,a)) + (1-p(s^A_t))E_{\not{s^A_t}}$ 
            \EndFor
            \State select $a^*$ from a greedy policy derived using  $f(U(s^A_{t+1}))$
  			\State select $a^\prime$ from an exploratory policy derived using $f(U(s^A_{t+1}))$
            \State $\delta = R_t + \gamma Q(s^A_{t+1},a^*) - Q(s^A_t,a_t)$
            \State $e(s^A_t,a_t)$ = 1
            \For {each augmented state $s^A$ and action $a$}
            	\State $Q(s^A,a) = Q(s^A,a) + \alpha\delta e(s^A,a)$
                 \If {$a^\prime = a^*$} 
                    \State $e(s^A,a) = \gamma \lambda e(s^A,a)$
                 \Else
                    \State $e(s^A,a) = 0$                 
                 \EndIf
            \EndFor
            \State $s^A_t = s^A_{t+1}, a_t = a^\prime$
    	\EndFor
    \EndWhile
  \end{algorithmic}
\end{algorithm}
\section{Result}
\begin{figure}[hbt!]
     \centering
     \begin{subfigure}[h]{0.45\textwidth}
         \centering
         \includegraphics[width=\textwidth]{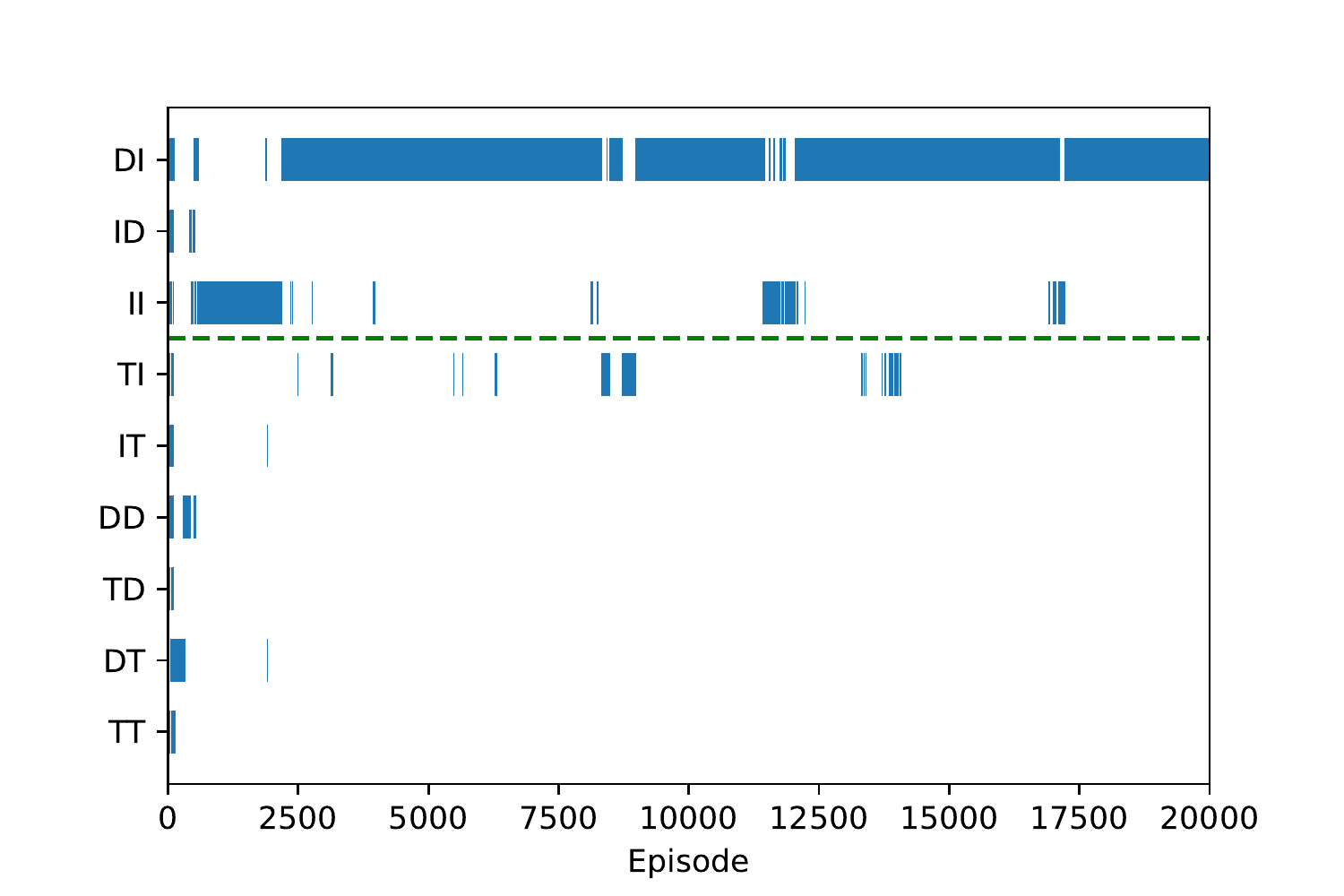}
         \caption{Policy DI}
         \label{fig:MOSSTP DI}
     \end{subfigure}
     \begin{subfigure}[h]{0.45\textwidth}
         \centering
         \includegraphics[width=\textwidth]{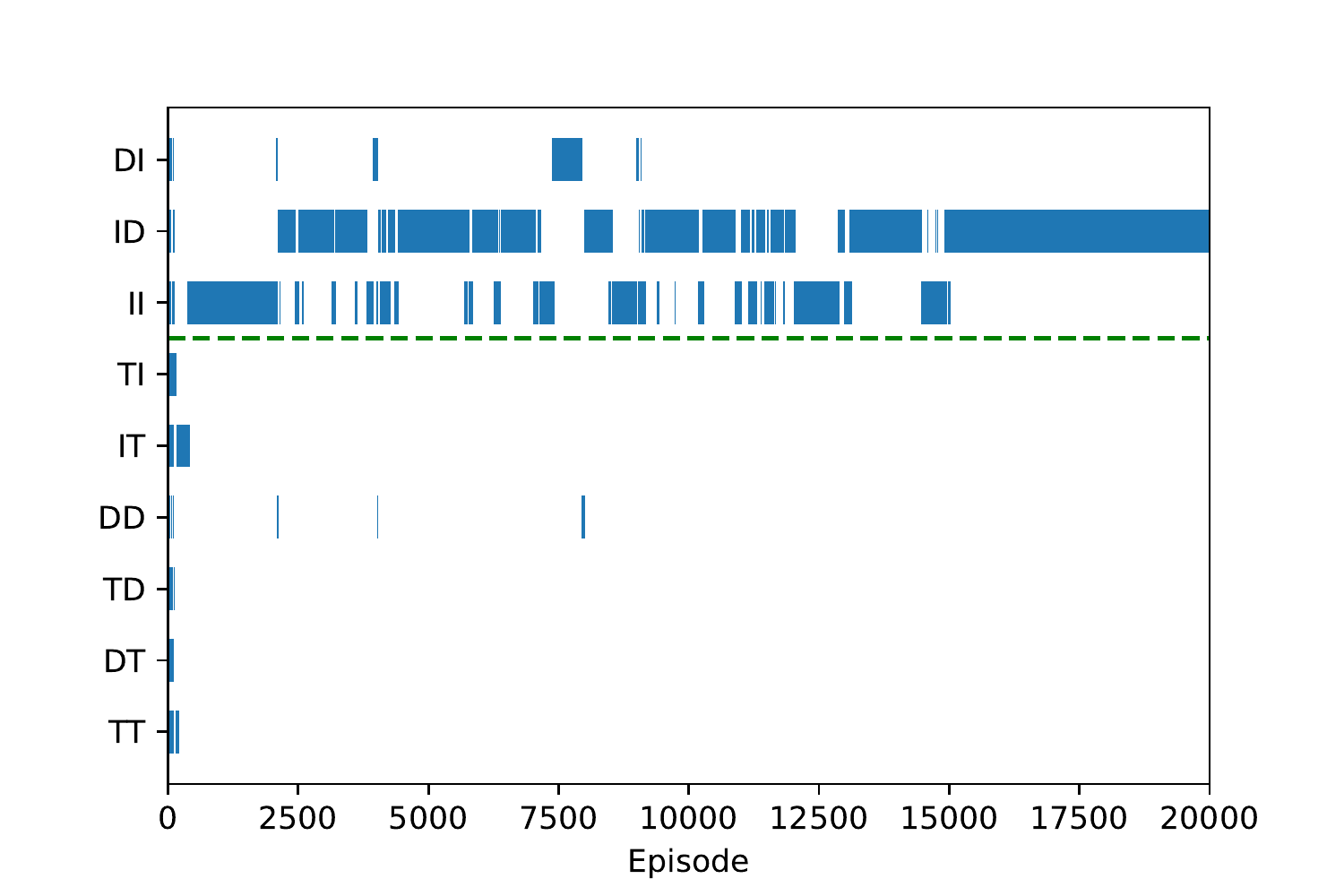}
         \caption{Policy ID}
         \label{fig:MOSSTP ID}
     \end{subfigure}
     \begin{subfigure}[h]{0.45\textwidth}
         \centering
         \includegraphics[width=\textwidth]{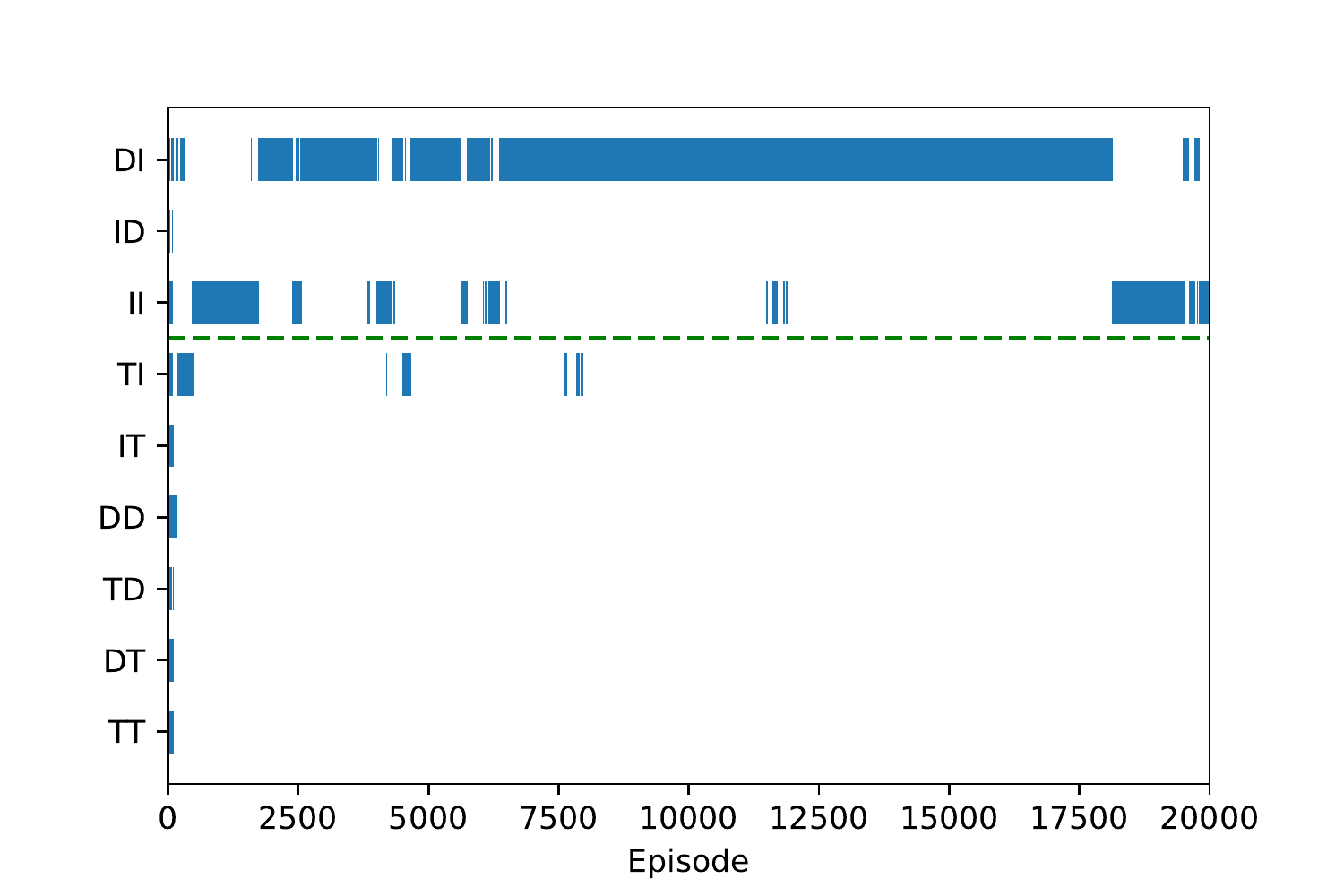}
         \caption{Policy II}
         \label{fig:MOSSTP II}
     \end{subfigure}
     \caption[Policy charts for Two-Phase MOSS]{3 policy charts for Two-Phase MOSS Algorithm}
     \label{fig:MOSSTP Policy Charts}
\end{figure}
\begin{table}[hbt!]
    \centering
    \begin{tabular}{@{}|c|c|c|c|c|c|c|@{}}
    \toprule
    Policy & DI & ID & II & IT & TI & DD\\ \midrule
    Baseline & 1 & 13 & 4  & 2  & 0 & 0\\ \midrule
    Reward Design & 10 & 5 & 1 & 1 & 2 & 1\\ \midrule
    Single-Phase MOSS & 15 & 0 & 0 & 3 & 2 & 0\\ \midrule
    Two-Phase MOSS & 13 & 6 & 1 & 0 & 0 & 0\\ \bottomrule
    \end{tabular}%
    \caption[20 independent runs of the Algorithm \ref{algo:mossql-phased}]{The final greedy policies learned in 20 independent runs of the Algorithm \ref{algo:mossql-phased} for Space Traders environment}
    \label{tab:Space-Traders-MOSSTP}
\end{table}
As we can see from table \ref{tab:Space-Traders-MOSSTP}, the most common result (13/20 runs) is the DI policy, which is the desired optimal policy, but the ID policy (6 repetitions) and II policy (1) also occur in some trials. Figure \ref{fig:MOSS Policy Charts} also indicates that even the final policy can converge to Policy II at the end of training. Most of time, agent believes policy DI is the desired optimal policy. But surprisingly in policy ID chart, agent stays on ID policy most of time compared with policy II and optimal policy DI.
\section{Discussion}
\begin{figure}[hbt!]
     \centering
     \begin{subfigure}[h]{0.45\textwidth}
         \centering
         \includegraphics[width=\textwidth]{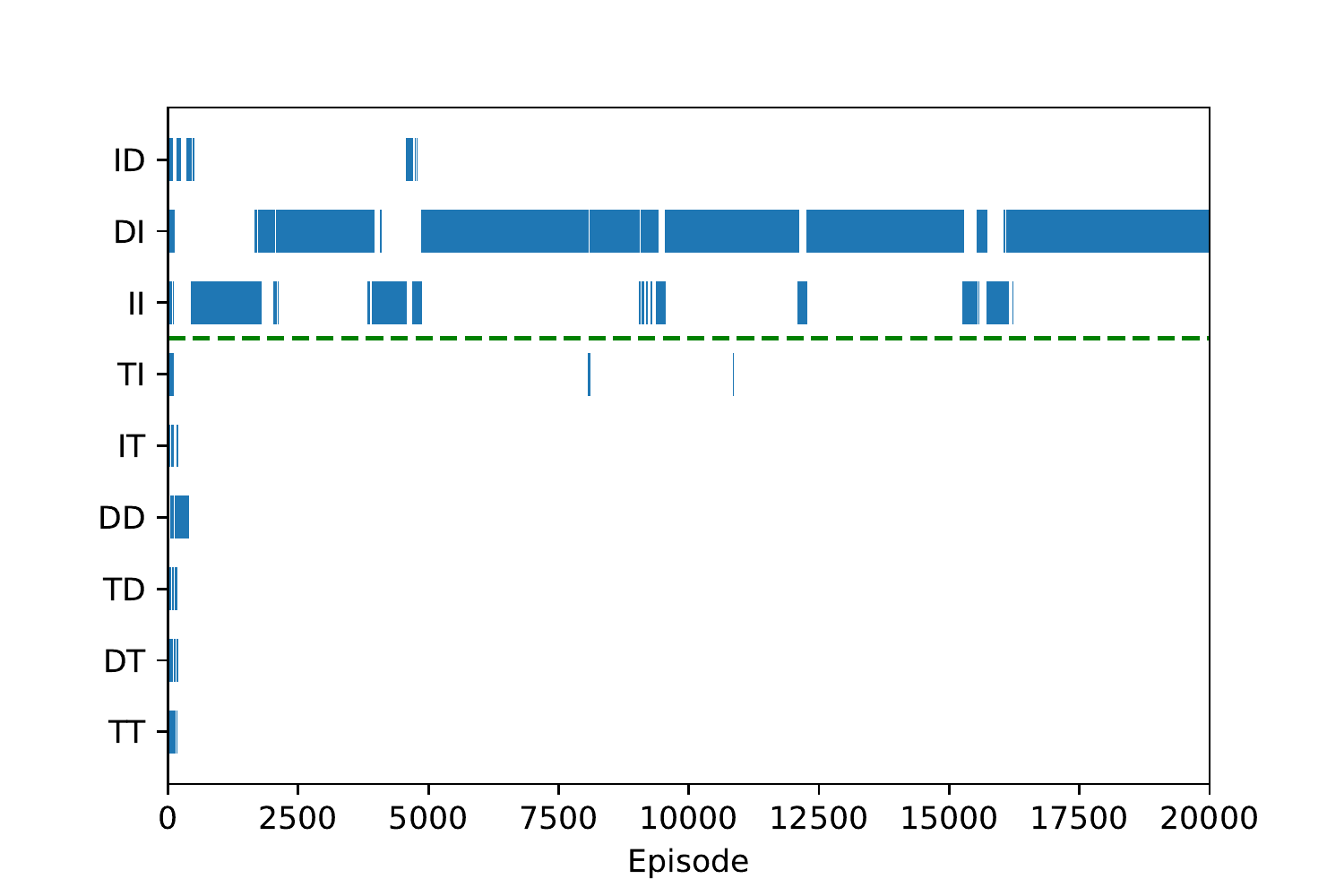}
         \caption{Policy DI}
         \label{fig:MOSSTP-v DI}
     \end{subfigure}
     \begin{subfigure}[h]{0.45\textwidth}
         \centering
         \includegraphics[width=\textwidth]{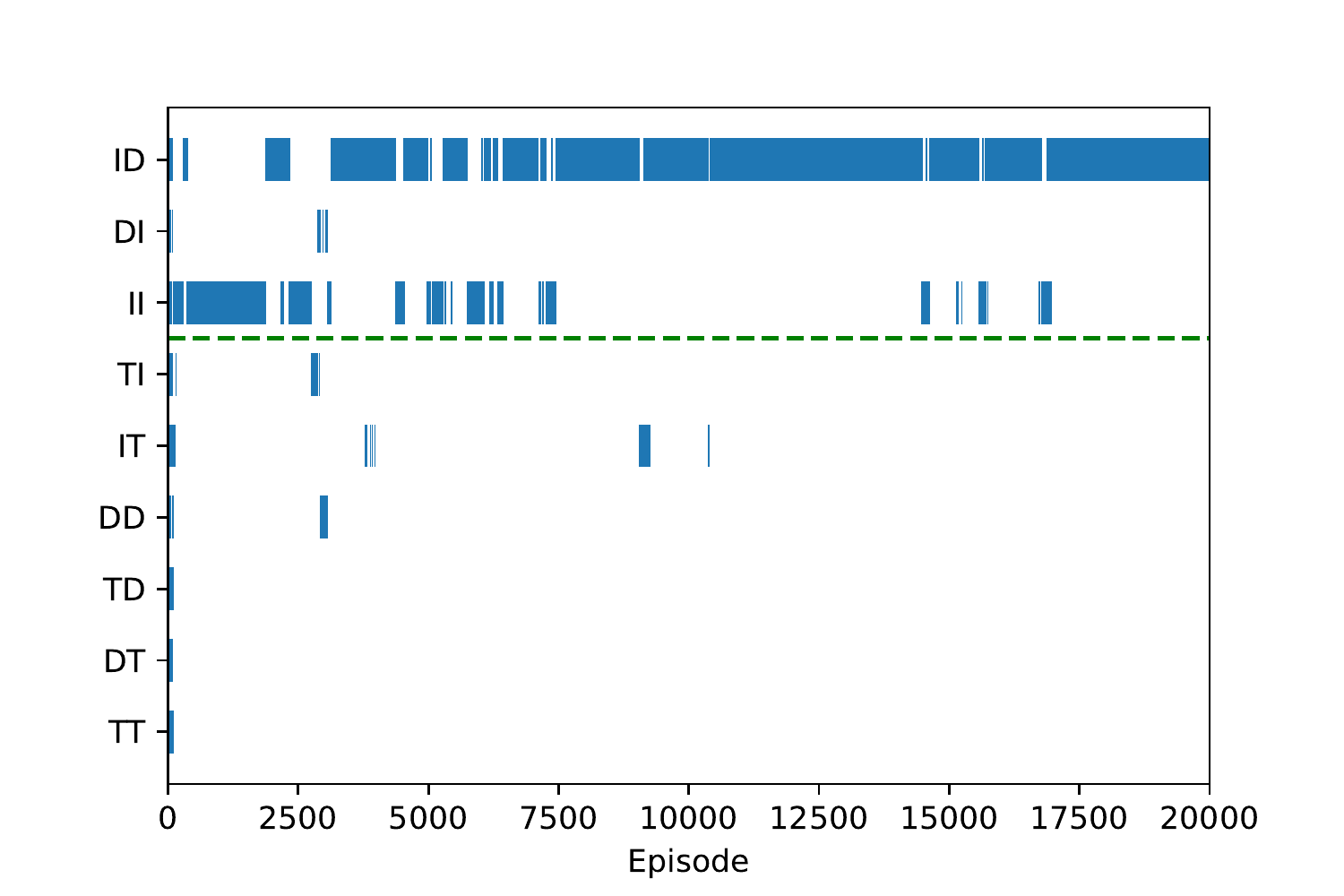}
         \caption{Policy ID}
         \label{fig:MOSSTP-v ID}
     \end{subfigure}
     \caption[Policy charts for Two-Phase MOSS in new Space Traders MOMDPs]{2 policy charts for Two-Phase MOSS in new Space Traders MOMDPs}
     \label{fig:MOSSTP-v Policy Charts}
\end{figure}

Compared with benchmark result in table \ref{tab:Space-Traders-MOSSTP}, Two Phase MOSS algorithm still outperform the baseline method in original Space Traders problem. But in order to test whether this MOSS algorithm is going to work for new variant of Space Traders as well. Another 20 trials of experiment have been conducted. The empirical result in table \ref{tab:Space-Traders-MOSSTP-v} shows that the desired optimal policy (ID) was converged only 6 out 20 runs in practice. The most common result (15/20 runs) is still policy (DI). The potential explanation for this could be even during the data-gathering phase, the executed actions are based on current best knowledge of greedy policy. It is not necessary aligned with the final optimal policy which we are looking for. As the result, the data-gathering phase still includes sub-optimal trajectories data. In another words, those estimated global statistic values do not converge to those values which are associated with the desired optimal policy as we are expected early. Because of this reason, the two-phase MOSS algorithm is easy trapped into sub-optimal policy and this is also explain why in Figure \ref{fig:MOSSTP-v Policy Charts} most of time agent thinks DI is optimal policy in Figure \ref{fig:MOSSTP-v DI} but most of time agent thinks ID is optimal policy in Figure \ref{fig:MOSSTP-v ID}. Therefore, there is no guarantee to find SER-optimal policy at the end of training. \newline
Also there is one clear limitation for Two-Phase MOSS Algorithm, which is the learning time has been increased because of two extra parameters - the episode for data-gather phase $D_D$ and the episode for learning phase $D_L$. Currently, there is no good method to adjust these two extra parameters automatically. So, in practise, trial-and-error is still the only option to change these two phase parameters. 
\begin{table}[hbt!]
    \centering
    \begin{tabular}{@{}|c|c|c|c|c|c|c|@{}}
    \toprule
    Policy & DI & ID & II & IT & TI & DD\\ \midrule
    Original & \textcolor{red}{13} & 6 & 1 & 0 & 0 & 0\\ \midrule
    New variant & 14 & \textcolor{red}{6} & 0 & 0 & 0 & 0\\ \bottomrule
    \end{tabular}%
    \caption[20 independent runs of the Algorithm \ref{algo:mossql-phased} for new Space Traders MOMDPs]{The final greedy policies learned in 20 independent runs of the Algorithm \ref{algo:mossql-phased} for new Space Traders environment}
    \label{tab:Space-Traders-MOSSTP-v}
\end{table}
\section{Conclusion}
Despite the two-phase MOSS algorithm does not address the stochastic SER issue in general, it still outperforms the baseline method in original Space Traders problem. In order to solve the stochastic SER issue at least for small and simple test environment likes Space Traders, we propose option learning which will be discussed in next chapter.
\chapter{Option learning}
\label{chapter8}
The third approach is to use the concept of options. In here every option is a 'meta-action' which determines the action to be selected by agent over multiple time-steps rather than single time-step. Normally in \acrlong{RL}, agent needs to learn the options by interacting with the environment. But for original Space Traders, agent will simply pre-define 9 options instead before running the experiment. Also agent needs to strictly follow those options until reach to the terminal state. Compared with normal state-action value in RL, agent need to learn state-option value instead. Each state-option value agent has learnt at state A should match the mean reward for each nine deterministic policies in table \ref{tab:Space-Traders-O2} from chapter 3.
\section{Algorithm}
\begin{algorithm}
  \caption{Multiobjective Q($\lambda$) with policy options.}
  \label{algo:moql-options}
  \begin{algorithmic}[1]
    \Statex input: learning rate $\alpha$, discounting term $\gamma$, eligibility trace decay term $\lambda$, number of objectives $n$, action-selection function $f$ and any associated parameters, set of policy options $P$
    \For {all states $s$, options $p$ and objectives $o$}
    	\State initialise $Q_o(s,p)$
    \EndFor
    \For {each episode} 
        \For {all states $s$ and options $p$}
    		\State $e(s,p)$=0
        \EndFor
    	\State observe initial state $s_t$
        \State select option $p_e$ using $f(Q(s_t))$ (with possible exploratory selection)
        \State select $a_t$ from $p_e(s_t)$
        \For {each step of the episode}
 			\State execute $a_t$, observe $s_{t+1}$ and reward $R_t$
 			\State select $a^\prime$ from $p_e(s_{t+1})$
            \State $\delta = R_t + \gamma Q(s_{t+1},p_e) - Q(s_t,P_e)$
            \State $e(s_t,p_e)$ = 1
            \For {each state $s$}
            	\State $Q(s,p_e) = Q(s,p_e) + \alpha\delta e(s,p_e)$
                \State $e(s,p_e) = \gamma \lambda e(s,p_e)$
            \EndFor
            \State $s_t = s_{t+1}, a_t = a^\prime$
    	\EndFor
    \EndFor
  \end{algorithmic}
\end{algorithm}
\section{Result}
\begin{figure}[hbt!]
     \centering
     \begin{subfigure}[h]{0.45\textwidth}
         \centering
         \includegraphics[width=\textwidth]{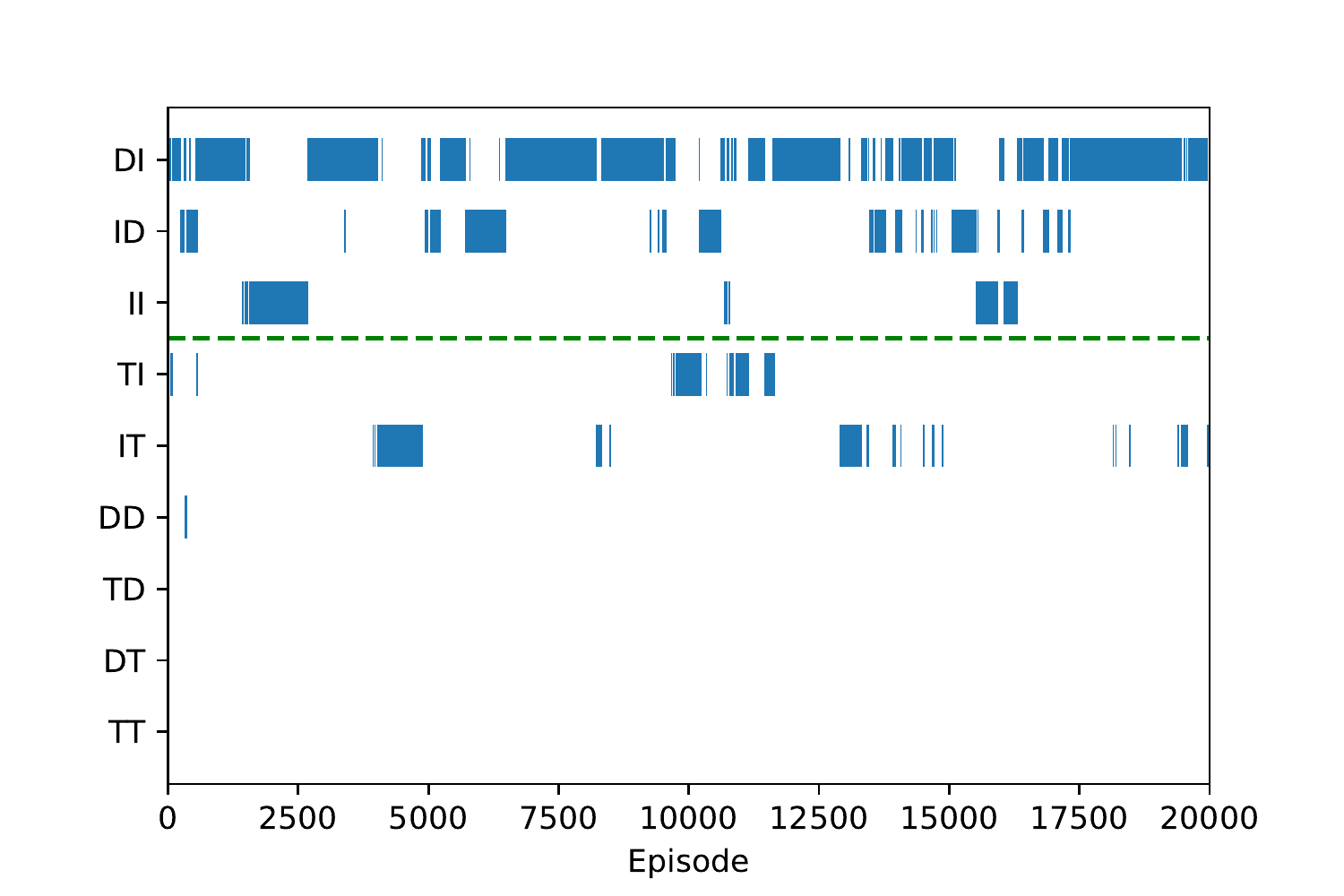}
         \caption{Policy DI}
         \label{fig:Option DI}
     \end{subfigure}
     \begin{subfigure}[h]{0.45\textwidth}
         \centering
         \includegraphics[width=\textwidth]{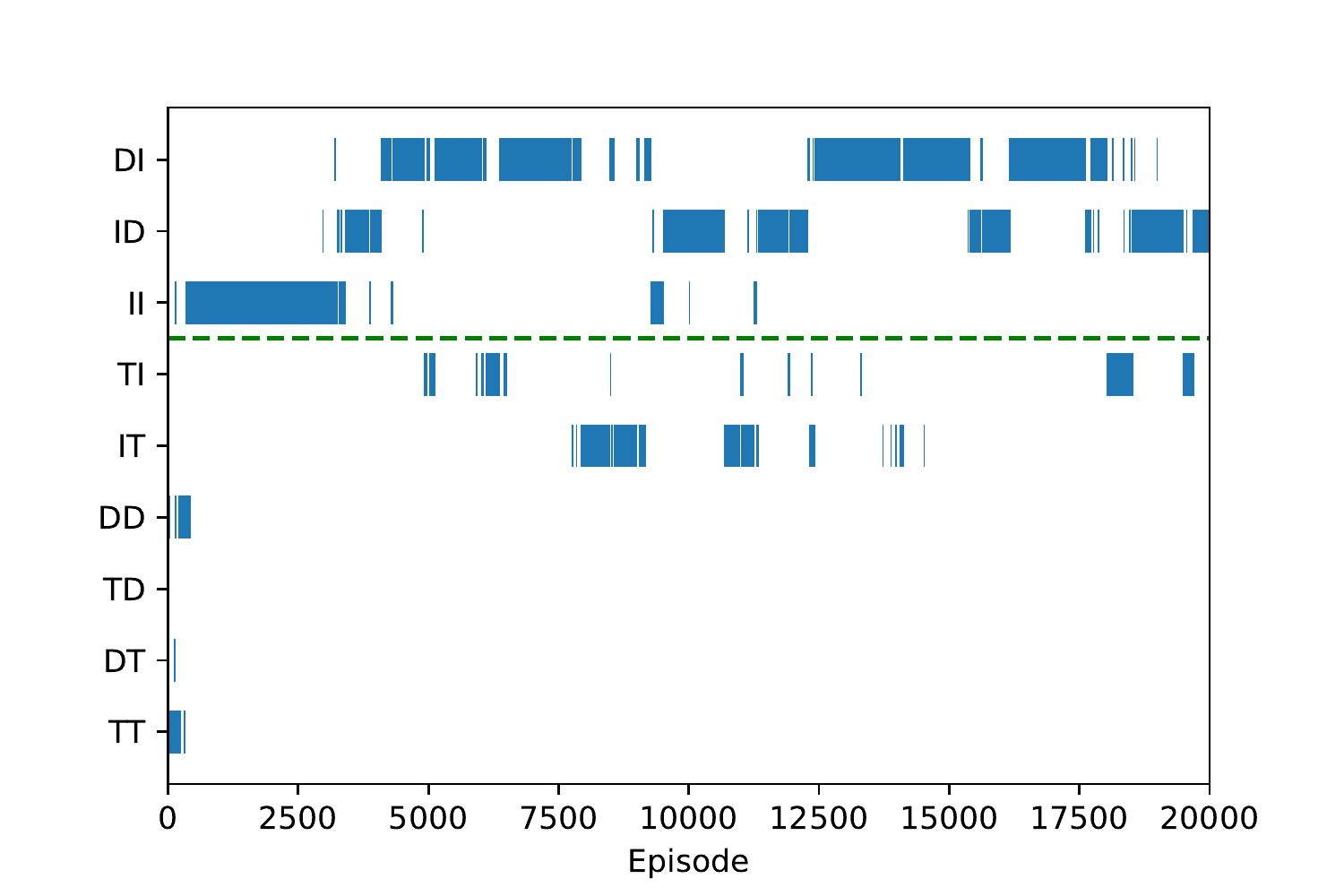}
         \caption{Policy ID}
         \label{fig:Option ID}
     \end{subfigure}
     \begin{subfigure}[h]{0.45\textwidth}
         \centering
         \includegraphics[width=\textwidth]{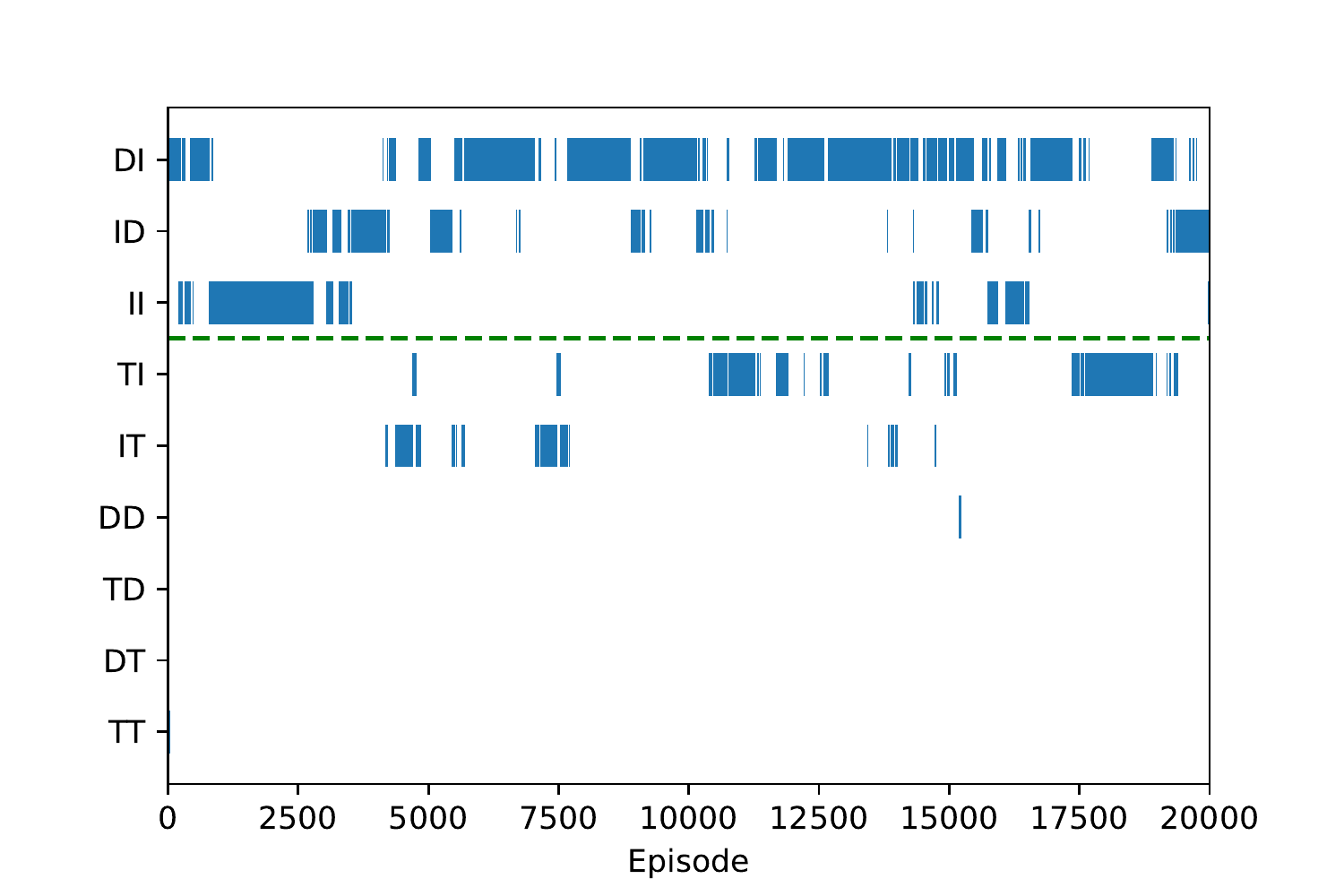}
         \caption{Policy II}
         \label{fig:Option II}
     \end{subfigure}
          \begin{subfigure}[h]{0.45\textwidth}
         \centering
         \includegraphics[width=\textwidth]{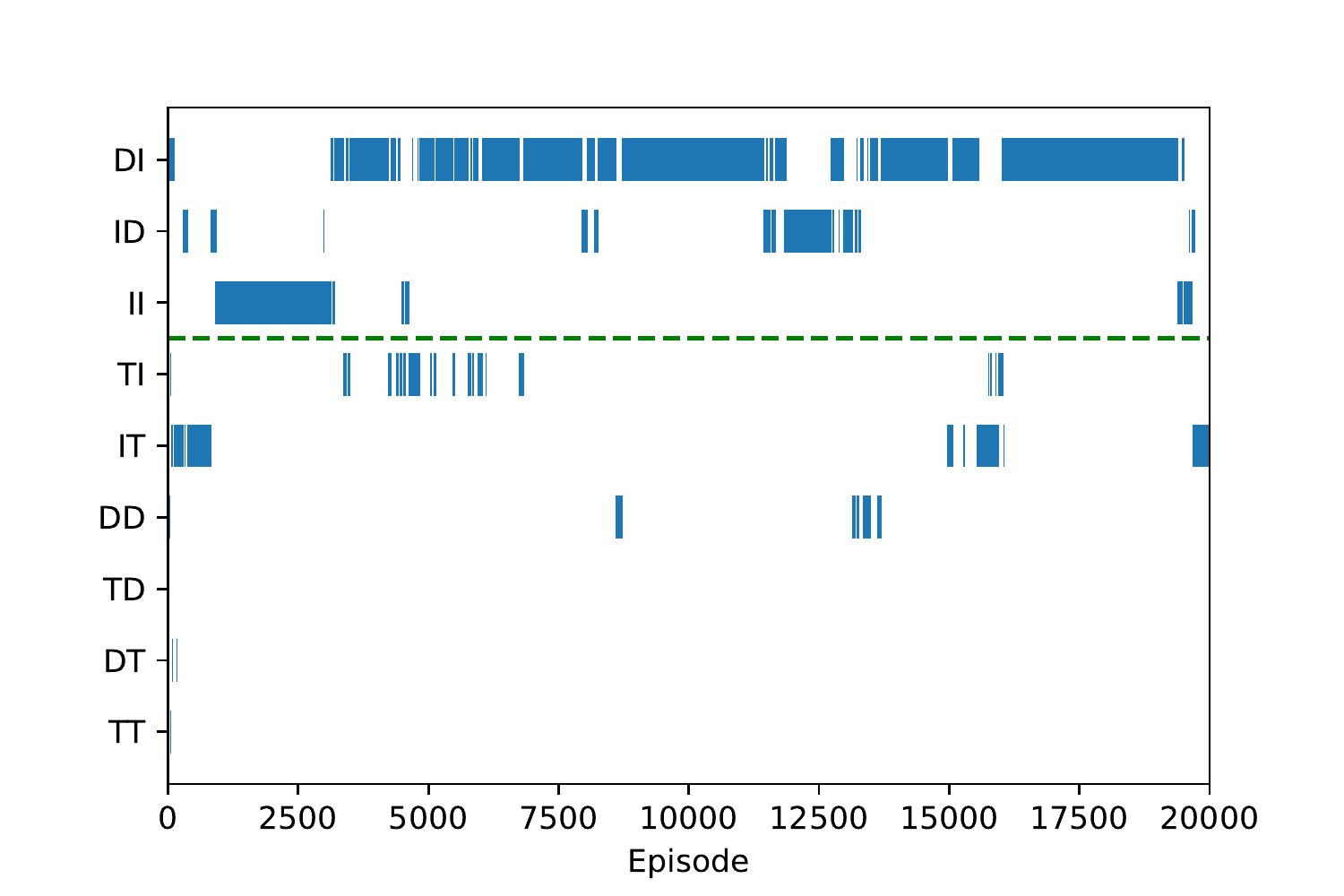}
         \caption{Policy IT}
         \label{fig:Option IT}
     \end{subfigure}
     \begin{subfigure}[h]{0.45\textwidth}
         \centering
         \includegraphics[width=\textwidth]{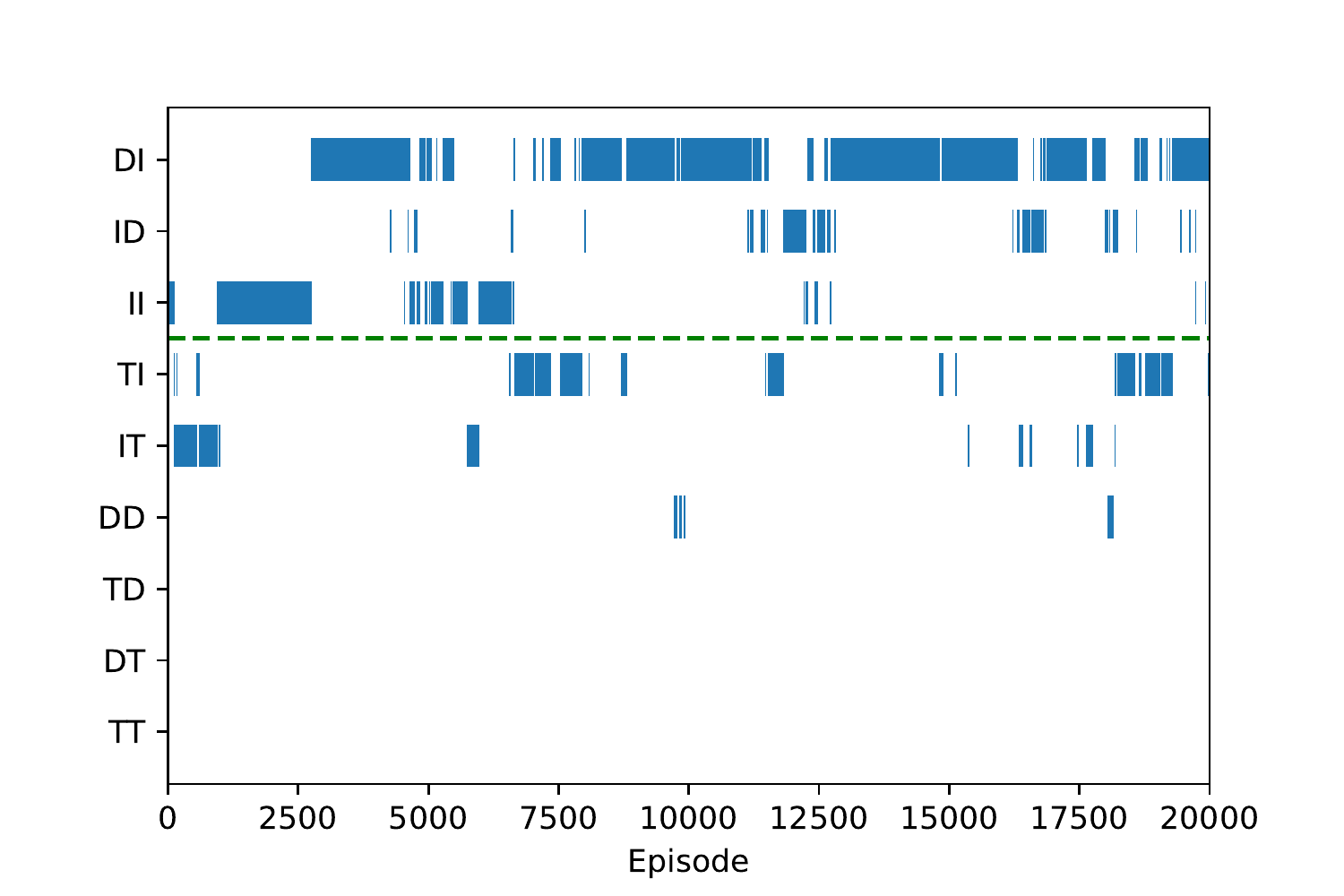}
         \caption{Policy TI}
         \label{fig:Option TI}
     \end{subfigure}
     \caption[Policy charts for Option Learning]{5 policy charts for sample runs of the Options MOQ-Learning algorithm}
     \label{fig:Option Policy Charts}
\end{figure}
\begin{table}[hbt!]
    \centering
    \begin{tabular}{@{}|c|c|c|c|c|c|c|@{}}
    \toprule
    Policy & DI & ID & II & IT & TI & DD\\ \midrule
    Baseline & 1 & 13 & 4  & 2  & 0 & 0\\ \midrule
    Reward Design & 10 & 5 & 1 & 1 & 2 & 1\\ \midrule
    Single-Phase MOSS & 15 & 0 & 0 & 3 & 2 & 0\\ \midrule
    Two-Phase MOSS & 13 & 6 & 1 & 0 & 0 & 0\\ \midrule
    Option & 14 & 2 & 1 & 1 & 2 & 0\\ \bottomrule
    \end{tabular}%
    \caption[20 independent runs of the Algorithm \ref{algo:moql-options}]{The final greedy policies learned in 20 independent runs of the Options MOQ-Learning algorithm for Space Traders environment}
    \label{tab:Space-Traders-Option}
\end{table}
As we can see from table \ref{tab:Space-Traders-Option}, the most common result (14/20 runs) is the DI policy, which is the desired optimal policy, but the ID policy (2 repetitions) and TI policy (1) also occur in some trials. Figure \ref{fig:Option Policy Charts} also indicates that even the final policy converge to Policy ID or IT at the end of training. Most of time, agent believes policy DI is the desired optimal policy. 
\section{Discussion}
Compared with benchmark result in table \ref{tab:Space-Traders-Option}, Option learning algorithm still outperform the baseline method in original Space Traders problem. But in order to test whether this algorithm is going to work for new variant of Space Traders as well. Another 20 trials of experiment have been conducted. 
\subsection{New variant of Space Traders}
The empirical result in table \ref{tab:Space-Traders-Option-v} shows that the desired optimal policy (ID) was successfully converged 13 out 20 runs in practice. However, we notice that in theory the option learning should converge to desired optional policy 20 out 20 because there is not stochastic SER issue any more as agent just strictly follows pre-defined option until it reaches to the terminal state. A closer examination of one trial of result from Option learning algorithm reveals that all previous methods including option learning algorithm suffer from Noisy Q Value estimate issue which has been previously described in \shortciteA{vamplew2021potential} \citeyear{VamplewEnvironmental2022}, the extent to which this interferes with the learning of SER-optimal policies had not previously been made clear.
\begin{figure}[hbt!]
     \centering
     \begin{subfigure}[h]{0.45\textwidth}
         \centering
         \includegraphics[width=\textwidth]{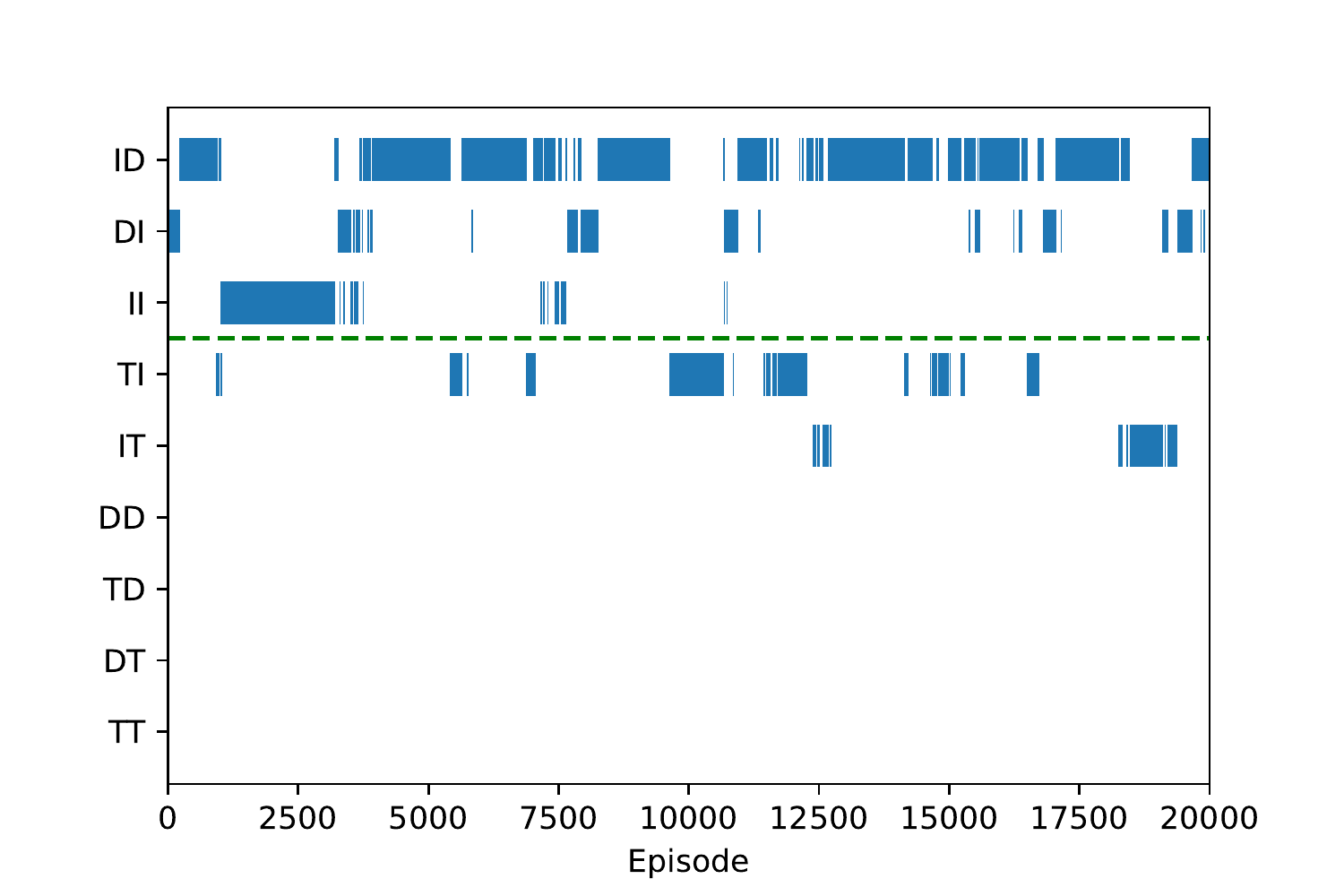}
         \caption{Policy DI}
         \label{fig:Option-v DI}
     \end{subfigure}
     \begin{subfigure}[h]{0.45\textwidth}
         \centering
         \includegraphics[width=\textwidth]{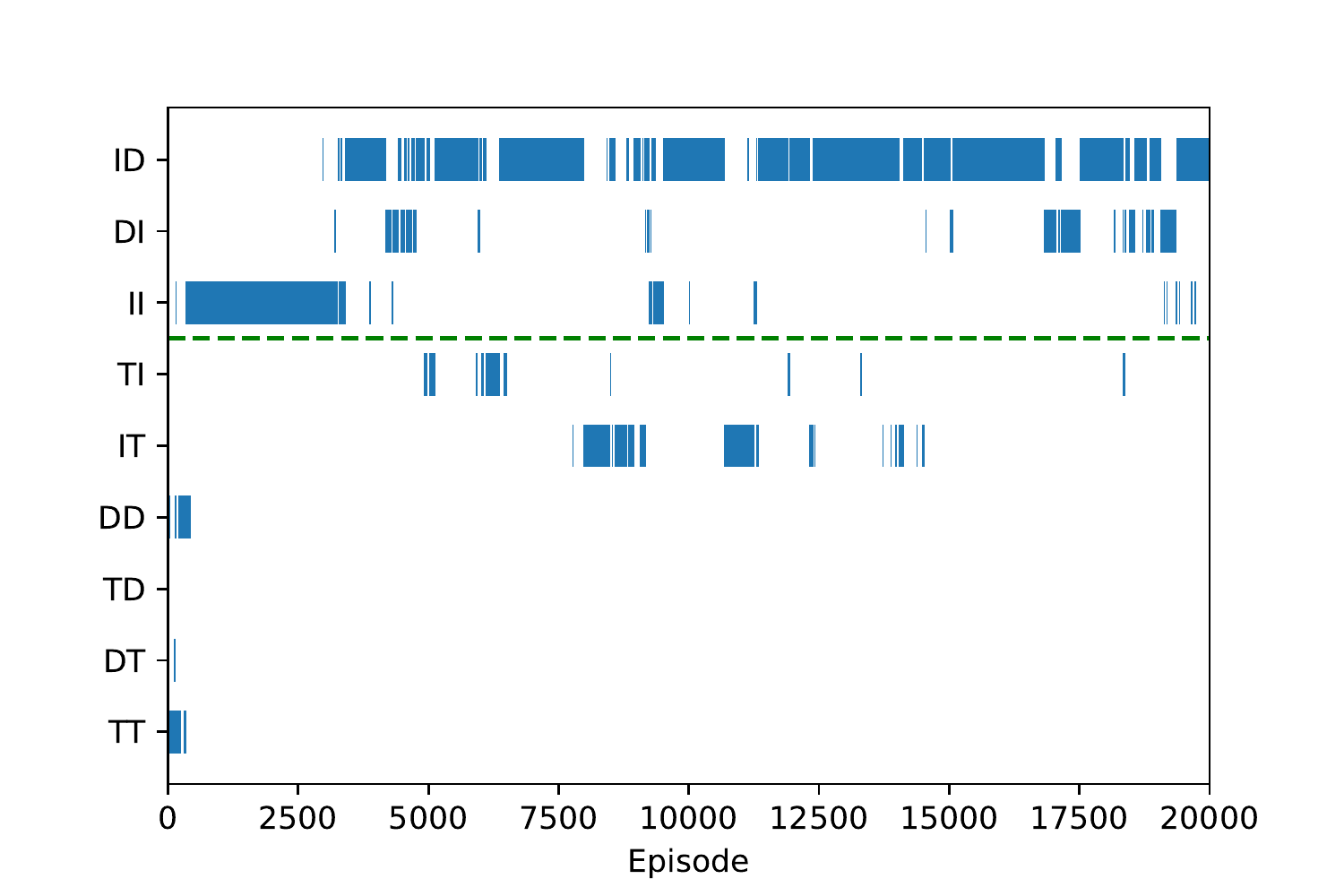}
         \caption{Policy ID}
         \label{fig:Option-v ID}
     \end{subfigure}
     \begin{subfigure}[h]{0.45\textwidth}
         \centering
         \includegraphics[width=\textwidth]{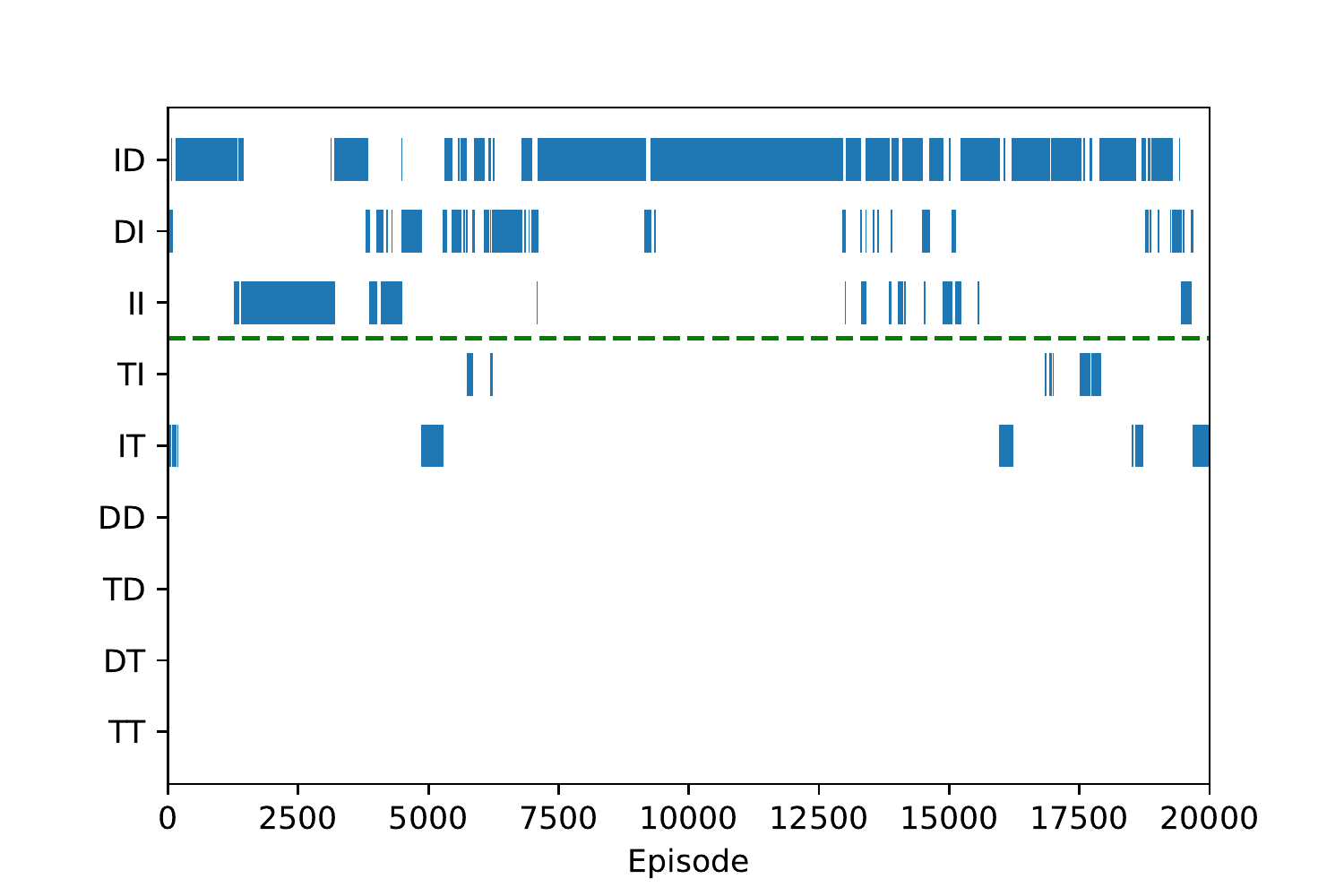}
         \caption{Policy IT}
         \label{fig:Option-v IT}
     \end{subfigure}
     \begin{subfigure}[h]{0.45\textwidth}
         \centering
         \includegraphics[width=\textwidth]{Images/Option-v-IT.pdf}
         \caption{Policy TI}
         \label{fig:Option-v TI}
     \end{subfigure}
     \caption[Policy charts for Option Learning in new Space Traders MOMDPs]{4 policy charts for Option Learning in new Space Traders MOMDPs}
     \label{fig:Option-v Policy Charts}
\end{figure}
\begin{table}[hbt!]
    \centering
    \begin{tabular}{@{}|c|c|c|c|c|c|c|@{}}
    \toprule
    Policy & DI & ID & II & IT & TI & DD\\ \midrule
    Original & 14 & 2 & 1 & 1 & 2 & 0\\ \midrule
    New variant & 3 & 13 & 0 & 2 & 2 & 0\\ \bottomrule
    \end{tabular}%
    \caption[20 independent runs of the Algorithm \ref{algo:moql-options} for new Space Traders MOMDPs]{The final greedy policies learned in 20 independent runs of the Algorithm \ref{algo:moql-options} for new Space Traders environment}
    \label{tab:Space-Traders-Option-v}
\end{table}

\begin{figure}[hbt!]
    \centering
    \includegraphics[width=13.5cm]{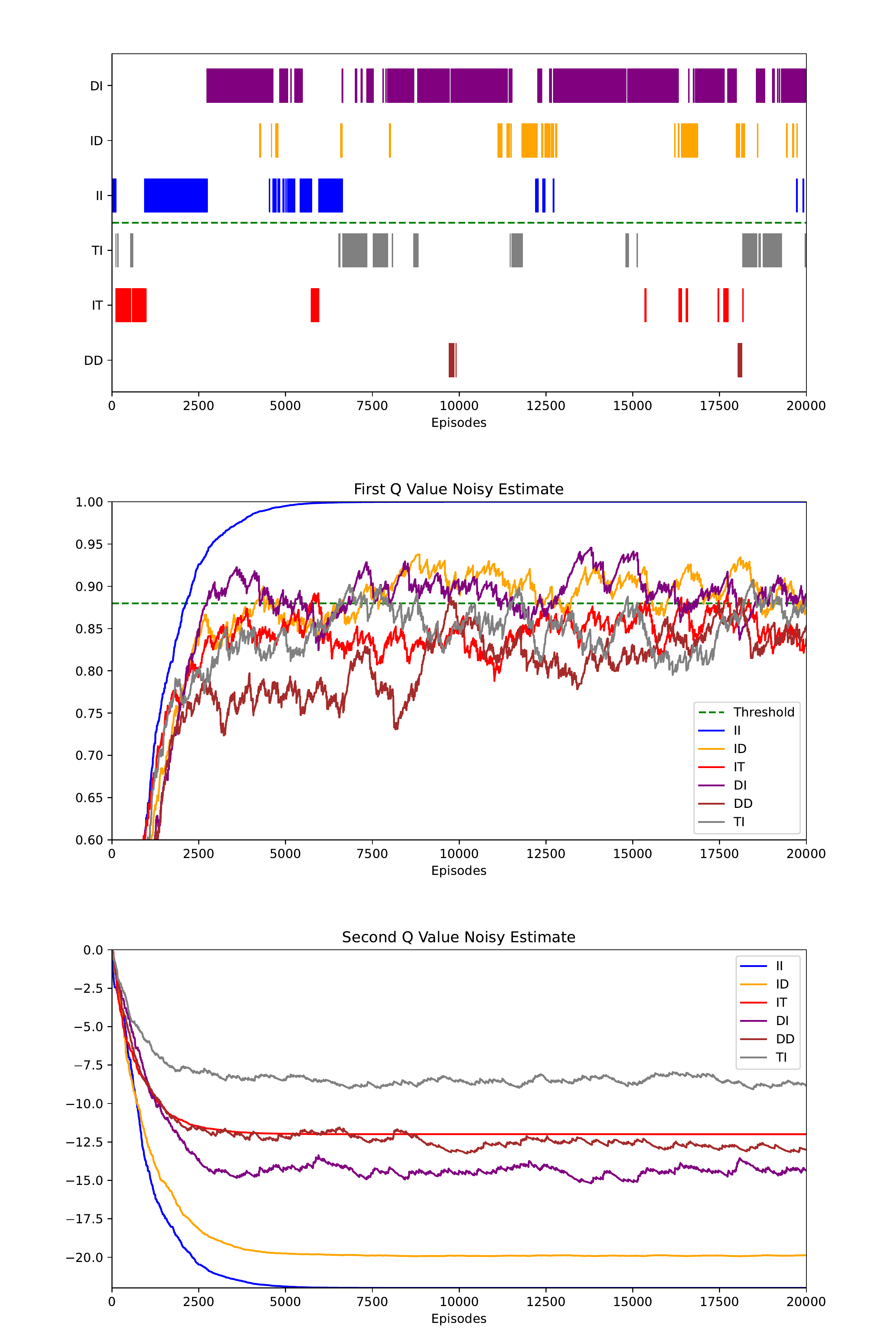}
    \caption[The Noisy Q Value Estimate Issue]{The Noisy Q Value Estimate issue in Option learning. These graphs illustrate agent behaviour for a single run. The top graph shows which option/policy is viewed as optimal after each episode, while the lower graphs show the estimated Q-value for each objective for each option.}
    \label{fig:NOVE-c}
\end{figure}

\begin{figure}[hbt!]
    \centering
    \includegraphics[width=13.5cm]{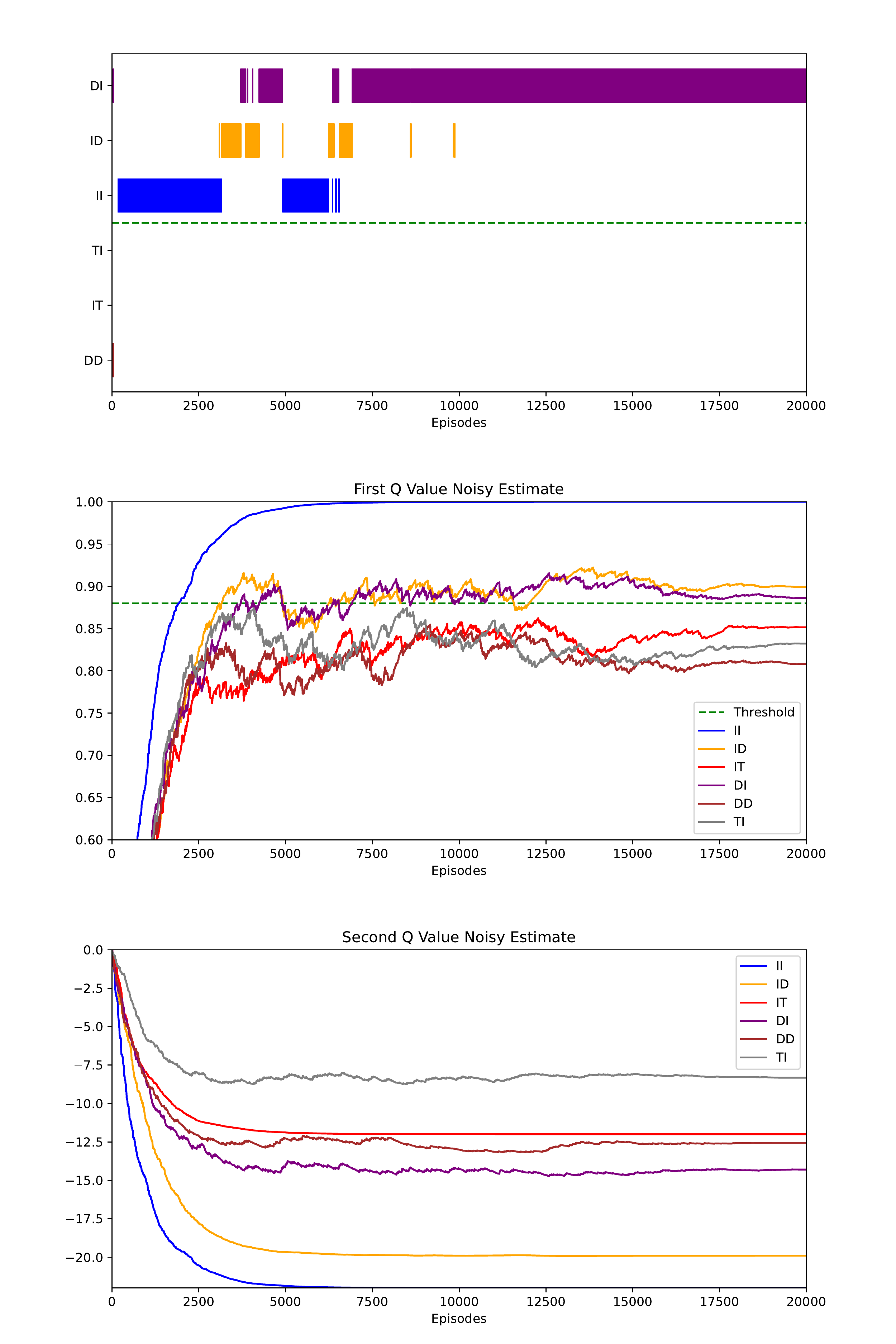}
    \caption[Decay Learning rate in Option learning]{Decay Learning rate in Option learning. Notice the increased stability of both the Q-values and option selection in the later episodes when the learning rate has decayed to a small value.}
    \label{fig:NOVE-d}
\end{figure}
\subsection{Noisy Q value estimate}
Figure \ref{fig:NOVE-c} is one trial of experiment which eventually selects policy (TI) for option learning in original Space Traders problem. The first layer is the normal policy chart where each policy has a unique color for better comparison. The middle layer indicates the first objective in Q value at state A and the bottom layer shows the the second objective in the Q vector value at state A. As we can see from these three graphs, due to the combination of the stochastic environment and hard code threshold, the optimal policy never stabilized even though it stays on DI the desired optimal policy most of time. There is an extreme case around 8,000 to 10,000 episode, where the policy DD (in brown color) gets extremely lucky and it's estimated value rises above the threshold and therefore on the policy chart at the top around this time the agent thinks policy DD is the optimal policy.\newline\newline
One potential explanation for this issue is that all previous methods including option learning algorithm use a constant learning rate which is 0.01 as showed in table \ref{tab:Space-Traders-O3} from Methodology chapter. Even 0.01 is quite a small learning rate already for a tabular problem likes Space Traders. But in practices, this value is still too large for agent to converge at the end of 20,000 episodes. Therefore, one potential solution for this Noisy Q value estimate issue is to gradually decay the learning rate. Figure \ref{fig:NOVE-d} is one trial of experiment where option learning algorithm uses decay learning rate instead in original Space Traders problem. As we can see from the top of policy chart, this time agent did converge to desired optimal policy DI because the decaying learning rate which reduces the influences of the occasional unsuccessful or successful runs leading the Q Values for an action to move from one side of the threshold to the others.\newline
\section{Conclusion}
Combined with decaying the learning rate, options learning is able to address both non-linear scalarisation function for SER criteria and noisy Q value estimate under stochastic environment likes Space Traders Problem. However this method still suffers from a more fundamental problem – the curse of dimension. Because option learning needs to pre-define each options before running the experiment.  For the problems with more states and actions, the numbers of pre-defined options are going to increase exponentially. So this method is not able to scale up to solve more complex problem in real-life. Because the noisy Q value estimate issue existing for all of the previous methods, we ran further experiments on the baseline method and MOSS with the decayed learning rate and which will be discussed in next chapter.
\chapter{Decayed learning rate}
\label{chapter9}
As we discussed in previous chapter, the noisy Q value estimate issue exists for all of the previous methods. The results in Chapter 8 shows that addressing this issue by decaying the learning rate produced a major improvement in the Options MOQ-Learning approach, enabling it to reliably converge to the desired optimal policy. Therefore, there is a need to run further experiments on the baseline method and MOSS algorithm with the decayed learning rate as well. One of the reasons is to isolate the core stochastic SER problem from the effect of the noisy Q value estimation on the 20 independent runs of experiment. 
\section{Baseline method Result and Discussion}
\begin{figure}[hbt!]
    \centering
    \begin{subfigure}[h]{0.45\textwidth}
         \centering
         \includegraphics[width=\textwidth]{Images/Baseline-ID.pdf}
         \caption{Policy ID with constant learning rate}
         \label{fig:Original-ID-1}
    \end{subfigure}
    \begin{subfigure}[h]{0.45\textwidth}
         \centering
         \includegraphics[width=\textwidth]{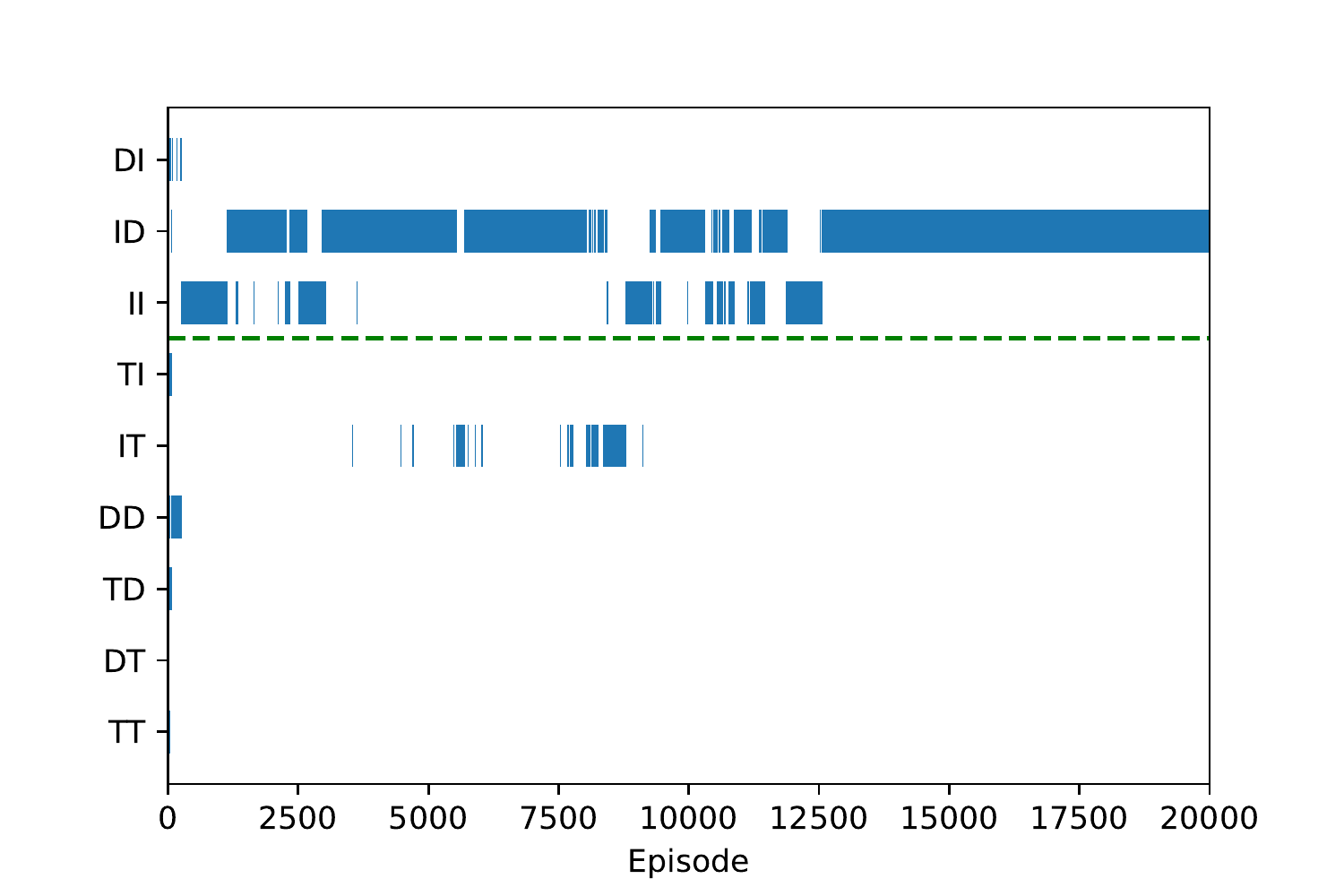}
         \caption{Policy ID with decayed learning rate}
         \label{fig:d-Original-ID}
    \end{subfigure}
    \caption[The Policy chart for baseline method with the decayed learning rate]
    {The Policy chart for baseline method with the decayed learning rate in original Space Traders Environment}
    \label{fig:d-Original-Policy-Chart}
\end{figure}
\begin{table}[hbt!]
    \centering
    \begin{tabular}{@{}|c|c|c|c|c|@{}}
    \toprule
    Policy & DI & ID & II & IT  \\ \midrule
    Constant learning rate & 1 & 13 & 4  & 2    \\ \midrule
    Decayed learning rate & 0 & 20 & 0  & 0    \\ \bottomrule
    \end{tabular}%
    \caption[20 independent runs of the Algorithm \ref{algo:moql-expected} with decayed learning rate]{The final greedy policies learned in 20 independent runs of the Algorithm \ref{algo:moql-expected} with decayed learning rate for Space Traders environment}
    \label{tab:Space-Traders-Baseline-d}
\end{table}
As we can see from table \ref{tab:Space-Traders-Baseline-d}, agent converges to sub-optimal policy ID 20 out of 20 this time with the decayed learning rate. The policy chart in Figure \ref{fig:d-Original-Policy-Chart} also indicates that by gradually decaying the learning rate for agent which reduces the influences of the occasional unsuccessful or successful runs from stochastic environment. After 15,000 episode, agent converges to single policy until end of experiment. So clearly the decayed learning rate helps to eliminate the impact of environmental stochasticity on this agent, allowing it to reliably converge to the same solution. But still the baseline method suffers from the main stochastic SER issue as we discussed early in chapter 4, as the solution it settles on is not actually SER-optimal.
\section{Single-phase MOSS Result and Discussion}
\begin{figure}[hbt!]
    \centering
    \begin{subfigure}[h]{0.45\textwidth}
         \centering
         \includegraphics[width=\textwidth]{Images/MOSS-DI.pdf}
         \caption{Policy DI with constant learning rate}
         \label{fig:d-MOSS-ID-1}
    \end{subfigure}
     \begin{subfigure}[h]{0.45\textwidth}
         \centering
         \includegraphics[width=\textwidth]{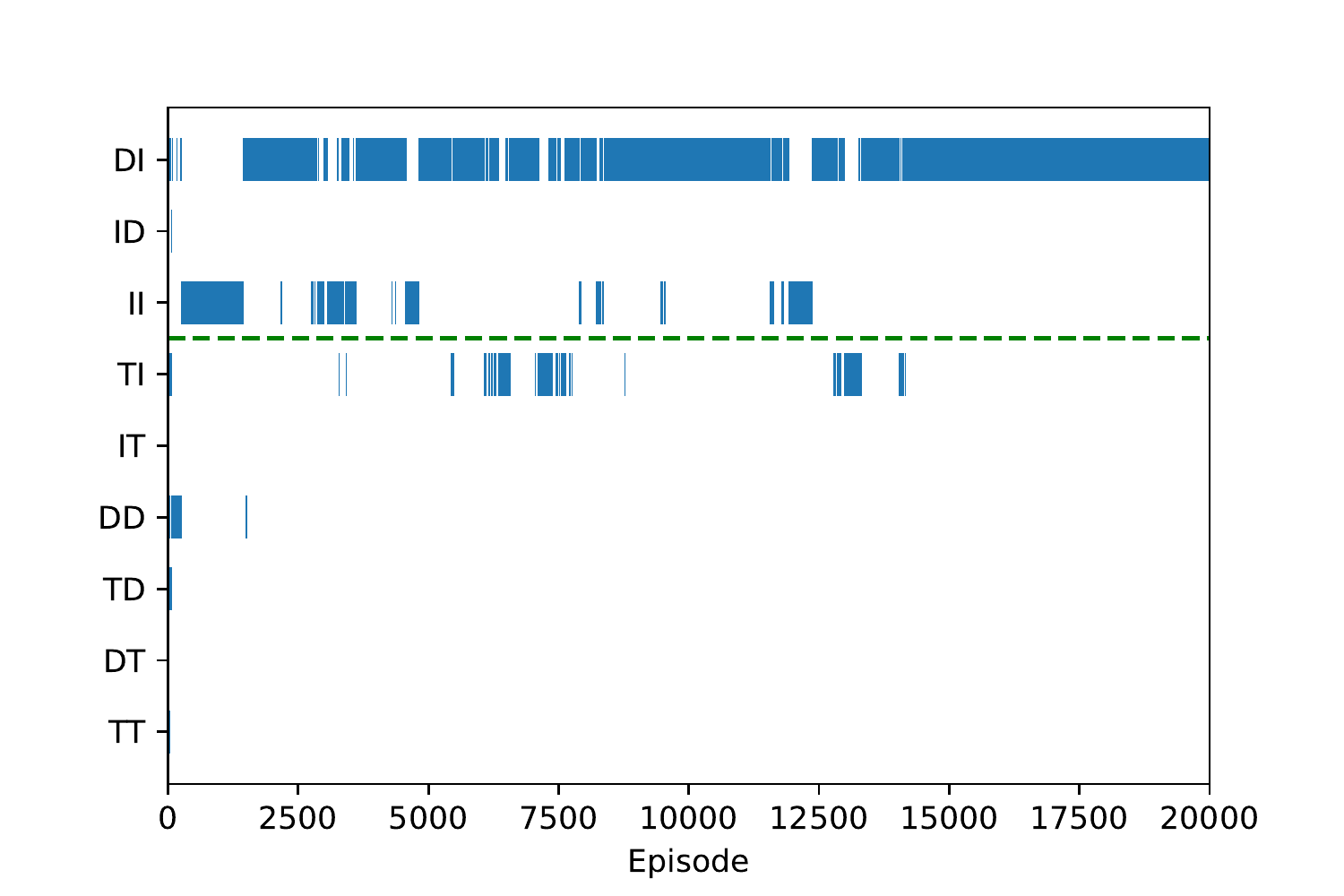}
         \caption{Policy DI with decayed learning rate}
         \label{fig:d-MOSS-ID}
    \end{subfigure}
    \caption[The Policy chart for single-phase MOSS algorithm with the decayed learning rate]
    {The Policy chart for single-phase MOSS algorithm with the decayed learning rate in original Space Traders Environment}
    \label{fig:d-MOSS-Policy-Chart}
\end{figure}
\begin{table}[hbt!]
    \centering
    \begin{tabular}{@{}|c|c|c|c|c|c|@{}}
    \toprule
    Policy & DI & ID & II & IT & TI \\ \midrule
    Constant learning rate & 15 & 0 & 0 & 3 & 2 \\ \midrule
    Decayed learning rate & 20 & 0 & 0 & 0 & 0 \\ \bottomrule
    \end{tabular}%
    \caption[20 independent runs of the Algorithm \ref{algo:mossql}]{The final greedy policies learned in 20 independent runs of the single-phase MOSS algorithm for Space Traders environment}
    \label{tab:Space-Traders-MOSS-d}
\end{table}
The single-phase MOSS algorithm has the similar pattern with baseline method. Agent successfully finds out the desired optional policy DI around 15,000 episodes in Figure \ref{fig:d-MOSS-Policy-Chart}. Compared with the policy chart on the left where agent is still struggling to stabilize the final policy before the end of experiment. The results in Table \ref{tab:Space-Traders-MOSS-d} show that, for the original Space Traders environment, the combination of single-Phase MOSS algorithm and a decayed learning rate does reliably converge to the correct SER-optimal policy. \newline\newline
But when we apply them in the new variant of Space Traders Environment, agent fails to find out the desired optional policy ID again as showed in Table \ref{tab:Space-Traders-MOSS-v-d}. Whether using constant or decayed learning rate, the single-phase MOSS algorithm is still having the same problems which has been covered early in chapter 6.
\begin{table}[hbt!]
    \centering
    \begin{tabular}{@{}|c|c|c|c|c|c|@{}}
    \toprule
    Policy & DI & ID & II & IT & TI \\ \midrule
    Constant learning rate & 15 & \textcolor{red}{0} & 0 & 3 & 2 \\ \midrule
    Decayed learning rate & 20 & \textcolor{red}{0} & 0 & 0 & 0 \\ \bottomrule
    \end{tabular}%
    \caption[20 independent the runs of single-phase MOSS algorithm with the decayed learning rate in new variant of Space Traders]{The final greedy policies learned in 20 independent runs of the single-phase MOSS algorithm with the decayed learning rate in new variant of Space Traders environment}
    \label{tab:Space-Traders-MOSS-v-d}
\end{table}

\section{Two-phase MOSS Result and Discussion}
\begin{figure}[hbt!]
    \centering
     \begin{subfigure}[h]{0.45\textwidth}
         \centering
         \includegraphics[width=\textwidth]{Images/MOSSTP-ID.pdf}
         \caption{Policy ID with constant learning rate}
         \label{fig:MOSSTP-ID-1}
    \end{subfigure}
    \begin{subfigure}[h]{0.45\textwidth}
         \centering
         \includegraphics[width=\textwidth]{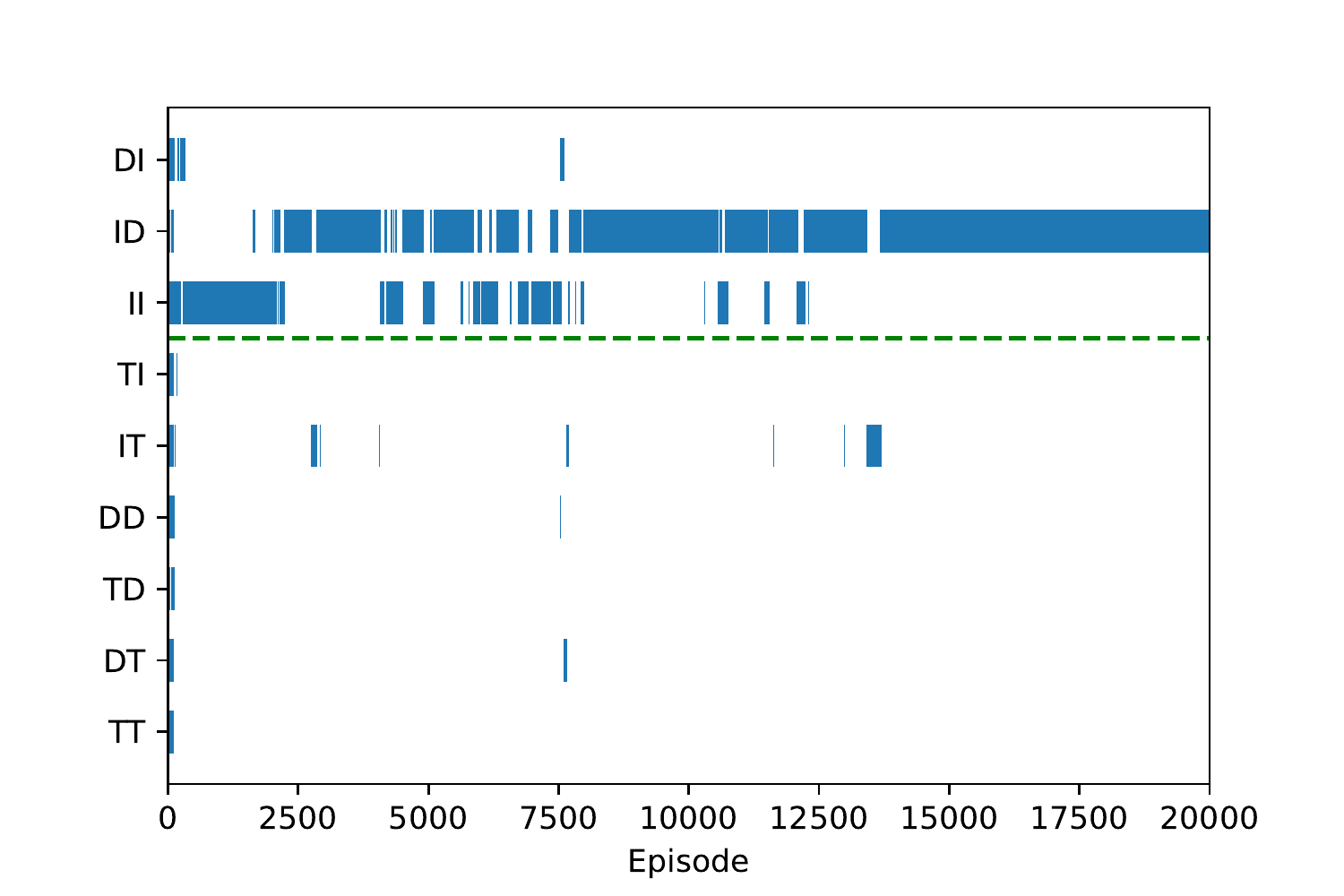}
         \caption{Policy ID with decayed learning rate}
         \label{fig:d-MOSSTP-ID}
    \end{subfigure}
    \begin{subfigure}[h]{0.45\textwidth}
         \centering
         \includegraphics[width=\textwidth]{Images/MOSSTP-DI.pdf}
         \caption{Policy DI with constant learning rate}
         \label{fig:d-MOSSTP-DI-1}
    \end{subfigure}
    \begin{subfigure}[h]{0.45\textwidth}
         \centering
         \includegraphics[width=\textwidth]{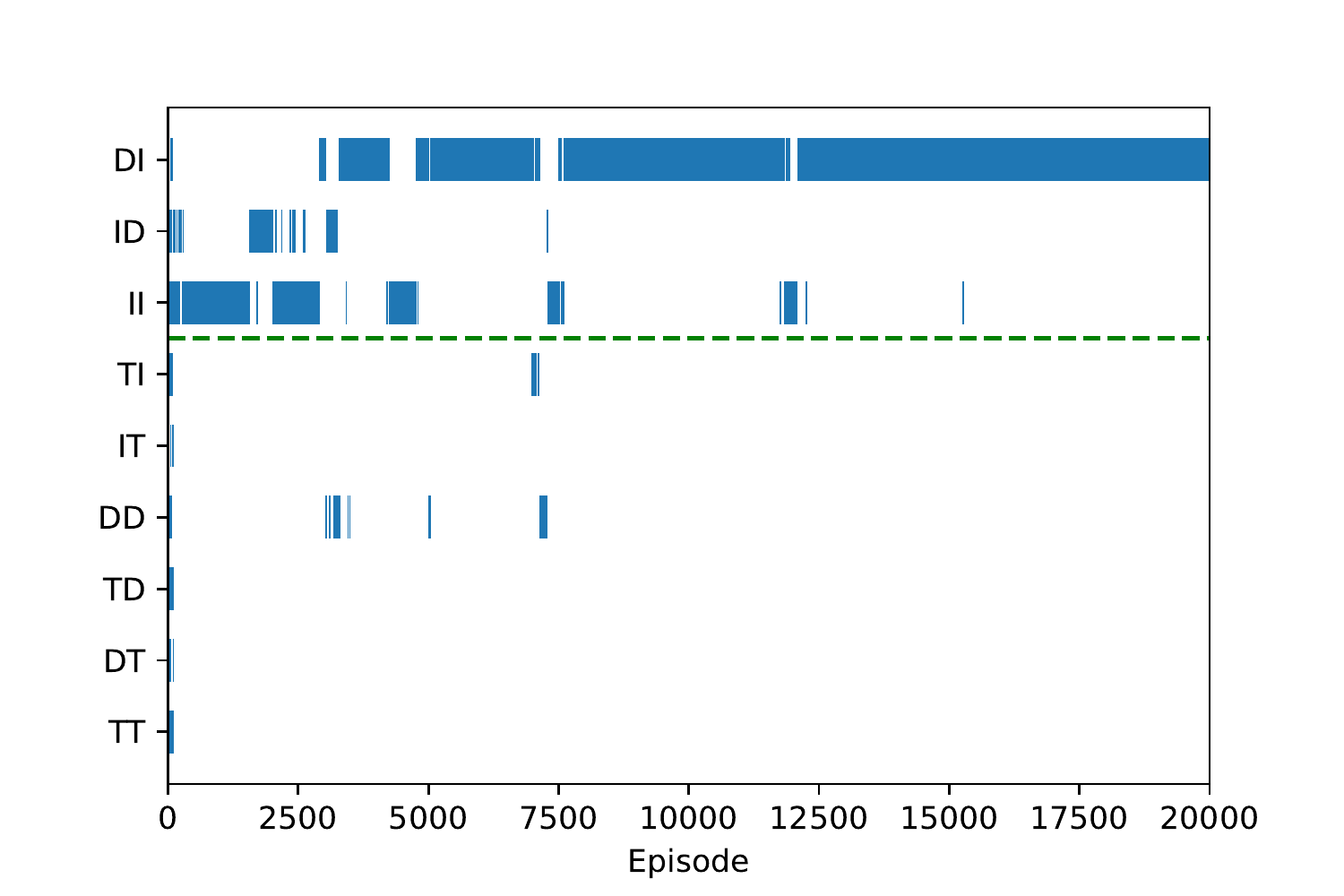}
         \caption{Policy DI with decayed learning rate}
         \label{fig:d-MOSSTP-DI}
    \end{subfigure}
    \caption[The Policy chart for two-phase MOSS algorithm with the decayed learning rate]
    {The Policy chart for two-phase MOSS algorithm with the decayed learning rate in original Space Traders Environment}
    \label{fig:d-MOSSTP-Policy-Chart}
\end{figure}

Compared with single-phase MOSS algorithm, the decayed learning rate did not change much in the final result of 20 independent trials for two-phase MOSS algorithm as we can see from table \ref{tab:Space-Traders-MOSSTP-d}. One potential explanation is that the global statistic that are gathered during the data collection phase already assist in reducing the problem of noisy estimates. But the downside of two-phase MOSS algorithm still exists which is easy trapped into sub-optimal policy as showed in Figure \ref{fig:d-MOSSTP-Policy-Chart}
\begin{table}[hbt!]
    \centering
    \begin{tabular}{@{}|c|c|c|c|c|c|@{}}
    \toprule
    Policy & DI & ID & II & IT & TI \\ \midrule
    Constant learning rate & 13 & 6 & 1 & 0 & 0 \\ \midrule
    Decayed learning rate & 13 & 7 & 0 & 0 & 0 \\ \bottomrule
    \end{tabular}%
    \caption[20 independent runs of the Algorithm \ref{algo:mossql-phased} with the decayed learning rate]{The final greedy policies learned in 20 independent runs of the two-phase MOSS algorithm with the decayed learning rate for Space Traders environment}
    \label{tab:Space-Traders-MOSSTP-d}
\end{table}


\section{Conclusion}
Compared with constant learning rate, the decayed learning rate does help to mitigate the noisy estimates issue in both baseline method and MOSS algorithm. But even with the noisy estimates problem fixed, these methods still fail to solve the major stochastic SER issue, as we already discussed it early in this research.
\chapter{Conclusion}
\label{chapter10}
An extension of scalar value Q-learning, multi-objective Q-learning algorithm, has been widely used in the multi-objective reinforcement learning literature. This research builds on prior works, and focuses on what factors influence the frequency with which value-based MORL Q-learning algorithms find out the SER optimal policy combined with using non-linear scalarisation function under stochastic state environment.
\section{Major Findings}
There are three aims that in this study which were explored. The first one is how different reward signal affects MORL agent's performance in stochastic environments. The second aim was to investigate what impact does the augmented state with global statistics have on MORL agent's ability to learn in stochastic environment. The final aim of this study was to find out what effect does the use of the options have on MORL agent's capacity to learn under stochastic environment. \newline \newline
The results for the first scenario clearly showed that with the new reward signal,the baseline method does improve the accuracy to find out the desired optimal policy in original Space Traders problem. However, just by modifying the reward signal is not enough to address issues in stochastic environments under SER criteria. Because in general, it may be too hard or even impossible to design a suitable reward structure. \newline \newline
It was found in the second case that the augmented state combined with use of global statistics in MOSS algorithm clearly outperforms the baseline method in original Space Traders problem. However, there is still a lot of improvement need to make for how agent collects those global information as the MOSS algorithm fails to find out SER optimal policy in general.\newline \newline
The results for the third aim reveals that options learning is able to solve non-linear scalarisation function for SER criteria under relatively small stochastic environment likes Space Traders Problem. Which means this method still fails from a more fundamental problem – the curse of dimension. Because each options need to be pre-defined before running the experiment. Therefore, for the problems with more states and actions, the numbers of pre-defined options are going to increase exponentially. In another word, this method is not able to scale up to address more complex problem in real-life.\newline \newline
While not an initial aim of this project, a key finding of this work is in the final experiment of Option learning which discovered the extent to which the issue of noisy Q-value estimates. Combined with stochastic environment, TLO action-selection and constant learning rate, the optimal policy never stabilized at end of each training. The further experiments reveals that with the help of decayed learning rate, all the methods including baseline algorithm was able to mitigate the influences of the occasional unsuccessful or successful runs from stochastic environment and converge to one final policy.
\section{Implications}
Because of the flaws in each investigated methods, none of them could be directly applied into real-world applications. However, there are variety of applications which are likely to involve a stochastic environment and SER criteria as the goal. The first good example is the advanced traffic control System \shortcite{jin2019multi}. The function of the system is to make trade-offs between various policy goals, such as energy efficiency and traffic mobility. Another example is the Large-Scale Power System \shortcite{Deng2018Integrated}, due to the short-term voltage security problems, traditional solution is normally suffered from expensive cost and load loss. Therefore a coordinated optimization strategy for generators and capacitor banks is more efficient way to mitigate this short-term voltage crisis.
\section{Conclusion and Future directions}
There are two issues existing for MOQ-learning in stochastic environments (the core stochastic SER issue AND noisy Q value estimates), therefore a successful algorithm must address both of those problems together.\newline\newline
The first recommendation for future research is to look at policy-based methods such as Policy gradient. As these methods directly maximise the policy as a whole by defining a set of policy parameters, therefore they do not have the local decision-making issue faced by model-free value-based methods such as MOQ-learning. Several researchers
have developed and assessed policy-based methods for multi-objective problems \shortcite{parisi2014policy} \shortcite{bai2021joint}. However most policy-based MORL methods produce stochastic policies, whereas in some applications deterministic policies may be required. So these algorithms may required modification in order to deal with this constraint.\newline \newline
The second research direction should investigate \acrfull{DRL}. The conventional value-based RL learns a single value per state-action pair which is representing the expected return. Distributional reinforcement learning on the other hand works directly with the full distribution of the reward instead. This can be beneficial for MORL, as shown by \shortciteA{hayesmulti2022} who applied distributional multi-objective Distributional Value Iteration to find optimal policies for the ESR criteria. Therefore it could also potentially solve both the noisy estimates and stochastic SER issues.


\pagebreak
\bibliographystyle{apacite}
\bibliography{Reference}

\end{document}